%% file: main.tex
\PassOptionsToPackage{table}{xcolor}
\documentclass{article}

\usepackage{times}
\usepackage{latexsym}
\usepackage{graphicx}
\usepackage{tikz}
\usetikzlibrary{arrows.meta,backgrounds,calc,positioning}
\usepackage{subcaption}

\usepackage{hyperref}

\usepackage{titletoc}

\let\originaladdcontentsline\addcontentsline
\usepackage[preprint]{neurips_2026}
\let\addcontentsline\originaladdcontentsline

\usepackage{mathtools}
\usepackage{amsthm}

\usepackage[capitalize,noabbrev]{cleveref}

\usepackage[utf8]{inputenc}
\usepackage[T1]{fontenc}
\usepackage{url}
\usepackage{fontawesome5}
\usepackage{booktabs}
\usepackage{amsfonts}
\usepackage{amsmath}
\usepackage{nicefrac}
\usepackage{microtype}
\usepackage{inconsolata}
\usepackage[bb=boondox]{mathalfa}
\usepackage{booktabs,arydshln}

\makeatletter
\def\adl@drawiv#1#2#3{        \hskip.5\tabcolsep
        \xleaders#3{#2.5\@tempdimb #1{1}#2.5\@tempdimb}                #2\z@ plus1fil minus1fil\relax
        \hskip.5\tabcolsep}
\newcommand{\cdashlinelr}[1]{  \noalign{\vskip\aboverulesep
           \global\let\@dashdrawstore\adl@draw
           \global\let\adl@draw\adl@drawiv}
  \cdashline{#1}
  \noalign{\global\let\adl@draw\@dashdrawstore
           \vskip\belowrulesep}}
\makeatother

\usepackage[acronym,nomain]{glossaries}
\usepackage{amssymb}
\usepackage{algorithm}
\usepackage{multirow}
\usepackage{caption}
\usepackage{enumitem}
\setlist[enumerate]{label=\Roman*.} \newlist{checklist}{enumerate}{2}
\setlist[checklist]{label=\arabic*.}
\usepackage{changepage}
\usepackage[most]{tcolorbox}
\usepackage{soul}

\newtcolorbox{softbox}[1][]{
  enhanced,
  breakable,
  colback=gray!5,
  colframe=black!10,
  coltext=black!80,
  colbacktitle=gray!5,
  coltitle=black!80,
  fonttitle=\sffamily\bfseries\scriptsize,
  title=#1,
  lefttitle=20pt,
  righttitle=20pt,
  toptitle=5pt,
  bottomtitle=4pt,
  boxrule=0.8pt,
  arc=10pt,
  left=5pt,
  right=5pt,
  top=5pt,
  bottom=5pt,
  fontupper=\sffamily\scriptsize,
}

\definecolor{red}{rgb}{0.96, 0.00, 0.29}
\definecolor{green}{rgb}{0.22, 1.00, 0.53}
\newcommand{\gain}[1]{\cellcolor{green!40}{#1}}
\newcommand{\loss}[1]{\cellcolor{red!25}{#1}}
\newcommand{\same}[1]{\cellcolor{gray!25}{#1}}
\newcommand{\smallgain}[1]{\cellcolor{green!10}{#1}}
\newcommand{\none}[1]{#1}

\newcommand\doubleplus{+\kern-1.8ex+\kern0.8ex}

\newacronym{llm}{LLM}{Large Language Models}
\newacronym{musr}{MuSR}{Multistep Soft Reasoning}

\title{ReHoPER: Receding-Horizon Planning\\for Enhanced Reasoning}

\author{  Saeed Ahmadnia \\
  Department of Computer Science\\
  University of Illinois Chicago\\
  \texttt{\color{blue} sahmad67@uic.edu} \\
  \And
  Cornelia Caragea \\
  Department of Computer Science\\
  University of Illinois Chicago\\
  \texttt{\color{blue} cornelia@uic.edu} \\
}

\begin{document}

\maketitle

\begin{abstract}
We propose ReHoPER, an inference-only, zero-shot method that improves large language models' reasoning by generating and answering intermediate questions along multiple paths before the final answer. It iteratively plans a horizon of candidate intermediate questions, selects one to answer, and replans from the updated history. ReHoPER is task-agnostic, using the same generic instructions across datasets and models without labeled data or task-specific prompt design. Across multiple datasets, including iLLC, a new controlled benchmark for compositional reasoning, ReHoPER outperforms strong baselines, with the largest gains in the most compositional settings. Our implementation\footnote{\mbox{ }\faGithub \color{blue} \mbox{ } \url{https://github.com/theSaeed/mobius}} and the iLLC generator\footnote{\mbox{ }\faGithub \color{blue} \mbox{ } \url{https://github.com/theSaeed/iLLC}} are publicly available to support future work.
\end{abstract}

\input{figures/the-figure}

\section{Introduction}

Large Language Models (LLMs) trained at scale can be prompted in natural language to perform a wide range of NLP tasks, achieving strong zero- and few-shot performance, often close to fine-tuned systems, without task-specific training~\citep{brown2020language,min2021survey}.
Instruction tuning and reinforcement learning from human feedback further improve their ability to follow instructions and generalize to new tasks, reinforcing LLMs as general-purpose, prompt-driven systems rather than narrowly trained models~\citep{wei2022finetuned,ouyang2022training}.
Despite this progress, benchmarks designed to probe multi-step and compositional reasoning, such as MoreHopQA \citep{Schnitzler2024MoreHopQAMT} and MuSR \citep{sprague2024musr}, show that even state-of-the-art models still struggle to integrate multiple pieces of evidence and handle long-range dependencies.
\looseness=-1
Inference-time reasoning has improved with methods such as Tree-of-Thoughts \citep{Yao23neurips}, StrategyLLM \citep{gao2024strategyllm}, and Self-Ask \citep{press-etal-2023-measuring}, which guide models through multi-step procedures like search, strategy retrieval, or follow-up questioning.
However, these approaches often depend on labeled data (e.g., solved examples for learning or retrieving strategies) or a manually crafted reasoning structure tailored to each new task.
In contrast, Chain-of-Thought (CoT) \citep{wei2022chain} and Plan-and-Solve (PS) \citep{wang-etal-2023-plan} use simple, reusable instructions (e.g., ``let's think step by step'' or ``first outline a plan, then solve the problem'') to enable oracle-free reasoning across tasks, and Self-Discover (SD) \citep{zhou2024selfdiscover} further reduces manual effort by letting the model generate its own reasoning structure at inference time.
Still, these methods mostly follow linear reasoning patterns, and can potentially benefit from richer, more flexible planning mechanisms in the spirit of Tree-of-Thoughts and StrategyLLM, while remaining oracle-free and task-agnostic.

In this work, we introduce Receding-Horizon Planning for Enhanced Reasoning (ReHoPER), which targets this design space from a different angle. Rather than prescribing a fixed sequence of explanations or relying on discovered task-specific templates, we guide the model to generate and answer intermediate questions about the current instance before committing to a final prediction. ReHoPER operates entirely at inference time and remains task-agnostic as the same generic instructions for proposing and answering intermediate questions can be used across tasks, models, and answer-format prefixes (Input-Output, CoT, and PS) without any task-specific labeled data or demonstrations.
Figure~\ref{fig:rehoper_overview} summarizes the planning loop and how predictions from multiple paths are combined.

ReHoPER follows a receding-horizon strategy. At each step, it generates a list of intermediate questions and answers the first unresolved one given the current history, loosely analogous to classical receding-horizon control \citep{propoi1963application,kwon1977modified}.
This alternating loop of planning and execution combines look-ahead with step-by-step adaptivity. Each answer updates the reasoning state, triggering replanning that can revise, refine, or discard previously proposed questions, enabling a dynamically evolving reasoning process rather than a static question sequence.
We evaluate ReHoPER against strong baselines
on MuSR, MoreHopQA, and iLLC, a new controlled benchmark for compositional execution, using a wide range of LLMs. Across these settings, ReHoPER improves over the compared methods under matched prompting conditions.

Our contributions are as follows: (i) we introduce ReHoPER, a task-agnostic, inference-only method that uses adaptive planning to generate and answer intermediate questions across multiple reasoning paths and aggregates their predictions; (ii) we introduce iLLC, a controlled diagnostic benchmark for compositional execution; (iii) we provide a comprehensive empirical study that compares ReHoPER against strong baselines from prior work across diverse models, datasets, and reasoning styles; (iv) we offer a systematic analysis of ReHoPER design choices and behaviors, through ablations and trajectory analyses; and (v) we publish our implementation and iLLC data generator to facilitate further research.

\section{Related work}

Prompting strategies play a central role in eliciting reasoning behavior from LLMs. Early work predominantly evaluated LLMs using Input–Output (IO) prompting, where models directly predict a final answer without exposing intermediate reasoning. This paradigm changed with the introduction of Chain-of-Thought (CoT) prompting, which encourages models to generate explicit reasoning steps in natural language through either few-shot exemplars with rationales \citep{wei2022chain} or minimal zero-shot instructions such as “Let’s think step by step” \citep{kojima22neurips}.
However, while zero-shot CoT can elicit step-by-step reasoning, it does not require the model to commit to a complete reasoning outline, leading to missing steps and omission errors. Plan-and-Solve (PS) prompting \citep{wang-etal-2023-plan} addresses this limitation by decoupling planning from execution: the model first generates an explicit plan and then executes it, effectively encouraging structured reasoning via a “plan, then solve” instruction.

Several methods decompose compositional tasks into subquestions. Self-Ask \citep{press-etal-2023-measuring} generates and answers follow-up questions before producing a final response, while Least-to-Most (LtM) prompting \citep{ZhouSHWS0SCBLC23} first produces subproblems and then solves them sequentially. However, both rely on curated exemplars and predefined reasoning templates, in which the decomposition is determined upfront and executed in a single pass.  Self-Discover (SD) \citep{zhou2024selfdiscover} removes manual design by composing a reasoning structure in a zero-shot setting. Still, it commits to a fixed plan before solving, limiting its ability to adapt as new information emerges during reasoning.
ReHoPER instead uses generic instructions to generate and answer intermediate questions for each instance, gathers predictions along multiple paths, and aggregates them. Within each path, it selects one question from a short plan before updating the plan from the new answer. This allows subsequent questions to reflect newly obtained information without committing to a fixed decomposition.

This iterative plan–execute–replan paradigm connects ReHoPER to classical planning and control, particularly receding-horizon control (RHC), also known as model predictive control, where decisions are made by optimizing over a short horizon, executing one step, and replanning from the updated state \citep{propoi1963application,kwon1977modified,computers14030104}. However, unlike classical RHC, ReHoPER does not rely on an externally defined reward function; instead, it uses the model’s own predictions of the goal question, conditioned on the evolving reasoning state, as an implicit evaluation signal.

Beyond explicit decomposition, several methods enhance LLM reasoning through iterative sampling, structured search, or self-reflection. Self-consistency \citep{wang2023selfconsistency} samples multiple independent CoT trajectories and aggregates their final answers, using diversity to reduce variance. Tree-of-Thoughts (ToT) \citep{Yao23neurips} extends this idea by explicitly exploring a tree of intermediate reasoning states using search and evaluation, enabling broader exploration at the cost of increased inference complexity. StrategyLLM \citep{gao2024strategyllm} commits to a high-level plan prior to execution. However, these approaches typically rely on a fixed plan (with no ability to re-plan) or costly search over multiple branches.
Self-Refine \citep{madaan2023selfrefine} and CoVe \citep{dhuliawala-etal-2024-chain} iteratively improve outputs via self-feedback and verification, RISE \citep{he-etal-2025-rise} combines decomposition, retrieval, and self-critique through iterative self-exploration, and Think Twice \citep{tian2025thinktwiceenhancingllm} repeatedly re-answers the same question to correct errors. Different from the above approaches, ReHoPER integrates lightweight planning and adaptive question selection within multiple independent forward reasoning trajectories, enabling targeted exploration without handcrafting reasoning paths, manually labeled data, or post-hoc revision.

\section{Methodology}

We propose ReHoPER (\emph{Receding-Horizon Planning for Enhanced Reasoning}), an inference-only method that guides an instruction-tuned LLM through intermediate questions before answering the goal question \(Q^*\). ReHoPER runs multiple stochastic reasoning paths and within each path repeatedly looks ahead by generating a horizon of candidate intermediate questions conditioned on the current path history. It then executes only one question, incorporates the resulting answer into the history, and replans from this updated state. This makes the process inherently dynamic: each newly obtained intermediate answer can change what information is missing, render some planned questions irrelevant, or suggest a better follow-up direction. Thus, ReHoPER adapts its reasoning trajectory dynamically, correcting earlier plans and incorporating newly revealed information on the fly without committing to a fixed plan. By combining diversity across paths with per-step look-ahead planning and replanning within each path, ReHoPER maintains an organized decomposition while enabling course correction, leading to more reliable final predictions.

We formalize the task setting in Section~\ref{sec:problem}, defining the inference-only objective and notation.
In Section~\ref{sec:rehoper}, we then present ReHoPER in detail, describing its receding-horizon planning, path rollout, and aggregation procedure.

\subsection{Problem statement}
\label{sec:problem}

Each task example consists of an optional context \(C\) and a goal question \(Q^*\). Given an instruction-tuned LLM \(f\), the objective is to produce a prediction \(\hat{y}\) that matches the ground-truth label \(y\).
The proposed method operates in a fully inference-only setting and does not rely on any task-specific instruction, labeled examples, or additional unlabeled data beyond the given task test example.

\subsection{ReHoPER}
\label{sec:rehoper}

\looseness=-1
ReHoPER runs \(n\) stochastic reasoning paths by enabling non-deterministic decoding during model inference (e.g., via temperature sampling), and organizes each path as a sequence of intermediate questions and answers generated in a receding-horizon manner: at every step, it briefly looks ahead to organize a candidate trajectory of questions, but executes only one newly selected question before replanning from the updated history.
Intermediate questions are generated by prompting the model with a specific instruction \(I\) (Appendix~\ref{app:instruct}). For path \(i\in\{1,\,\ldots,\,n\}\), let \(m_i\) denote the number of intermediate questions ultimately asked along that path. The history of path \(i\) at step \(k\in\{0,\,\ldots,\,m_i\}\) is
\begin{equation}
\mathbf{H}^{(i)}_{k}=\big\langle (IQ^{(i)}_1,\ A^{(i)}_1),\ \ldots,\ (IQ^{(i)}_{k},\ A^{(i)}_{k})\big\rangle,\ \ \ \
\mbox{where }\ \mathbf{H}^{(i)}_{k=0}=\emptyset. \mbox{      }
\end{equation}
Here, $(IQ^{(i)}_k, A^{(i)}_k), k=1..m_i$ represent intermediate (asked/answered) questions and their answers, respectively. For simplicity, we denote $\mathbf{H}^{(i)}_{k=0}=\mathbf{H}_0$. \(I_{\mathrm{rh}}\) is receding-horizon instruction.
\paragraph{Initial prediction.} Prompt \(f\) once with context, empty history, and goal question \((C,\ \mathbf{H}_0,\ Q^*)\) to obtain response \(R_0\) and post-process \(R_0\) to compute an initial prediction \(\hat{y}_0\).

\paragraph{Path rollout.}
For each path \(i\), for steps \(k=1,2,\ldots\) until termination ($m_i$ steps or a user-defined cap of \(k=m_{\max}\)):
\begin{enumerate}[leftmargin=2.5em]
    \item \textbf{Plan.} Prompt \(f\) with \((C,\ \mathbf{H}^{(i)}_{k-1},\ I_{\mathrm{rh}})\) to instruct the model to generate a horizon of candidate questions:
    \(     \widetilde{\mathbf{IQ}}^{(i)}_{k}
    =\big\langle \widetilde{IQ}^{(i)}_{k,1},\ \ldots,\ \widetilde{IQ}^{(i)}_{k,h}\big\rangle,
    \)     where \(h\) is determined by \(f\) and may vary across steps and paths; $\widetilde{\mathbf{IQ}}^{(i)}_{k}$ is the set of generated questions.
    \item \textbf{Select.} Choose the first previously unanswered candidate question from $\widetilde{\mathbf{IQ}}^{(i)}_{k}$:
    \begin{equation}
    j^*=\min\Bigl\{j:\widetilde{IQ}^{(i)}_{k,j}\notin \{IQ^{(i)}_{1},\ \ldots,\ IQ^{(i)}_{k-1}\}\Bigr\},
    \label{jstar}
    \end{equation}
    and set
    \(     IQ^{(i)}_{k}
    \;\leftarrow\;
    \widetilde{IQ}^{(i)}_{k,j^*}.
    \)     If no such \(j^*\) exists (all candidates repeat or no valid question is produced), the path terminates. The remaining questions
    \(\widetilde{\mathbf{IQ}}^{(i)}_{k}\setminus\langle IQ^{(i)}_{k}\rangle\), where \(\setminus\) denotes difference, are discarded, and a fresh plan is generated at the next step. Note that if $k=1$ in Eq. \ref{jstar}, $j^*=1$, since no intermediate questions have yet been asked in the current path.
    \item \textbf{Answer.} Prompt \(f\) with \((C,\ \mathbf{H}^{(i)}_{k-1},\ IQ^{(i)}_k)\) to get \(A^{(i)}_k\).
    \item \textbf{Update.} Set
    \(     \mathbf{H}^{(i)}_{k}
    \leftarrow
    \mathbf{H}^{(i)}_{k-1}\ \|\ \big\langle (IQ^{(i)}_{k},\ A^{(i)}_{k})\big\rangle,
    \)     where \(\|\) denotes concatenation of sequences.
    \item \textbf{Predict.} Prompt \(f\) with \((C,\ \mathbf{H}^{(i)}_{k},\ Q^*)\) to obtain response \(R^{(i)}_k\), and post-process to extract candidate prediction \(\hat{y}^{(i)}_k\) for step $k$ of path $i$.
\end{enumerate}

\paragraph{Aggregation.}
Collect the list of predictions, from the initial step and all the steps $k$ of all paths $i$:
\begin{equation}
\langle \hat{y}_0\rangle \;\mathbin{\|}\; \Big( \mathbin{\|}_{i=1}^{n}\; \mathbin{\|}_{k=1}^{m_i} \langle \hat{y}^{(i)}_{k}\rangle \Big).
\end{equation}
Since \(\mathbf{H}_0=\emptyset\) is identical across all paths, the prediction at step \(k=0\) is counted at most once in the final aggregation. We apply a simple majority vote (ties broken alphabetically) over the predictions to incorporate a form of Self-Consistency across reasoning paths and intermediate steps: the final prediction \(\hat{y}\) is defined as the most frequent value, and is evaluated against the ground-truth label \(y\).

\vspace{-0.3em}
\section{Experimental setup}
\label{sec:setup}

\vspace{-0.3em}
\subsection{Datasets and tasks}

We evaluate ReHoPER on three reasoning benchmarks spanning long-context narrative inference, multi-hop evidence aggregation, and controlled compositional execution: MuSR~\citep{sprague2024musr}, MoreHopQA~\citep{Schnitzler2024MoreHopQAMT}, and our proposed \(i\)-th to Last Letter Concatenation (iLLC) benchmark. \textbf{MuSR} includes three long-context reasoning tasks, \textbf{MM} (Murder Mysteries), \textbf{OP} (Object Placement), and \textbf{TA} (Team Allocation), each requiring models to integrate scattered clues and constraints. \textbf{MoreHopQA} tests multi-hop question answering over contexts that require connecting multiple pieces of evidence. We follow the original setup by evaluating on the 150-example subset selected by the dataset authors. Finally, \textbf{iLLC} generalizes the established Last Letter Concatenation (LLC)~\citep{wei2022chain} by requiring models to extract the \(i\)-th-to-last character from each word and concatenate the resulting letters; \(i=1\) recovers standard LLC. We denote each subset as L\{\(i\)\}-\{\(w\)\}, where \(i\) is the target character position from the end and \(w\) is the number of words in the sequence. Our main experiments use \textbf{L1-4}, \textbf{L1-6}, \textbf{L2-4}, and \textbf{L2-6}, while additional results on the longer \textbf{L1-8} and \textbf{L2-8} settings are reported in Appendix~\ref{app:illc}.

We use iLLC to study compositional execution in a setting where the required operations are known. Solving an instance requires extracting one character from each word, preserving the order of the words, and combining the characters into the final answer. This structure fits our study of intermediate-question planning because the problem can be broken into well-defined steps. It also lets us vary difficulty in a controlled way: \(i\) changes the character-extraction operation, while \(w\) changes the number of operations that must be composed. Since the task does not depend on external knowledge, its results are less affected by differences in factual recall or retrieval. In addition, the generator can produce new labeled instances at different difficulty levels, making the benchmark easy to scale and extend. We use iLLC to test procedural compositional reasoning under controlled conditions, complementing the broader reasoning settings covered by the natural-language benchmarks. Further details on iLLC and the other datasets, including statistics, examples, and post-processing procedures, are provided in Appendix~\ref{app:data-examples}.

\subsection{Language models}

We evaluate ReHoPER on a diverse set of instruction-tuned LLMs spanning multiple model families, parameter scales, and deployment regimes, including dense models from Llama, Gemma, Phi, and Mistral, as well as sparse Mixture-of-Experts models such as gpt-oss and Qwen3. To make larger models practical at inference time, we use quantized AWQ or GPTQ variants when needed. For Qwen3, we evaluate both its non-thinking mode and its thinking-enabled mode; unless a result is explicitly labeled as Qwen3 with thinking, Qwen3 denotes the non-thinking configuration. Full model and deployment details are provided in Appendix~\ref{app:models}, Table~\ref{tab:models}.

\subsection{Baselines and setup details}
We use three different assistant output prefixes: IO (input-output), CoT (chain-of-thought) \citep{wei2022chain}, and PS (plan-and-solve) \citep{wang-etal-2023-plan}. Concretely, \textbf{IO} leaves the assistant output unconstrained (empty prefix), while \textbf{CoT} prepends the assistant’s response with
``Let’s think step by step.''
and \textbf{PS} prepends it with
``Let's first understand the problem and devise a plan to solve the problem. Then, let's carry out the plan to solve the problem step by step.''.
Notably, adding CoT or PS explicitly forces a particular reasoning style and thus can override or mask the model’s native reasoning behavior that would otherwise emerge under IO. For each prompt variant, we pair it with \textbf{Self-Consistency (SC)} \citep{wang2023selfconsistency} as our first baseline by sampling 61 independent completions (same prompt with stochastic decoding), post-processing each completion into a candidate answer,
and returning the majority-vote answer among the
candidates (ties broken deterministically by choosing the alphabetically smallest answer).

As another baseline, we include the \textbf{Self-Discover (SD)} \citep{zhou2024selfdiscover}, adapted to our instance-specific setup. For each test instance, we first prompt the model using only that single unlabeled example to propose a high-level reasoning strategy and decomposition tailored to that instance (e.g., what intermediate checks to perform, what constraints to track, how to verify the final answer). We then turn this into a one-off reasoning template and use it to answer the same instance.
For completeness on iLLC, we also compare against \textbf{Least-to-Most (LtM)} prompting, its original \textbf{CoT} baseline~\citep{ZhouSHWS0SCBLC23}, and \textbf{Self-Refine (SR)}~\citep{madaan2023selfrefine}. For SR, we use instruction-only feedback and refinement prompts for five iterations, since no few-shot exemplars are available for our datasets. We treat these as analysis-only references rather than core baselines, as they rely on labeled exemplars, manually curated reasoning structures, or task-specific refinement prompts.

\looseness=-1
ReHoPER is designed to be prefix-agnostic and compatible with all SC assistant output prefix variants. Concretely, ReHoPER can be combined with \{IO, CoT, PS\} by simply applying the chosen prefix during the evaluation stages, yielding composite configurations of the form \text{ReHoPER} \(\times\) \{\text{IO}, \text{CoT}, \text{PS}\} (e.g., ReHoPER with CoT, etc.). This design allows us to flexibly adopt the prefix that best matches the requirements of a given task, while keeping the underlying reasoning and evaluation procedure unchanged.
We use \(n=6\) reasoning paths, a maximum depth of \(m_{\max}=10\) steps, and temperature \(0.5\). This configuration produces up to \(1 + n\cdot m_{\max} = 61\) candidate answers per instance (one from the initial evaluation plus up to \(n\cdot m_{\max}\) intermediate evaluation outputs), enabling a majority vote analogous to the 61-sample SC baselines. We choose \(n=6\) and \(m_{\max}=10\) as a pragmatic trade-off that balances breadth (path diversity) and depth (multi-step refinement). Appendix~\ref{app:hyperparams} reports the complete hyperparameter study, including the selected temperature and path-step budget.

\looseness=-1
For the planning ablations, we evaluate ReHoPER without replanning, generating the intermediate-question list once per path and answering it in order, and without look-ahead, generating a single next question from the updated history at each step. Both use the same answering, history-update, prediction, and aggregation procedures, output prefixes, and path and step limits as the full method. The former stops when its initial list is exhausted or the step limit is reached; the latter runs to the step limit. Both replace the candidate-selection step. We evaluate these ablations on MuSR MM, MoreHopQA, and iLLC L2-4; Appendix~\ref{app:planning-ablations} gives implementation details, and Appendix~\ref{app:instruct} provides the prompts.

\input{tables/result}

\section{Results and discussion}

\subsection{ReHoPER results}

\looseness=-1
Table~\ref{tab:result} compares ReHoPER with SC and SD across all datasets. Across all models, datasets, and prefixes, ReHoPER improves over the SC setup under the same answer-format prefix in a majority of cases, with stronger gains for more structured prefixes.
Appendix~\ref{app:output-samples} shows examples of how ReHoPER generates and uses intermediate questions in both correct and incorrect cases.
On average (bottom-right block), ReHoPER improves accuracy by 5.8pp for IO, 8.4pp for CoT, and 6.4pp for PS. These gains are calculated before rounding the accuracies and show that our approach is beneficial even when models already produce explicit chains of thought. Appendix~\ref{app:macro-results} shows that ReHoPER also improves over SC when each benchmark family is given equal weight, with gains of 3.9pp for IO, 6.6pp for CoT, and 4.8pp for PS. Appendix~\ref{app:robustness} shows that these gains are stable across repeated runs.
ReHoPER also often recovers and surpasses IO-level performance even when raw CoT or PS lags behind, indicating that the full intermediate-question procedure can help beyond the surface prompt alone.
Notably, ReHoPER tends to be strongest and most reliable when paired with PS, making it a suitable default for new tasks where it is unclear which prefix works best.

Remarkably, the gains are most pronounced on the hardest, most compositional settings.
On the iLLC~L2-$\{w\}$ variants, where the SC baseline performance is low, ReHoPER yields large improvements, e.g.,
14.3pp for IO, 25.4pp for CoT, and 20.9pp for PS on L2-6. We also observe near-consistent gains on MuSR MM, with average accuracy increasing by 6.2pp for IO, 4.5pp for CoT, and 3.6pp for PS, and smaller but positive improvements on other MuSR subsets and MoreHopQA. In contrast, tasks that are either more straightforward or closer to ceiling (e.g., iLLC~L1-$\{w\}$ variants) show only modest gains and occasionally small regressions, suggesting limited headroom for further improvement through additional planning.

The iLLC results also show that these controlled changes produce meaningful differences in difficulty. Performance drops substantially from L1, where the target character is at the word boundary, to L2, where its absolute position depends on word length. Performance also generally declines as the number of words increases. ReHoPER performs best relative to the baselines in these harder settings, suggesting that intermediate-question planning becomes more useful as extraction and composition become more demanding. These results demonstrate the value of iLLC as a controlled diagnostic of procedural compositional reasoning, complementing the broader reasoning settings represented by MuSR and MoreHopQA.

On average, SC already tends to outperform SD on all benchmarks, and ReHoPER further widens that gap. In the final Average block, SD reaches 43.4\% overall, whereas SC ranges from 56.1--59.9\% depending on the prefix, and ReHoPER raises these results to 62.3--66.3\%. Overall, the table supports the view that ReHoPER is a robust, model- and prompt-agnostic planner that strengthens test-time reasoning, with its largest impact in the deepest compositional settings, even rescuing cases where naive CoT or PS prompting alone would underperform a simple IO baseline.

\input{tables/vs}

\subsection{Effect of replanning and look-ahead}
Table~\ref{tab:planning-ablations-full} compares ReHoPER with ablations without replanning or look-ahead on MuSR MM, MoreHopQA, and iLLC L2-4. Across models and datasets, ReHoPER exceeds the stronger ablation by 1.9pp, 2.7pp, and 2.5pp under IO, CoT, and PS, respectively. The advantage holds for each dataset and prefix average and is largest on iLLC L2-4 (3.5pp, 6.1pp, and 5.9pp for IO, CoT, and PS), suggesting that the combined planning design is particularly useful in this more compositional setting. The overall averages also show that generating each next question from the updated history, even without look-ahead, performs better than following a fixed initial plan: the ablation without look-ahead achieves 60.8\%, 61.7\%, and 64.2\% under IO, CoT, and PS, compared with 59.0\%, 61.3\%, and 62.9\% without replanning. This pattern is consistent with the benefit of adapting question generation to newly acquired information. ReHoPER improves further on these averages by combining history-conditioned replanning with a horizon of candidate questions and explicit selection of an unasked question. These aggregate trends support the combined design, although individual configurations can favor an ablation. Appendix~\ref{app:planning-ablations} details the ablation procedures; Appendix~\ref{app:iq-stats} reports trajectory lengths and within-path question diversity; Appendix~\ref{app:trajectory-changes} shows that replanning often changes the initial question trajectory.

\subsection{Effect of answer aggregation}

\begin{figure}[t]
    \centering
    \includegraphics[width=1.\linewidth, trim=0 70 0 25, clip]{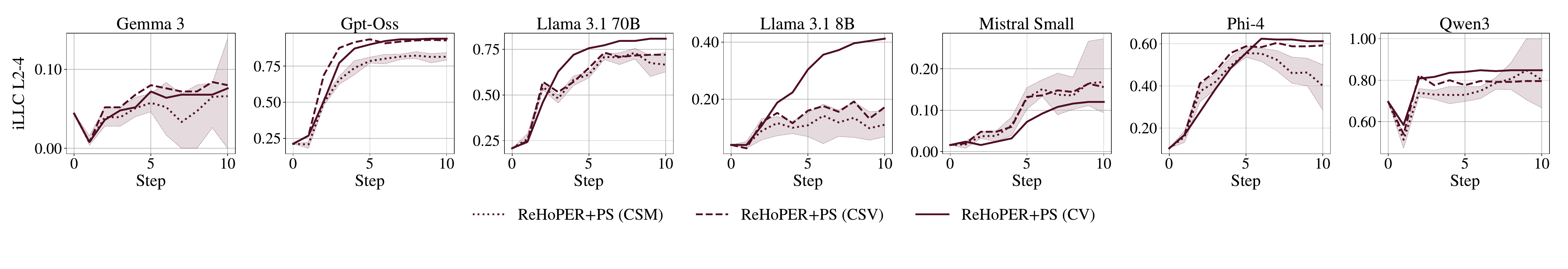}
    \includegraphics[width=1.\linewidth, trim=0 70 0 0, clip]{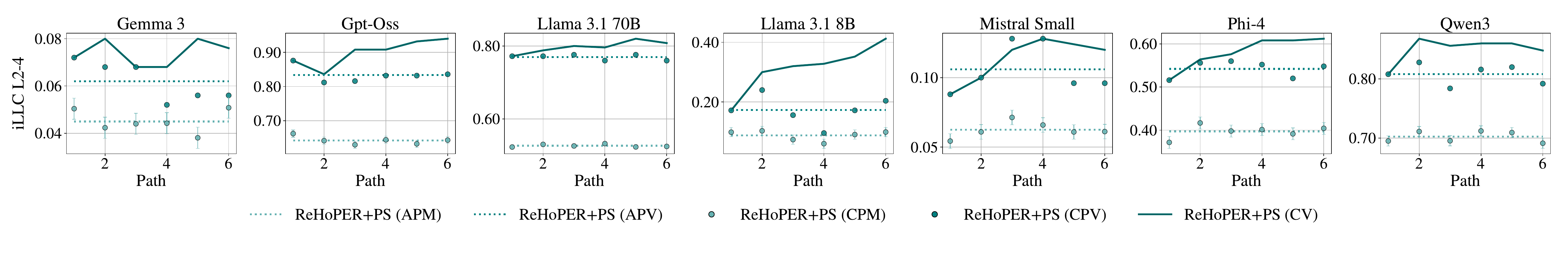}
    \caption{
        Answer aggregation for ReHoPER+PS on iLLC L2-4 across models. Top: accuracy over reasoning steps; bottom: accuracy over stochastic paths. CV is cumulative voting, CSV voting over each path’s latest prediction, and CSM their mean correctness. CPV/CPM are voting/mean correctness within a path; APV/APM are their averages across paths.
    }
    \vspace{-1em}
    \label{fig:agg-mini}
\end{figure}

Figure~\ref{fig:agg-mini} separates improvements in predictions produced by ReHoPER+PS during reasoning from the gains due to aggregating them on iLLC L2-4. In the upper row, both the mean correctness of the latest prediction from each path and the vote over those latest predictions tend to rise as more intermediate questions are answered. Later reasoning states are therefore often better in their own right. The cumulative vote, which retains predictions from earlier steps, typically rises further and finishes above the latest-prediction vote; this suggests that earlier states still contribute useful votes even when the current predictions have improved. In the lower row, the vote within an individual path varies across trajectories, while voting over all paths generally improves as paths are added and tends to finish above the path-local summaries. Thus, the underlying upward trend is strengthened by retaining and combining predictions across steps and paths. Appendices~\ref{app:step-aggregation} and~\ref{app:path-aggregation} define the measures and report the full results across models, output prefixes, and more datasets.

\input{tables/ltm}

\subsection{Paths vs.\ steps}

Figure~\ref{fig:pvs} compares how ReHoPER+PS and SC+PS improve as more candidate predictions are included in the vote. For ReHoPER, the dashed curves add reasoning steps across six paths, while the dotted curves add paths with a fixed maximum depth of ten steps; the solid curve adds independent SC samples. This comparison shows how the choice between deeper reasoning and more paths affects accuracy as the candidate budget grows. Appendix~\ref{app:pvs} gives the exact vote construction and results for other models and output prefixes.
ReHoPER outperforms SC in most configurations, often from small iteration budgets onward. Across datasets and models, both ReHoPER and SC exhibit diminishing returns as gains per additional step or path become small beyond a certain point. For ReHoPER, one contributing factor is that some paths terminate early, so later steps contribute fewer new candidates.
The figure also shows that the relative value of steps versus paths is task dependent. On iLLC, the dotted curves (more paths at fixed depth) usually rise faster than the dashed curves in the first few iterations, indicating that the earliest paths provide the largest marginal benefit; however, they plateau quickly while the dashed curves continue to improve, suggesting that deeper reasoning steps are ultimately more beneficial on this dataset. In contrast, for example, in MuSR with Phi-4, configurations that emphasize more paths can be advantageous, and the upward trend of the SC curve (figuratively, many step-0 paths)
is consistent with this behavior.
\subsection{Comparison with Least-to-Most and Self-Refine}

Table~\ref{tab:ltm} compares ReHoPER+PS with Least-to-Most (LtM)~\citep{ZhouSHWS0SCBLC23}, its few-shot CoT baseline, and Self-Refine (SR)~\citep{madaan2023selfrefine} on iLLC. For L1, we use the original CoT and LtM prompts in the LtM paper~\citep{ZhouSHWS0SCBLC23}; for L2, we construct analogous prompts following the same template to isolate differences in reasoning rather than prompt design. In SR, feedback and refinement are implemented as few-shot prompts in the original work, but no exemplars exist for our datasets and creating task-specific ones would defeat the purpose of our zero-shot setting. We therefore use instruction-only feedback and refinement prompts, perform five refinement iterations, and take the final refined answer. SR improves over LtM in several cases, but does not match ReHoPER. Our method performs best overall and is particularly stronger on the more compositional L2 variants, where LtM, CoT, and SR degrade sharply. This indicates that while post-hoc refinement helps, explicit planning with intermediate questions is more effective for compositional reasoning even in a task-agnostic, zero-shot setting.

\input{tables/time}

\subsection{Compute–accuracy tradeoffs}

Table~\ref{tab:time} reports runtime and accuracy on iLLC, with accuracy averaged over IO/CoT/PS. Here, we denote the original SC setup used in the main experiments as SC$_{61}$, and include SC$_{181}$, which increases the sampling budget from 61 to 181 samples to better match ReHoPER's runtime and number of forward passes. For efficiency, we batch the three prefix-specific conversations in a single inference request for all methods; in ReHoPER, planning and intermediate questions are shared across prefixes, and prefixes are applied only at final prediction. Increasing the sampling budget yields limited gains: consistent with Figure~\ref{fig:pvs}, SC$_{181}$ is only slightly better and sometimes worse than SC$_{61}$, indicating that sampling more final answers is not the main limitation of Self-Consistency. In contrast, ReHoPER achieves larger improvements by allocating compute to planning and intermediate reasoning, with gains evident on L1 and substantially larger on L2. SC$_{181}$ also incurs higher runtime, often exceeding ReHoPER while rarely matching its accuracy. Overall, ReHoPER's advantage comes from structured computation via receding-horizon planning, not simply more forward passes.

\section{Conclusion and future work}

In this work, we introduced ReHoPER, a task-agnostic, inference-only method that reasons through intermediate questions across multiple paths and aggregates their predictions. It improves over Self-Consistency and other strong baselines under the same answer-format prefix in most settings, with the largest gains on harder compositional tasks. Our controlled iLLC benchmark helps expose how performance changes with compositional difficulty, while planning ablations examine design choices within the method. An interesting direction of future work will be to make the compute budget more adaptive (e.g., dynamically increasing the number of paths when some terminate early). Our code and iLLC generator are publicly available on GitHub.

\subsection*{AI use statement}

We used generative AI tools to help look for additional related work in the
literature, polish the writing of this paper, and assist with parts of the
implementation and debugging. The authors reviewed all
AI-assisted changes and take responsibility for the final manuscript, code,
and results.

\subsection*{Reproducibility statement}

Section~\ref{sec:rehoper} specifies the ReHoPER planning, rollout, termination,
and answer-aggregation procedure. Section~\ref{sec:setup} describes the
datasets, baselines, output prefixes, and shared inference settings. The
appendix provides the full intermediate-question instructions
(Appendix~\ref{app:instruct}), dataset selection and answer post-processing
(Appendix~\ref{app:data-examples}), exact model checkpoints and deployment
details (Appendix~\ref{app:models}), and the hyperparameter study
(Appendix~\ref{app:hyperparams}). Repeated-run results with means and standard
deviations are reported in Appendix~\ref{app:robustness}. The ReHoPER implementation and iLLC generator are published at the GitHub repositories linked in the abstract.

\bibliographystyle{unsrt}
\bibliography{custom}

\clearpage

\appendix

\titlecontents{section}
  [2em]              {}                 {\contentslabel{2em}}   {}                 {\titlerule*[0.5pc]{.}\contentspage\vspace{1em}} \startcontents[appendix]
\section*{Appendix contents}
\vspace{0.5em}
\printcontents[appendix]{}{1}{}

\include{appendix}

\end{document}

%% file: figures/the-figure.tex
\usetikzlibrary{arrows.meta,backgrounds,calc,positioning}

\definecolor{rhOverviewMuted}{HTML}{64748B}
\definecolor{rhOverviewBlue}{HTML}{EFF6FF}
\definecolor{rhOverviewBlueLine}{HTML}{2563EB}
\definecolor{rhOverviewTeal}{HTML}{F0FDFA}
\definecolor{rhOverviewTealLine}{HTML}{0F766E}
\definecolor{rhOverviewAmber}{HTML}{FFFBEB}
\definecolor{rhOverviewAmberLine}{HTML}{D97706}
\definecolor{rhOverviewPurple}{HTML}{F5F3FF}
\definecolor{rhOverviewPurpleLine}{HTML}{7C3AED}
\definecolor{rhOverviewRose}{HTML}{FFF1F2}
\definecolor{rhOverviewRoseLine}{HTML}{E11D48}
\definecolor{rhOverviewPathLine}{HTML}{94A3B8}
\definecolor{rhOverviewPathBack}{HTML}{E2E8F0}
\definecolor{rhOverviewPathMiddle}{HTML}{F1F5F9}
\definecolor{rhOverviewPathFront}{HTML}{FAFCFE}

\tikzset{
  rhOverviewBox/.style={
    rectangle, rounded corners=4pt, line width=0.6pt,
    text width=2.55cm, minimum height=0.76cm,
    inner sep=3pt, align=center, font=\footnotesize
  },
  rhOverviewPredictionBox/.style={
    rhOverviewBox, text width=3.70cm, minimum height=0.76cm,
    fill=rhOverviewPurple, draw=rhOverviewPurpleLine,
    text=rhOverviewPurpleLine
  },
  rhOverviewOutputBox/.style={
    rhOverviewBox, text width=3.70cm, minimum height=0.76cm,
    fill=rhOverviewRose, draw=rhOverviewRoseLine,
    text=rhOverviewRoseLine
  },
  rhOverviewArrow/.style={
    -{Stealth[length=5pt,width=4pt]},
    line width=0.8pt, draw=rhOverviewMuted
  }
}

\begin{figure}[!ht]
  \centering
  \resizebox{0.98\linewidth}{!}{    \begin{tikzpicture}[node distance=0.34cm]
      \node[rhOverviewBox, fill=rhOverviewBlue, draw=rhOverviewBlueLine,
        text=rhOverviewBlueLine]
        (input) {\textbf{Task input}\\[-2pt]{\scriptsize Question + context}};
      \node[rhOverviewBox, fill=rhOverviewAmber, draw=rhOverviewAmberLine,
        text=rhOverviewAmberLine, text width=4.10cm, right=of input]
        (plan) {\textbf{Plan \& look ahead}\\[-1pt]{\scriptsize Generate intermediate questions\\[-1pt]using the current history}};
      \node[rhOverviewBox, fill=rhOverviewTeal, draw=rhOverviewTealLine,
        text=rhOverviewTealLine, text width=3.40cm, right=of plan]
        (answer) {\textbf{Answer one}\\[-2pt]{\scriptsize First unanswered question}};
      \node[rhOverviewBox, fill=rhOverviewBlue, draw=rhOverviewBlueLine,
        text=rhOverviewBlueLine, text width=3.70cm, right=of answer]
        (update) {\textbf{Update history}\\[-2pt]{\scriptsize Store new intermediate Q\&A}};

      \draw[rhOverviewArrow] (input) -- (plan);
      \draw[rhOverviewArrow] (plan) -- (answer);
      \draw[rhOverviewArrow] (answer) -- (update);
      \coordinate (loopRight) at ($(update.north)+(0,0.45cm)$);
      \coordinate (loopLeft) at (loopRight -| plan.north);
      \draw[rhOverviewArrow, rounded corners=5pt]
        (update.north) -- (loopRight) -- (loopLeft) -- (plan.north);
      \node[font=\scriptsize\bfseries, text=rhOverviewBlueLine,
        fill=white, inner sep=2pt]
        at ($(loopLeft)!0.5!(loopRight)$)
        {Replan from updated history};

      \node[rhOverviewPredictionBox]
        (initial) at ($(input.south)+(0,-1.00cm)$)
        {\textbf{Initial prediction}\\[-2pt]{\scriptsize Before intermediate questions}};
      \node[rhOverviewPredictionBox]
        (intermediate) at (initial -| update)
        {\textbf{Subsequent predictions}\\[-2pt]{\scriptsize After each answered question}};
      \node[rhOverviewOutputBox]
        (aggregate) at ($(initial)!0.5!(intermediate)$)
        {\textbf{Aggregator}\\[-2pt]{\scriptsize Vote across paths and steps}};
      \node[rhOverviewOutputBox,
        text width=3.20cm, minimum height=0.58cm,
        below=0.30cm of aggregate]
        (final) {\textbf{Final prediction}};

      \draw[rhOverviewArrow] (input.south) -- (initial.north);
      \draw[rhOverviewArrow] (update.south) -- (intermediate.north);
      \draw[rhOverviewArrow] (initial.east) -- (aggregate.west);
      \draw[rhOverviewArrow]
        ($(intermediate.west)+(0,0.17cm)$) --
        ($(aggregate.east)+(0,0.17cm)$);
      \draw[rhOverviewArrow] (aggregate.south) -- (final.north);

      \coordinate (pathTopY) at ($(loopRight)+(0,0.34cm)$);
      \coordinate (pathShoulderY) at ($(plan.south)+(0,-0.12cm)$);
      \coordinate (pathBottomY) at ($(intermediate.south)+(0,-0.13cm)$);
      \coordinate (pathLeftX) at ($(plan.west)+(-0.14cm,0)$);
      \coordinate (pathElbowX) at ($(intermediate.west)+(-0.14cm,0)$);
      \coordinate (pathRightX) at ($(intermediate.east)+(0.14cm,0)$);
      \coordinate (pathNW) at (pathLeftX |- pathTopY);
      \coordinate (pathNE) at (pathRightX |- pathTopY);
      \coordinate (pathSE) at (pathRightX |- pathBottomY);
      \coordinate (pathSW) at (pathElbowX |- pathBottomY);
      \coordinate (pathElbow) at (pathElbowX |- pathShoulderY);
      \coordinate (pathShoulderW) at (pathLeftX |- pathShoulderY);
      \begin{pgfonlayer}{background}
        \path[rounded corners=5pt, line width=0.5pt,
          fill=rhOverviewPathBack, draw=rhOverviewPathLine,
          transform canvas={xshift=0.20cm,yshift=-0.20cm}]
          (pathNW) -- (pathNE) -- (pathSE) -- (pathSW) --
          (pathElbow) -- (pathShoulderW) -- cycle;
        \draw[rhOverviewArrow, opacity=0.55]
          ($(intermediate.west)+(0.20cm,-0.17cm)$) --
          ($(aggregate.east)+(0,-0.17cm)$);
        \path[rounded corners=5pt, line width=0.5pt,
          fill=rhOverviewPathMiddle, draw=rhOverviewPathLine,
          transform canvas={xshift=0.10cm,yshift=-0.10cm}]
          (pathNW) -- (pathNE) -- (pathSE) -- (pathSW) --
          (pathElbow) -- (pathShoulderW) -- cycle;
        \draw[rhOverviewArrow, opacity=0.75]
          ($(intermediate.west)+(0.10cm,0)$) -- (aggregate.east);
        \path[rounded corners=5pt, line width=0.6pt,
          fill=rhOverviewPathFront, draw=rhOverviewPathLine]
          (pathNW) -- (pathNE) -- (pathSE) -- (pathSW) --
          (pathElbow) -- (pathShoulderW) -- cycle;
      \end{pgfonlayer}
      \node[anchor=north west, font=\scriptsize\bfseries,
        text=rhOverviewMuted, inner sep=0pt]
        at ($(pathNW)+(0.14cm,-0.06cm)$) {Path $i$ of $n$};
    \end{tikzpicture}  }
  \caption{ReHoPER overview. Layered frames indicate independent paths. Each path plans questions, answers one, and replans from updated history. A majority vote over the initial and per-step predictions across paths yields the final prediction.}
  \label{fig:rehoper_overview}
  \vspace{-1em}
\end{figure}

%% file: tables/result.tex
\begin{table*}[t]
\centering
\Large
\caption{
Accuracy (\%) of ReHoPER (RH) on MuSR, MoreHopQA, and iLLC benchmarks under IO, CoT, and PS prefixes. SD refers to the Self-Discover baseline. SC denotes simple question answering with Self-Consistency using the corresponding output prefix, while RH applies our planning
approach with that prefix.
\colorbox{green!40}{\small Green}, \colorbox{red!25}{\small red}, and \colorbox{gray!25}{\small gray} indicate
performance changes of ReHoPER relative to SC. The bottom-right block reports averages over the eight displayed subsets for each model and prompt.
}
\label{tab:result}
\resizebox{1.\textwidth}{!}{
\begin{tabular}{l p{0pt} c c cccccc p{5pt} c c cccccc p{5pt} c c cccccc}

\toprule

& &
\multicolumn{8}{c}{\textbf{MuSR MM}} & &
\multicolumn{8}{c}{\textbf{MuSR OP}} & &
\multicolumn{8}{c}{\textbf{MuSR TA}} \\

\cmidrule(lr){3-10}
\cmidrule(lr){12-19}
\cmidrule(lr){21-28}

& &
& & \multicolumn{2}{c}{\textbf{IO}} & \multicolumn{2}{c}{\textbf{CoT}} & \multicolumn{2}{c}{\textbf{PS}} & &
& & \multicolumn{2}{c}{\textbf{IO}} & \multicolumn{2}{c}{\textbf{CoT}} & \multicolumn{2}{c}{\textbf{PS}} & &
& & \multicolumn{2}{c}{\textbf{IO}} & \multicolumn{2}{c}{\textbf{CoT}} & \multicolumn{2}{c}{\textbf{PS}} \\

\cmidrule(lr){5-6} \cmidrule(lr){7-8} \cmidrule(lr){9-10}
\cmidrule(lr){14-15} \cmidrule(lr){16-17} \cmidrule(lr){18-19}
\cmidrule(lr){23-24} \cmidrule(lr){25-26} \cmidrule(lr){27-28}

\textbf{Model} & &
\textbf{SD} & &
\textbf{SC} & \textbf{RH} &
\textbf{SC} & \textbf{RH} &
\textbf{SC} & \textbf{RH} & &
\textbf{SD} & &
\textbf{SC} & \textbf{RH} &
\textbf{SC} & \textbf{RH} &
\textbf{SC} & \textbf{RH} & &
\textbf{SD} & &
\textbf{SC} & \textbf{RH} &
\textbf{SC} & \textbf{RH} &
\textbf{SC} & \textbf{RH} \\

\midrule

Gemma 3 & &
57.6 & &
54.4 & \gain{61.2} &
61.2 & \gain{62.0} &
61.6 & \gain{65.6} & &
50.0 & &
50.4 & \gain{55.1} &
51.6 & \gain{55.5} &
54.7 & \gain{55.9} & &
50.0 & &
58.8 & \loss{47.6} &
56.0 & \loss{51.2} &
58.0 & \loss{56.0} \\

gpt-oss & &
74.8 & &
70.8 & \gain{76.0} &
70.4 & \gain{76.4} &
73.2 & \gain{76.8} & &
53.9 & &
55.5 & \loss{53.9} &
53.1 & \gain{53.5} &
57.0 & \loss{53.5} & &
70.8 & &
69.2 & \loss{68.0} &
56.4 & \gain{63.2} &
65.2 & \gain{68.0} \\

Llama 3.1 8B & &
54.0 & &
56.8 & \gain{62.4} &
56.4 & \gain{60.8} &
57.2 & \gain{63.6} & &
44.5 & &
49.2 & \gain{53.1} &
52.0 & \loss{48.4} &
52.0 & \loss{48.4} & &
34.8 & &
35.2 & \gain{43.6} &
43.2 & \gain{47.6} &
45.2 & \gain{47.6} \\

Llama 3.1 70B & &
64.8 & &
65.6 & \gain{71.2} &
66.0 & \gain{70.4} &
66.8 & \gain{71.2} & &
42.6 & &
46.1 & \gain{55.9} &
46.1 & \gain{54.3} &
48.4 & \gain{55.9} & &
57.6 & &
62.0 & \loss{57.6} &
62.0 & \loss{59.6} &
60.4 & \loss{59.2} \\

Mistral Small & &
58.8 & &
60.4 & \gain{73.2} &
68.4 & \gain{78.0} &
70.8 & \gain{79.6} & &
48.8 & &
50.4 & \loss{47.3} &
46.5 & \gain{50.0} &
47.7 & \gain{50.0} & &
51.6 & &
48.4 & \gain{58.8} &
51.6 & \gain{62.0} &
55.6 & \gain{58.8} \\

Phi-4 & &
72.0 & &
75.6 & \loss{74.8} &
74.8 & \gain{76.0} &
78.8 & \loss{75.2} & &
44.1 & &
52.3 & \loss{50.0} &
48.0 & \gain{50.0} &
47.7 & \gain{52.3} & &
56.0 & &
56.0 & \gain{56.4} &
55.6 & \gain{61.6} &
56.8 & \gain{62.4} \\

Qwen3 w/o Thinking & &
60.4 & &
58.4 & \gain{72.0} &
67.2 & \gain{71.6} &
70.8 & \gain{72.8} & &
40.2 & &
53.1 & \loss{51.6} &
50.8 & \gain{54.7} &
50.8 & \gain{51.6} & &
59.2 & &
57.2 & \gain{64.0} &
71.2 & \loss{68.8} &
68.0 & \gain{68.8} \\

Qwen3 w/ Thinking & &
70.4 & &
72.0 & \gain{72.4} &
67.2 & \gain{72.0} &
70.8 & \gain{73.6} & &
49.2 & &
52.0 & \gain{57.4} &
50.8 & \gain{53.5} &
50.8 & \gain{53.5} & &
70.0 & &
71.6 & \loss{66.4} &
71.2 & \loss{70.0} &
68.0 & \loss{67.2} \\

\cdashlinelr{1-28}

Average & &
64.1 & &
64.3 & \gain{70.4} &
66.5 & \gain{70.9} &
68.8 & \gain{72.3} & &
46.7 & &
51.1 & \gain{53.0} &
49.9 & \gain{52.5} &
51.1 & \gain{52.6} & &
56.3 & &
57.3 & \gain{57.8} &
58.4 & \gain{60.5} &
59.7 & \gain{61.0} \\

\midrule

& &
\multicolumn{8}{c}{\textbf{iLLC L1-4}} & &
\multicolumn{8}{c}{\textbf{iLLC L1-6}} & &
\multicolumn{8}{c}{\textbf{MoreHopQA}} \\

\cmidrule(lr){3-10}
\cmidrule(lr){12-19}
\cmidrule(lr){21-28}

& &
& & \multicolumn{2}{c}{\textbf{IO}} & \multicolumn{2}{c}{\textbf{CoT}} & \multicolumn{2}{c}{\textbf{PS}} & &
& & \multicolumn{2}{c}{\textbf{IO}} & \multicolumn{2}{c}{\textbf{CoT}} & \multicolumn{2}{c}{\textbf{PS}} & &
& & \multicolumn{2}{c}{\textbf{IO}} & \multicolumn{2}{c}{\textbf{CoT}} & \multicolumn{2}{c}{\textbf{PS}} \\

\cmidrule(lr){5-6} \cmidrule(lr){7-8} \cmidrule(lr){9-10}
\cmidrule(lr){14-15} \cmidrule(lr){16-17} \cmidrule(lr){18-19}
\cmidrule(lr){23-24} \cmidrule(lr){25-26} \cmidrule(lr){27-28}

\textbf{Model} & &
\textbf{SD} & &
\textbf{SC} & \textbf{RH} &
\textbf{SC} & \textbf{RH} &
\textbf{SC} & \textbf{RH} & &
\textbf{SD} & &
\textbf{SC} & \textbf{RH} &
\textbf{SC} & \textbf{RH} &
\textbf{SC} & \textbf{RH} & &
\textbf{SD} & &
\textbf{SC} & \textbf{RH} &
\textbf{SC} & \textbf{RH} &
\textbf{SC} & \textbf{RH} \\

\midrule

Gemma 3 & &
73.6 & &
20.8 & \gain{46.0} &
82.8 & \gain{83.2} &
90.4 & \loss{88.0} & &
46.0 & &
10.0 & \gain{34.4} &
56.0 & \gain{62.8} &
74.0 & \loss{66.0} & &
44.0 & &
61.3 & \gain{64.7} &
64.7 & \gain{68.7} &
66.7 & \gain{74.0} \\

gpt-oss & &
78.8 & &
90.0 & \loss{87.6} &
95.6 & \loss{88.4} &
96.0 & \loss{88.0} & &
66.8 & &
83.2 & \loss{79.6} &
85.2 & \loss{78.8} &
82.4 & \loss{80.0} & &
39.3 & &
80.7 & \loss{78.7} &
81.3 & \loss{80.0} &
79.3 & \gain{81.3} \\

Llama 3.1 8B & &
46.8 & &
74.8 & \loss{70.0} &
67.6 & \same{67.6} &
80.0 & \loss{77.6} & &
18.8 & &
50.0 & \loss{40.8} &
53.2 & \loss{52.8} &
56.0 & \loss{52.8} & &
34.0 & &
59.3 & \loss{57.3} &
61.3 & \loss{60.7} &
64.0 & \loss{63.3} \\

Llama 3.1 70B & &
93.2 & &
89.6 & \loss{89.2} &
88.8 & \gain{91.2} &
98.0 & \loss{97.6} & &
77.6 & &
73.2 & \gain{77.2} &
71.6 & \gain{76.8} &
83.2 & \gain{86.4} & &
62.0 & &
68.0 & \gain{71.3} &
68.0 & \gain{71.3} &
68.7 & \gain{74.0} \\

Mistral Small & &
70.4 & &
88.4 & \gain{90.4} &
92.0 & \gain{92.4} &
91.6 & \loss{91.2} & &
42.8 & &
73.2 & \loss{69.6} &
76.8 & \loss{72.0} &
78.4 & \gain{78.8} & &
54.7 & &
70.0 & \gain{71.3} &
70.0 & \same{70.0} &
71.3 & \gain{72.7} \\

Phi-4 & &
88.0 & &
89.6 & \gain{91.2} &
92.8 & \same{92.8} &
91.2 & \gain{92.0} & &
68.8 & &
68.8 & \gain{72.4} &
76.4 & \gain{80.4} &
74.4 & \gain{76.0} & &
64.0 & &
77.3 & \same{77.3} &
76.0 & \gain{80.7} &
77.3 & \loss{76.7} \\

Qwen3 w/o Thinking & &
71.6 & &
82.4 & \gain{85.2} &
79.2 & \gain{84.4} &
83.6 & \gain{86.0} & &
60.4 & &
62.4 & \gain{80.0} &
70.0 & \gain{78.8} &
75.6 & \gain{80.0} & &
34.0 & &
70.0 & \loss{68.0} &
69.3 & \gain{70.7} &
73.3 & \loss{72.0} \\

Qwen3 w/ Thinking & &
1.2 & &
97.2 & \same{97.2} &
79.2 & \gain{95.6} &
83.6 & \gain{95.6} & &
0.8 & &
87.2 & \gain{90.0} &
70.0 & \gain{86.8} &
75.6 & \gain{85.6} & &
42.0 & &
75.3 & \gain{80.0} &
69.3 & \gain{78.7} &
73.3 & \gain{78.0} \\

\cdashlinelr{1-28}

Average & &
65.5 & &
79.1 & \gain{82.1} &
84.8 & \gain{87.0} &
89.3 & \gain{89.5} & &
47.8 & &
63.5 & \gain{68.0} &
69.9 & \gain{73.7} &
75.0 & \gain{75.7} & &
46.8 & &
70.3 & \gain{71.1} &
70.0 & \gain{72.6} &
71.8 & \gain{74.0} \\

\midrule

& &
\multicolumn{8}{c}{\textbf{iLLC L2-4}} & &
\multicolumn{8}{c}{\textbf{iLLC L2-6}} & &
\multicolumn{8}{c}{\textbf{Average}} \\

\cmidrule(lr){3-10}
\cmidrule(lr){12-19}
\cmidrule(lr){21-28}

& &
& & \multicolumn{2}{c}{\textbf{IO}} & \multicolumn{2}{c}{\textbf{CoT}} & \multicolumn{2}{c}{\textbf{PS}} & &
& & \multicolumn{2}{c}{\textbf{IO}} & \multicolumn{2}{c}{\textbf{CoT}} & \multicolumn{2}{c}{\textbf{PS}} & &
& & \multicolumn{2}{c}{\textbf{IO}} & \multicolumn{2}{c}{\textbf{CoT}} & \multicolumn{2}{c}{\textbf{PS}} \\

\cmidrule(lr){5-6} \cmidrule(lr){7-8} \cmidrule(lr){9-10}
\cmidrule(lr){14-15} \cmidrule(lr){16-17} \cmidrule(lr){18-19}
\cmidrule(lr){23-24} \cmidrule(lr){25-26} \cmidrule(lr){27-28}

\textbf{Model} & &
\textbf{SD} & &
\textbf{SC} & \textbf{RH} &
\textbf{SC} & \textbf{RH} &
\textbf{SC} & \textbf{RH} & &
\textbf{SD} & &
\textbf{SC} & \textbf{RH} &
\textbf{SC} & \textbf{RH} &
\textbf{SC} & \textbf{RH} & &
\textbf{SD} & &
\textbf{SC} & \textbf{RH} &
\textbf{SC} & \textbf{RH} &
\textbf{SC} & \textbf{RH} \\

\midrule

Gemma 3 & &
1.6 & &
7.2 & \loss{6.0} &
7.2 & \loss{6.4} &
3.6 & \gain{7.6} & &
0.4 & &
0.8 & \same{0.8} &
1.2 & \gain{1.6} &
2.0 & \loss{0.8} & &
40.4 & &
33.0 & \gain{39.5} &
47.6 & \gain{48.9} &
51.4 & \gain{51.7} \\

gpt-oss & &
67.2 & &
95.2 & \loss{93.6} &
81.6 & \gain{93.2} &
40.0 & \gain{94.0} & &
56.0 & &
86.4 & \loss{86.0} &
67.2 & \gain{84.4} &
14.0 & \gain{84.0} & &
63.5 & &
78.9 & \loss{77.9} &
73.9 & \gain{77.2} &
63.4 & \gain{78.2} \\

Llama 3.1 8B & &
0.0 & &
9.6 & \gain{12.8} &
9.6 & \gain{18.4} &
28.8 & \gain{41.2} & &
0.0 & &
3.6 & \gain{8.0} &
4.8 & \gain{8.0} &
5.6 & \gain{14.4} & &
29.1 & &
42.3 & \gain{43.5} &
43.5 & \gain{45.5} &
48.6 & \gain{51.1} \\

Llama 3.1 70B & &
8.0 & &
13.6 & \gain{78.4} &
36.4 & \gain{78.8} &
47.6 & \gain{80.8} & &
3.6 & &
4.8 & \gain{68.0} &
17.6 & \gain{63.2} &
29.2 & \gain{74.4} & &
51.2 & &
52.9 & \gain{71.1} &
57.1 & \gain{70.7} &
62.8 & \gain{74.9} \\

Mistral Small & &
0.4 & &
4.0 & \gain{11.2} &
2.0 & \gain{11.6} &
4.4 & \gain{12.0} & &
0.0 & &
0.0 & \gain{1.6} &
0.0 & \gain{2.4} &
0.0 & \gain{2.0} & &
40.9 & &
49.3 & \gain{52.9} &
50.9 & \gain{54.8} &
52.5 & \gain{55.6} \\

Phi-4 & &
14.0 & &
12.0 & \gain{54.0} &
10.8 & \gain{62.8} &
10.4 & \gain{61.2} & &
6.0 & &
4.4 & \gain{27.2} &
3.6 & \gain{38.0} &
6.4 & \gain{41.6} & &
51.6 & &
54.5 & \gain{62.9} &
54.8 & \gain{67.8} &
55.4 & \gain{67.2} \\

Qwen3 w/o Thinking & &
1.2 & &
72.4 & \gain{81.6} &
54.0 & \gain{84.4} &
86.8 & \loss{84.8} & &
0.4 & &
36.0 & \gain{59.6} &
23.2 & \gain{62.0} &
72.8 & \loss{68.0} & &
40.9 & &
61.5 & \gain{70.2} &
60.6 & \gain{71.9} &
72.7 & \gain{73.0} \\

Qwen3 w/ Thinking & &
1.2 & &
92.4 & \gain{93.6} &
54.0 & \gain{91.6} &
86.8 & \gain{92.4} & &
1.6 & &
85.2 & \loss{84.4} &
23.2 & \gain{84.4} &
72.8 & \gain{84.8} & &
29.6 & &
79.1 & \gain{80.2} &
60.6 & \gain{79.1} &
72.7 & \gain{78.8} \\

\cdashlinelr{1-28}

Average & &
11.7 & &
38.3 & \gain{53.9} &
32.0 & \gain{55.9} &
38.6 & \gain{59.3} & &
8.5 & &
27.7 & \gain{42.0} &
17.6 & \gain{43.0} &
25.4 & \gain{46.3} & &
43.4 & &
56.4 & \gain{62.3} &
56.1 & \gain{64.5} &
59.9 & \gain{66.3} \\

\bottomrule

\end{tabular}
}
\vspace{-1em}
\end{table*}

%% file: tables/vs.tex
\begin{table*}[t]
\centering
\large
\caption{
Planning ablations without replanning (-RP), without look-ahead (-LA), and the full ReHoPER method (RH) across three prompting styles, IO, CoT, and PS, on MuSR MM, MoreHopQA, and iLLC~L2-4. Values are accuracies (\%), and \colorbox{green!40}{\small green} and \colorbox{green!15}{\small light green} mark the best and second best variant, respectively, for each (model, dataset, prefix) triplet. The bottom-right block reports averages over all datasets for each model and prompt.
}
\label{tab:planning-ablations-full}
\resizebox{.91\textwidth}{!}{
\begin{tabular}{
l p{0pt}
ccc p{0pt} ccc p{0pt} ccc p{0pt}
ccc p{0pt} ccc p{0pt} ccc
}

\toprule

& &
\multicolumn{11}{c}{\textbf{MuSR MM}} & &
\multicolumn{11}{c}{\textbf{MoreHopQA}} \\

\cmidrule(lr){3-13}
\cmidrule(lr){15-25}

& &
\multicolumn{3}{c}{\textbf{IO}} & & \multicolumn{3}{c}{\textbf{CoT}} & & \multicolumn{3}{c}{\textbf{PS}} & &
\multicolumn{3}{c}{\textbf{IO}} & & \multicolumn{3}{c}{\textbf{CoT}} & & \multicolumn{3}{c}{\textbf{PS}} \\

\cmidrule(lr){3-5} \cmidrule(lr){7-9} \cmidrule(lr){11-13}
\cmidrule(lr){15-17} \cmidrule(lr){19-21} \cmidrule(lr){23-25}

\textbf{Model} & &
\textbf{-RP} & \textbf{-LA} & \textbf{RH} & &
\textbf{-RP} & \textbf{-LA} & \textbf{RH} & &
\textbf{-RP} & \textbf{-LA} & \textbf{RH} & &
\textbf{-RP} & \textbf{-LA} & \textbf{RH} & &
\textbf{-RP} & \textbf{-LA} & \textbf{RH} & &
\textbf{-RP} & \textbf{-LA} & \textbf{RH} \\
\midrule

Gemma 3 & &
\none{57.6} & \smallgain{58.8} & \gain{61.2} & &
\smallgain{59.2} & \none{58.4} & \gain{62.0} & &
\none{62.4} & \none{62.4} & \gain{65.6} & &
\none{64.0} & \gain{66.0} & \smallgain{64.7} & &
\none{66.7} & \none{66.7} & \gain{68.7} & &
\smallgain{70.0} & \none{69.3} & \gain{74.0} \\

gpt-oss & &
\smallgain{72.4} & \none{70.0} & \gain{76.0} & &
\smallgain{72.0} & \none{68.0} & \gain{76.4} & &
\smallgain{74.0} & \none{71.2} & \gain{76.8} & &
\gain{80.7} & \none{78.7} & \none{78.7} & &
\none{79.3} & \none{79.3} & \gain{80.0} & &
\smallgain{80.7} & \none{78.7} & \gain{81.3} \\

Llama 3.1 8B & &
\smallgain{61.6} & \none{58.4} & \gain{62.4} & &
\gain{61.6} & \smallgain{61.2} & \none{60.8} & &
\gain{64.8} & \none{60.4} & \smallgain{63.6} & &
\none{52.0} & \gain{57.3} & \gain{57.3} & &
\none{56.7} & \smallgain{58.7} & \gain{60.7} & &
\none{60.7} & \gain{66.0} & \smallgain{63.3} \\

Llama 3.1 70B & &
\smallgain{70.4} & \none{70.0} & \gain{71.2} & &
\smallgain{72.0} & \gain{72.4} & \none{70.4} & &
\gain{71.6} & \none{71.2} & \none{71.2} & &
\none{68.7} & \gain{74.0} & \smallgain{71.3} & &
\none{70.7} & \gain{72.7} & \smallgain{71.3} & &
\none{72.7} & \gain{74.0} & \gain{74.0} \\

Mistral Small & &
\none{70.0} & \smallgain{70.4} & \gain{73.2} & &
\none{74.8} & \smallgain{77.2} & \gain{78.0} & &
\none{77.2} & \gain{79.6} & \gain{79.6} & &
\none{66.0} & \smallgain{68.7} & \gain{71.3} & &
\none{68.0} & \none{68.0} & \gain{70.0} & &
\none{71.3} & \gain{72.7} & \gain{72.7} \\

Phi-4 & &
\none{74.4} & \gain{75.2} & \smallgain{74.8} & &
\smallgain{77.2} & \gain{77.6} & \none{76.0} & &
\smallgain{78.0} & \gain{79.6} & \none{75.2} & &
\smallgain{76.7} & \none{74.7} & \gain{77.3} & &
\smallgain{78.7} & \none{76.7} & \gain{80.7} & &
\gain{77.3} & \none{76.0} & \smallgain{76.7} \\

Qwen3 & &
\smallgain{74.0} & \gain{76.8} & \none{72.0} & &
\smallgain{72.8} & \gain{78.0} & \none{71.6} & &
\smallgain{75.2} & \gain{77.2} & \none{72.8} & &
\smallgain{66.7} & \none{65.3} & \gain{68.0} & &
\smallgain{69.3} & \none{68.0} & \gain{70.7} & &
\none{69.3} & \smallgain{70.7} & \gain{72.0} \\

\cdashlinelr{1-25}

Average & &
\smallgain{68.6} & \none{68.5} & \gain{70.1} & &
\none{69.9} & \smallgain{70.4} & \gain{70.7} & &
\smallgain{71.9} & \none{71.7} & \gain{72.1} & &
\none{67.8} & \smallgain{69.2} & \gain{69.8} & &
\none{69.9} & \smallgain{70.0} & \gain{71.7} & &
\none{71.7} & \smallgain{72.5} & \gain{73.4} \\

\midrule

& &
\multicolumn{11}{c}{\textbf{iLLC~L2-4}} & &
\multicolumn{11}{c}{\textbf{Average}} \\

\cmidrule(lr){3-13}
\cmidrule(lr){15-25}

& &
\multicolumn{3}{c}{\textbf{IO}} & & \multicolumn{3}{c}{\textbf{CoT}} & & \multicolumn{3}{c}{\textbf{PS}} & &
\multicolumn{3}{c}{\textbf{IO}} & & \multicolumn{3}{c}{\textbf{CoT}} & & \multicolumn{3}{c}{\textbf{PS}} \\

\cmidrule(lr){3-5} \cmidrule(lr){7-9} \cmidrule(lr){11-13}
\cmidrule(lr){15-17} \cmidrule(lr){19-21} \cmidrule(lr){23-25}

\textbf{Model} & &
\textbf{-RP} & \textbf{-LA} & \textbf{RH} & &
\textbf{-RP} & \textbf{-LA} & \textbf{RH} & &
\textbf{-RP} & \textbf{-LA} & \textbf{RH} & &
\textbf{-RP} & \textbf{-LA} & \textbf{RH} & &
\textbf{-RP} & \textbf{-LA} & \textbf{RH} & &
\textbf{-RP} & \textbf{-LA} & \textbf{RH} \\
\midrule

Gemma 3 & &
\none{4.8} & \gain{8.8} & \smallgain{6.0} & &
\none{5.2} & \gain{9.2} & \smallgain{6.4} & &
\none{3.2} & \smallgain{7.2} & \gain{7.6} & &
\none{42.1} & \gain{44.5} & \smallgain{44.0} & &
\none{43.7} & \smallgain{44.8} & \gain{45.7} & &
\none{45.2} & \smallgain{46.3} & \gain{49.1} \\

gpt-oss & &
\none{92.8} & \gain{93.6} & \gain{93.6} & &
\none{89.6} & \gain{93.6} & \smallgain{93.2} & &
\none{86.4} & \smallgain{92.8} & \gain{94.0} & &
\smallgain{82.0} & \none{80.8} & \gain{82.8} & &
\none{80.3} & \none{80.3} & \gain{83.2} & &
\none{80.4} & \smallgain{80.9} & \gain{84.0} \\

Llama 3.1 8B & &
\none{2.8} & \gain{21.6} & \smallgain{12.8} & &
\none{5.6} & \smallgain{12.4} & \gain{18.4} & &
\none{10.0} & \smallgain{36.8} & \gain{41.2} & &
\none{38.8} & \gain{45.8} & \smallgain{44.2} & &
\none{41.3} & \smallgain{44.1} & \gain{46.6} & &
\none{45.2} & \smallgain{54.4} & \gain{56.0} \\

Llama 3.1 70B & &
\none{76.0} & \gain{80.8} & \smallgain{78.4} & &
\none{77.6} & \gain{82.0} & \smallgain{78.8} & &
\gain{86.0} & \smallgain{83.2} & \none{80.8} & &
\none{71.7} & \gain{74.9} & \smallgain{73.6} & &
\none{73.4} & \gain{75.7} & \smallgain{73.5} & &
\gain{76.8} & \smallgain{76.1} & \none{75.3} \\

Mistral Small & &
\none{2.8} & \smallgain{4.0} & \gain{11.2} & &
\none{2.4} & \smallgain{4.8} & \gain{11.6} & &
\smallgain{3.2} & \none{2.8} & \gain{12.0} & &
\none{46.3} & \smallgain{47.7} & \gain{51.9} & &
\none{48.4} & \smallgain{50.0} & \gain{53.2} & &
\none{50.6} & \smallgain{51.7} & \gain{54.8} \\

Phi-4 & &
\none{20.8} & \smallgain{28.8} & \gain{54.0} & &
\smallgain{45.6} & \none{30.0} & \gain{62.8} & &
\smallgain{40.8} & \none{33.2} & \gain{61.2} & &
\none{57.3} & \smallgain{59.6} & \gain{68.7} & &
\smallgain{67.2} & \none{61.4} & \gain{73.2} & &
\smallgain{65.4} & \none{62.9} & \gain{71.0} \\

Qwen3 & &
\gain{84.0} & \none{75.6} & \smallgain{81.6} & &
\smallgain{83.2} & \none{80.8} & \gain{84.4} & &
\gain{85.6} & \none{84.0} & \smallgain{84.8} & &
\gain{74.9} & \none{72.6} & \smallgain{73.9} & &
\none{75.1} & \gain{75.6} & \gain{75.6} & &
\smallgain{76.7} & \gain{77.3} & \none{76.5} \\

\cdashlinelr{1-25}

Average & &
\none{40.6} & \smallgain{44.7} & \gain{48.2} & &
\none{44.2} & \smallgain{44.7} & \gain{50.8} & &
\none{45.0} & \smallgain{48.6} & \gain{54.5} & &
\none{59.0} & \smallgain{60.8} & \gain{62.7} & &
\none{61.3} & \smallgain{61.7} & \gain{64.4} & &
\none{62.9} & \smallgain{64.2} & \gain{66.7} \\

\bottomrule

\end{tabular}
}
\vspace{-1em}
\end{table*}

%% file: tables/ltm.tex
\begin{table*}[ht]
\vspace{-0.5em}
\centering
\setlength{\fboxsep}{0pt}
\setlength{\fboxrule}{0pt}

\fbox{\begin{minipage}{.99\textwidth}
\centering

\vspace{.7em}

\begin{minipage}[t]{0.50\textwidth}
\centering
\vspace{0pt}

\includegraphics[width=\linewidth]{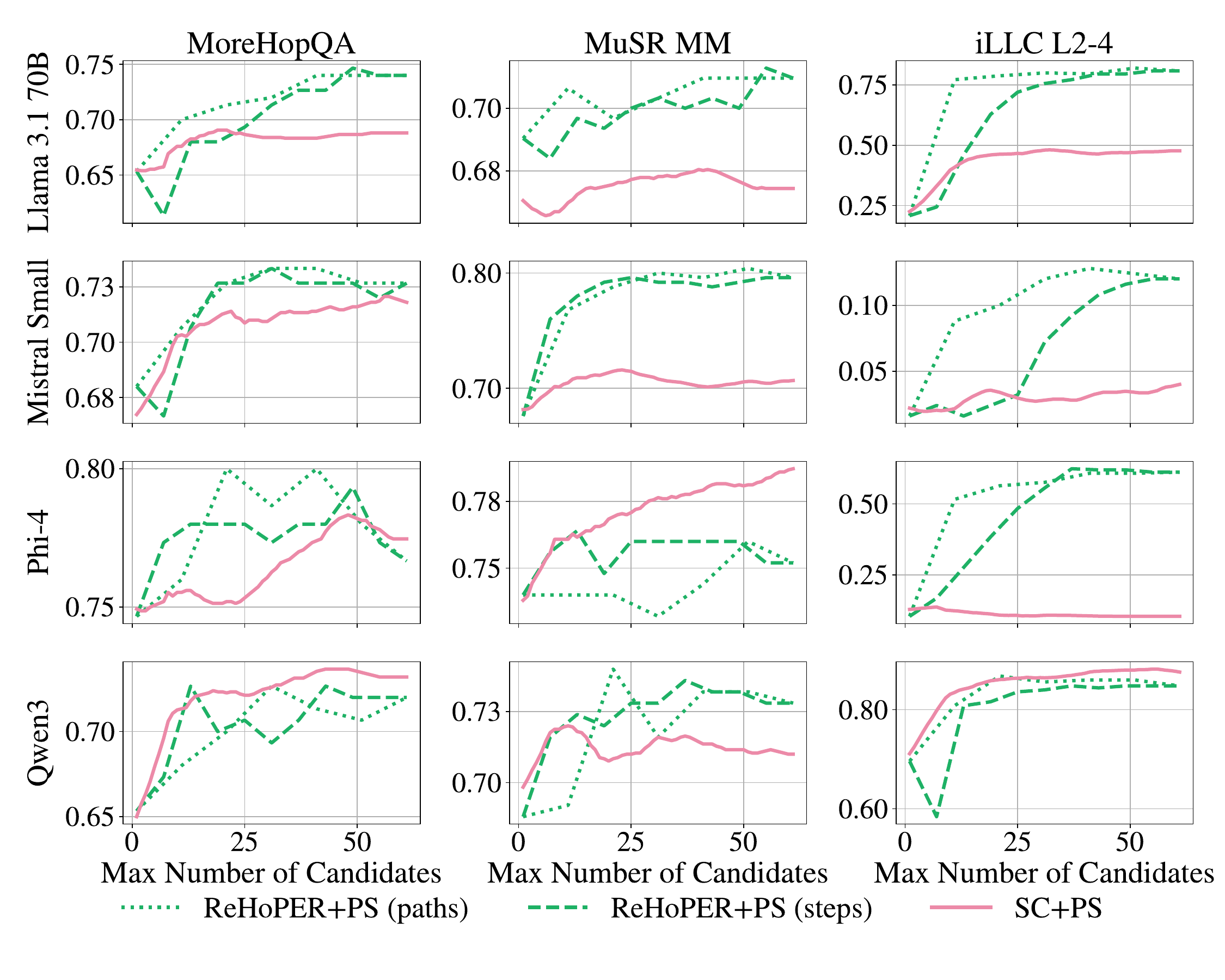}

\captionof{figure}{
Accuracy from allocating iterations to steps vs.\ paths.
}
\label{fig:pvs}
\end{minipage}
\hfill
\begin{minipage}[t]{0.47\textwidth}
\centering
\large
\captionof{table}{
Accuracy (\%) of ReHoPER+PS (RH), Least-to-Most (LtM), its proposed few-shot CoT baseline, and Self-Refine (SR). }
\label{tab:ltm}

\resizebox{\linewidth}{!}{\begin{tabular}{l p{0pt} cccc p{5pt} cccc}

\toprule

& &
\multicolumn{4}{c}{\textbf{iLLC~L1-4}} & &
\multicolumn{4}{c}{\textbf{iLLC~L2-4}} \\

\cmidrule(lr){3-6}
\cmidrule(lr){8-11}

\textbf{Model} & &
\textbf{CoT} & \textbf{LtM} & \textbf{SR} & \textbf{RH} & &
\textbf{CoT} & \textbf{LtM} & \textbf{SR} & \textbf{RH} \\

\midrule

Llama 3.1 70B & &
90.0 & 95.2 & 95.2 & \textbf{97.6} & &
13.2 & 12.0 & 28.0 & \textbf{80.8} \\

Mistral Small & &
0.0 & 81.6 & 86.8 & \textbf{91.2} & &
0.0 & 2.0 & 0.8 & \textbf{12.0} \\

Phi-4 & &
90.0 & \textbf{92.8} & \textbf{92.8} & 92.0 & &
8.8 & 6.4 & 10.0 & \textbf{61.2} \\

Qwen3 & &
53.2 & 45.6 & 79.2 & \textbf{86.0} & &
0.8 & 3.6 & 51.6 & \textbf{84.8} \\

\midrule

& &
\multicolumn{4}{c}{\textbf{iLLC~L1-6}} & &
\multicolumn{4}{c}{\textbf{iLLC~L2-6}} \\

\cmidrule(lr){3-6}
\cmidrule(lr){8-11}

\textbf{Model} & &
\textbf{CoT} & \textbf{LtM} & \textbf{SR} & \textbf{RH} & &
\textbf{CoT} & \textbf{LtM} & \textbf{SR} & \textbf{RH} \\

\midrule

Llama 3.1 70B & &
65.6 & 85.2 & 83.6 & \textbf{86.4} & &
2.0 & 5.6 & 15.6 & \textbf{74.4} \\

Mistral Small & &
0.0 & 58.0 & 69.2 & \textbf{78.8} & &
0.0 & 0.0 & 0.0 & \textbf{2.0} \\

Phi-4 & &
46.4 & \textbf{76.0} & 72.4 & \textbf{76.0} & &
0.0 & 0.8 & 3.6 & \textbf{41.6} \\

Qwen3 & &
0.0 & 8.8 & 64.8 & \textbf{80.0} & &
0.0 & 0.4 & 23.6 & \textbf{68.0} \\

\midrule

& &
\multicolumn{4}{c}{\textbf{iLLC~L1-8}} & &
\multicolumn{4}{c}{\textbf{iLLC~L2-8}} \\

\cmidrule(lr){3-6}
\cmidrule(lr){8-11}

\textbf{Model} & &
\textbf{CoT} & \textbf{LtM} & \textbf{SR} & \textbf{RH} & &
\textbf{CoT} & \textbf{LtM} & \textbf{SR} & \textbf{RH} \\

\midrule

Llama 3.1 70B & &
24.8 & \textbf{75.6} & 67.6 & 70.4 & &
0.4 & 1.2 & 6.4 & \textbf{54.0} \\

Mistral Small & &
0.0 & 53.2 & 54.8 & \textbf{59.6} & &
\textbf{0.0} & \textbf{0.0} & \textbf{0.0} & \textbf{0.0} \\

Phi-4 & &
0.0 & \textbf{66.4} & 60.4 & 61.2 & &
0.0 & 0.0 & 0.0 & \textbf{35.6} \\

Qwen3 & &
0.0 & 2.0 & 48.4 & \textbf{72.0} & &
0.0 & 0.0 & 16.8 & \textbf{43.6} \\

\bottomrule
\end{tabular}
}
\end{minipage}

\end{minipage}
}
\vspace{-1em}
\end{table*}

%% file: tables/time.tex
\begin{table*}[htb]
\centering
\caption{
Runtime (h:mm) and accuracy (\%) comparison between Self-Consistency (SC$_{61}$) and ReHoPER (RH) across iLLC L1 and L2 subsets. SC$_{181}$ uses 181 samples as a higher-compute SC comparison.
}
\label{tab:time}
\resizebox{1.\textwidth}{!}{
\begin{tabular}{ll p{0pt} ccc p{5pt} ccc p{5pt} ccc p{5pt} ccc}

\toprule

& & &
\multicolumn{3}{c}{\textbf{iLLC~L1-4}} & &
\multicolumn{3}{c}{\textbf{iLLC~L1-6}} & &
\multicolumn{3}{c}{\textbf{iLLC~L2-4}} & &
\multicolumn{3}{c}{\textbf{iLLC~L2-6}} \\

\cmidrule(lr){4-6}
\cmidrule(lr){8-10}
\cmidrule(lr){12-14}
\cmidrule(lr){16-18}

\textbf{Model} & & &
\textbf{SC$_{61}$} & \textbf{SC$_{181}$} & \textbf{RH} & &
\textbf{SC$_{61}$} & \textbf{SC$_{181}$} & \textbf{RH} & &
\textbf{SC$_{61}$} & \textbf{SC$_{181}$} & \textbf{RH} & &
\textbf{SC$_{61}$} & \textbf{SC$_{181}$} & \textbf{RH} \\

\midrule

\multirow{2}{*}{Llama 3.1 70B} & Runtime & &
1:16 & 3:37 & 2:42 & &
1:26 & 4:23 & 4:12 & &
2:11 & 6:23 & 7:56 & &
2:37 & 9:04 & 11:15 \\
& Accuracy & &
92.1 & 92.4 & \textbf{92.7} & &
76.0 & 76.9 & \textbf{80.1} & &
32.5 & 35.1 & \textbf{79.3} & &
17.2 & 16.7 & \textbf{68.5} \\

\cdashlinelr{1-18}

\multirow{2}{*}{Mistral Small} & Runtime & &
0:38 & 1:38 & 0:54 & &
0:43 & 2:03 & 0:55 & &
0:48 & 2:08 & 1:15 & &
0:56 & 2:41 & 1:13 \\
& Accuracy & &
90.7 & 90.8 & \textbf{91.3} & &
76.1 & \textbf{76.9} & 73.5 & &
3.5 & 2.4 & \textbf{11.6} & &
0.0 & 0.1 & \textbf{2.0} \\

\cdashlinelr{1-18}

\multirow{2}{*}{Phi-4} & Runtime & &
0:45 & 1:46 & 1:06 & &
0:52 & 2:31 & 1:57 & &
0:50 & 2:01 & 1:18 & &
1:00 & 2:55 & 2:21 \\
& Accuracy & &
91.2 & 91.2 & \textbf{92.0} & &
73.2 & 73.5 & \textbf{76.3} & &
11.1 & 10.8 & \textbf{59.3} & &
4.8 & 4.4 & \textbf{35.6} \\

\cdashlinelr{1-18}

\multirow{2}{*}{Qwen3} & Runtime & &
0:31 & 1:08 & 0:42 & &
0:36 & 1:43 & 1:06 & &
0:48 & 1:44 & 1:20 & &
0:57 & 2:08 & 1:39 \\
& Accuracy & &
81.7 & 81.7 & \textbf{85.2} & &
69.3 & 68.5 & \textbf{79.6} & &
71.1 & 70.5 & \textbf{83.6} & &
44.0 & 44.0 & \textbf{63.2} \\

\bottomrule

\end{tabular}
}
\vspace{-1em}
\end{table*}

%% file: appendix.tex
\clearpage
\section{Macro-aggregated results}
\label{app:macro-results}

Table~\ref{tab:result-macro} complements the per-subset results in Table~\ref{tab:result} by giving each benchmark family equal weight. We first average accuracy within the three MuSR tasks (MM, OP, and TA) and the six iLLC variants (L1-4, L1-6, L1-8, L2-4, L2-6, and L2-8), then average those family-level results with MoreHopQA. Under this aggregation, ReHoPER improves over the corresponding Self-Consistency (SC) baseline by 3.9pp for IO, 6.6pp for CoT, and 4.8pp for PS, with gains calculated before rounding the accuracies. Thus, the overall advantage persists when families with more subsets do not receive more weight.

The gains extend across all three benchmark families and all three answer formats. Averaged across models, ReHoPER improves MuSR by 2.9pp, 3.1pp, and 2.1pp for IO, CoT, and PS, respectively; MoreHopQA by 0.8pp, 2.6pp, and 2.3pp; and iLLC by 8.0pp, 14.2pp, and 10.0pp. The especially large improvements on iLLC are consistent with the per-subset results in Table~\ref{tab:result}, where the gains are greatest in the more demanding compositional settings. Together, these macro-aggregated results show that ReHoPER's advantage is not driven by the number of subsets in any one benchmark family.

\input{tables/result-macro}

\clearpage
\section{Answer aggregation across intermediate reasoning steps}
\label{app:step-aggregation}

ReHoPER produces candidate predictions not only at the end of each reasoning path, but also after each intermediate question. We therefore analyze how different ways of aggregating these predictions affect performance over the rollout. Let $\hat{y}_0$ be the initial prediction and $\hat{y}^{(i)}_t$ be the prediction from path $i$ after step $t$. We compare three aggregation summaries at step $k$. \textbf{Cumulative Vote} (CV) applies majority voting over $\hat{y}_0$ and all predictions from all paths up to step $k$:
\begin{equation}
\mathrm{CV}_k =
\mathrm{MV}\!\left(
\langle \hat{y}_0 \rangle
\parallel
\Big(\parallel_{i=1}^{n}\parallel_{t=1}^{\min(k,m_i)}
\langle \hat{y}^{(i)}_t \rangle\Big)
\right),
\end{equation}
where $\mathrm{MV}$ denotes majority vote. This is the default ReHoPER aggregation when $k=m_{\max}$.

For step-wise aggregation, we define the carried-forward prediction for path $i$ at step $k$ as
\begin{equation}
\tilde{y}^{(i)}_k = \hat{y}^{(i)}_{\min(k,m_i)},
\end{equation}
where we set $\hat{y}^{(i)}_0=\hat{y}_0$ for all paths. Thus, if path $i$ has already terminated before step $k$, $\tilde{y}^{(i)}_k$ is its last available prediction. \textbf{Current-Step Vote} (CSV) applies majority voting over exactly one carried-forward prediction from each path:
\begin{equation}
\mathrm{CSV}_k =
\mathrm{MV}\!\left(
\left\langle
\tilde{y}^{(1)}_k,\tilde{y}^{(2)}_k,\ldots,\tilde{y}^{(n)}_k
\right\rangle
\right).
\end{equation}
\textbf{Current-Step Mean} (CSM) uses the same carried-forward predictions, but reports their mean correctness rather than applying majority voting:
\begin{equation}
\mathrm{CSM}_k =
\frac{1}{n}\sum_{i=1}^{n}
\mathbf{1}\!\left[\tilde{y}^{(i)}_k = y\right].
\end{equation}
Thus, CSM is a diagnostic measure of per-path quality rather than an inference-time aggregation rule.

\begin{figure}[t]
    \centering
    \includegraphics[width=.78\linewidth]{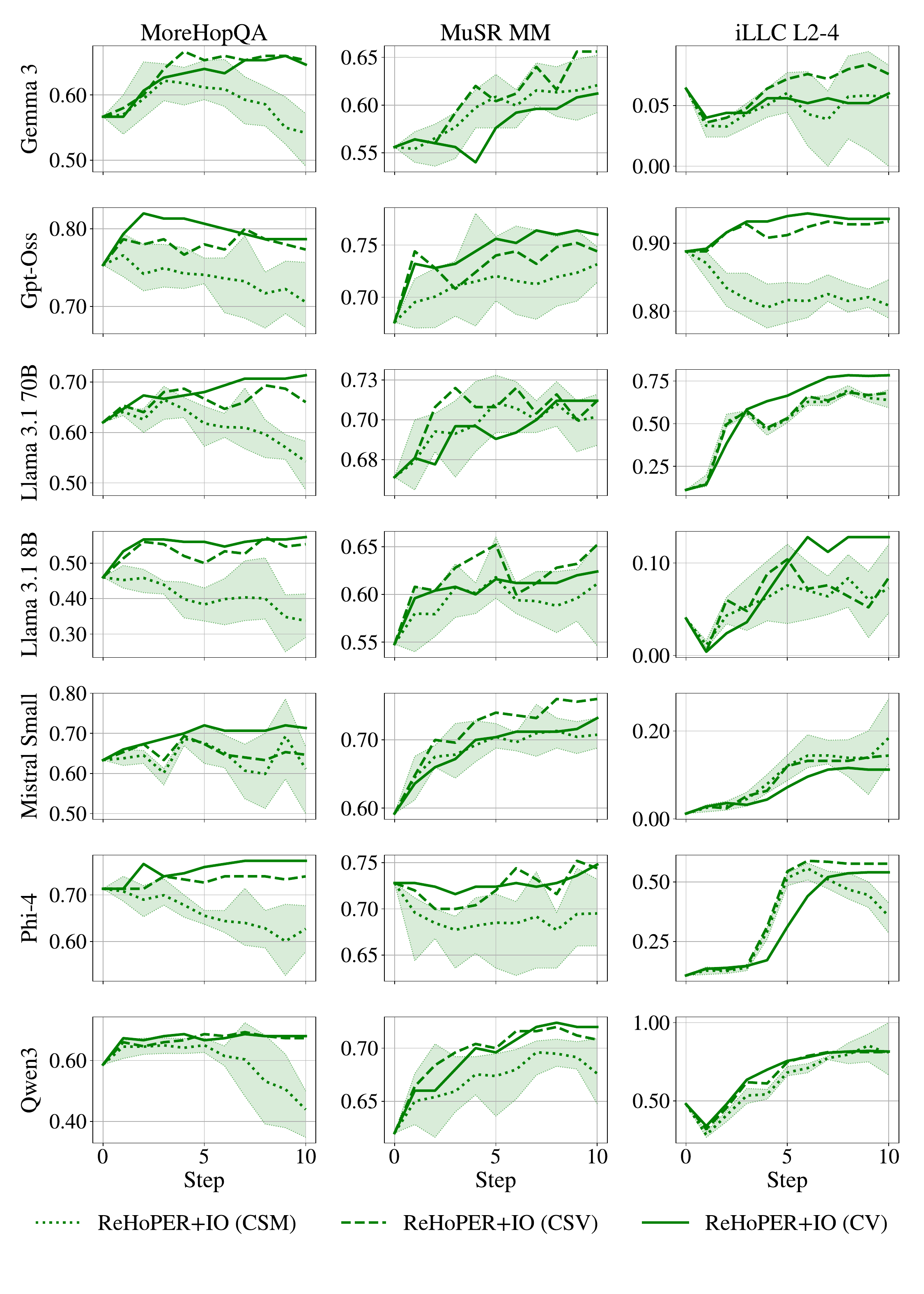}
    \caption{
    Step-wise aggregation analysis for ReHoPER under the \textbf{IO} answer format.
    Each panel shows accuracy as a function of the reasoning step for one model--dataset pair.
    CV denotes cumulative voting over all predictions up to the current step; CSV votes over one carried-forward prediction per path at the current step; CSM reports the mean carried-forward per-path correctness.
    Shaded bands show the standard deviation for the per-path CSM diagnostic.
    }
    \label{fig:agg-step-io}
\end{figure}

\begin{figure}[t]
    \centering
    \includegraphics[width=.78\linewidth]{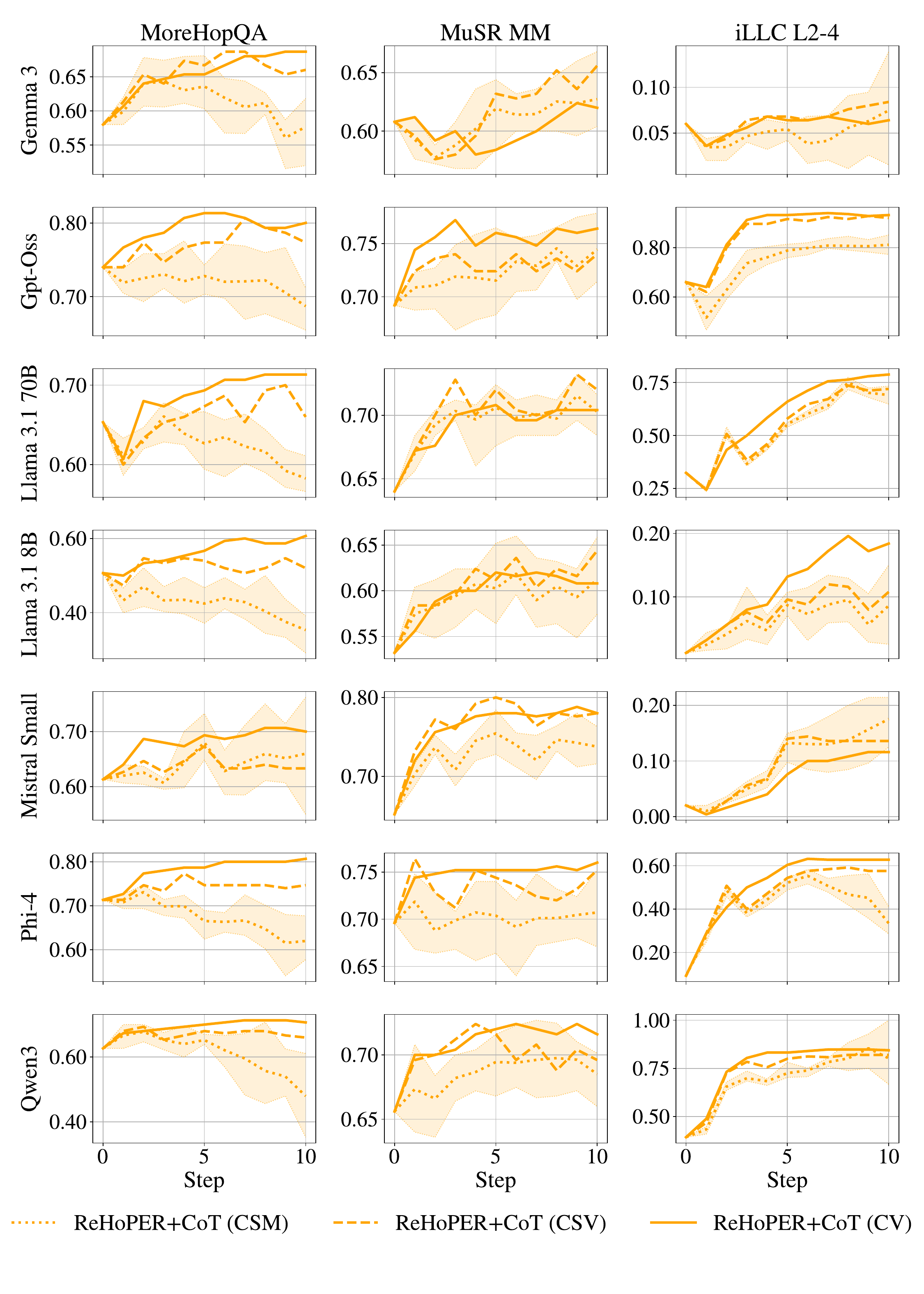}
    \caption{
    Step-wise aggregation analysis for ReHoPER under the \textbf{CoT} answer format.
    Each panel shows accuracy as a function of the reasoning step for one model--dataset pair.
    CV denotes cumulative voting over all predictions up to the current step; CSV votes over one carried-forward prediction per path at the current step; CSM reports the mean carried-forward per-path correctness.
    Shaded bands show the standard deviation for the per-path CSM diagnostic.
    }
    \label{fig:agg-step-cot}
\end{figure}

\begin{figure}[t]
    \centering
    \includegraphics[width=.78\linewidth]{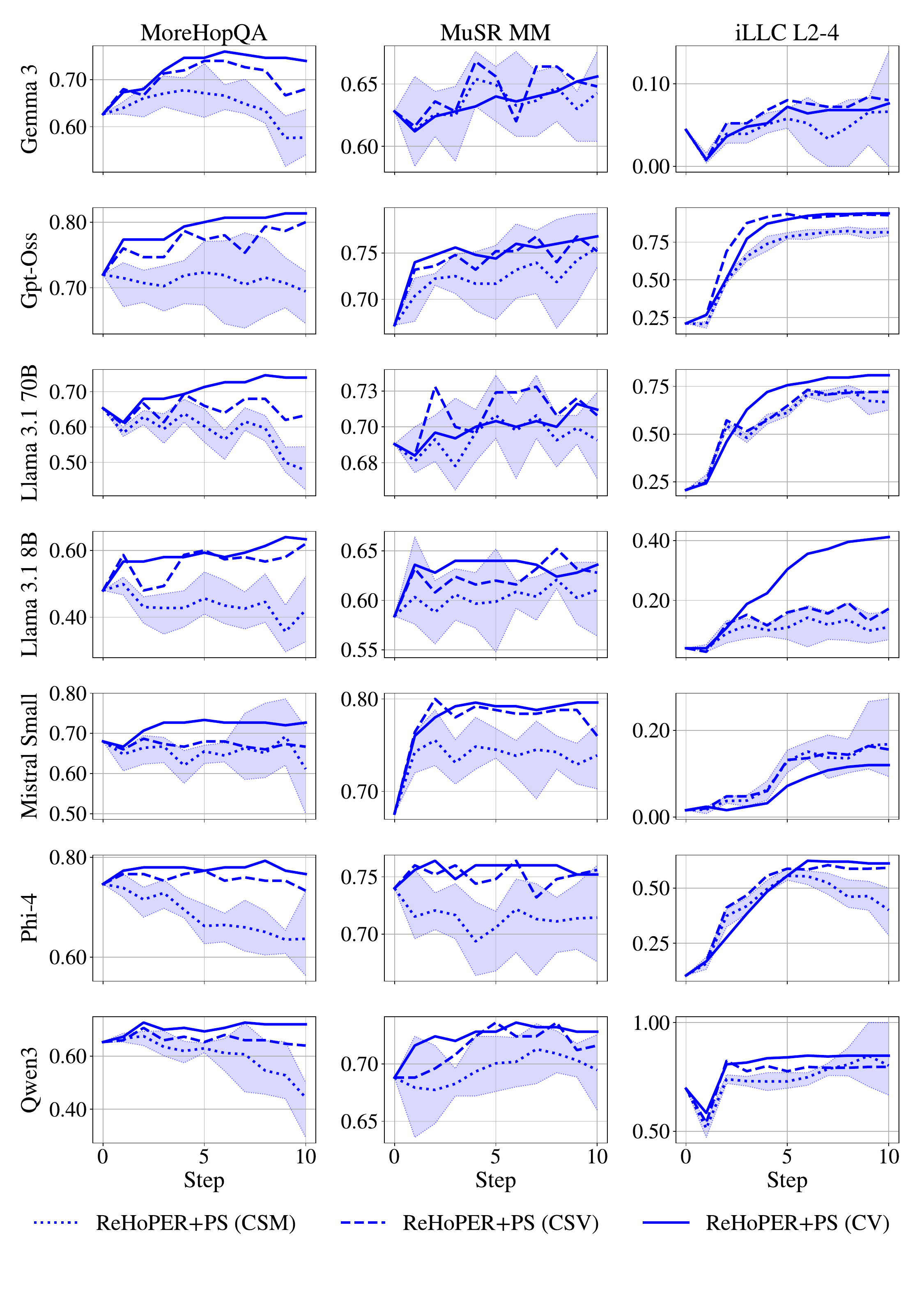}
    \caption{
    Step-wise aggregation analysis for ReHoPER under the \textbf{PS} answer format.
    Each panel shows accuracy as a function of the reasoning step for one model--dataset pair.
    CV denotes cumulative voting over all predictions up to the current step; CSV votes over one carried-forward prediction per path at the current step; CSM reports the mean carried-forward per-path correctness.
    Shaded bands show the standard deviation for the per-path CSM diagnostic.
    }
    \label{fig:agg-step-ps}
\end{figure}

Figures~\ref{fig:agg-step-io}--\ref{fig:agg-step-ps} show the results for IO, CoT, and PS across datasets and models. Overall, cumulative aggregation is consistently competitive and is usually the strongest final aggregation strategy. Averaged over all plotted dataset--model--prefix configurations, final-step CV reaches $64.6\%$ accuracy, compared with $62.2\%$ for CSV and $55.5\%$ for CSM. CV is also the best or tied for best final aggregation rule in 48 out of 63 configurations, and under PS in 18 out of 21 configurations. The gain over the initial prediction is largest on iLLC L2-4, where performance often continues to improve as more intermediate reasoning states are accumulated.

These results support the default ReHoPER design: useful evidence is distributed across multiple intermediate states, and retaining earlier predictions can make the final answer more robust than relying only on the most recent state. CSV is occasionally competitive, especially on flatter settings such as MuSR MM, but the cumulative vote provides a more reliable choice across prefixes, datasets, and model families. CSM is generally lower than both voting strategies, highlighting the value of ReHoPER's aggregation mechanism: predictions from different paths and intermediate steps provide complementary evidence that, when combined, yields more reliable final decisions.

\clearpage
\section{Answer aggregation across stochastic reasoning paths}
\label{app:path-aggregation}

We analyze how ReHoPER's predictions accumulate across stochastic reasoning paths. This complements the step-wise analysis in Appendix~\ref{app:step-aggregation}: instead of varying the reasoning step within a fixed set of paths, we vary the number of paths included in the aggregation. Let $\hat{y}_0$ denote the initial prediction, shared across all paths, and let $\hat{y}^{(i)}_t$ be the prediction from path $i$ after step $t$. For a path index $j \in \{1,\ldots,n\}$, \textbf{Cumulative Vote} (CV) aggregates predictions from paths $1$ through $j$:
\begin{equation}
c_j =
\mathrm{MV}\!\left(
\langle \hat{y}_0 \rangle
\parallel
\Big(\parallel_{i=1}^{j}\parallel_{t=1}^{m_i}
\langle \hat{y}^{(i)}_t \rangle\Big)
\right).
\end{equation}
Here, the initial prediction $\hat{y}_0$ is included only once, rather than being duplicated for each path.

To isolate the contribution of individual paths, \textbf{Current-Path Vote} (CPV) applies majority voting using only the predictions from path $j$, together with the shared initial prediction:
\begin{equation}
v_j =
\mathrm{MV}\!\left(
\langle \hat{y}_0 \rangle
\parallel
\Big(\parallel_{t=1}^{m_j}
\langle \hat{y}^{(j)}_t \rangle\Big)
\right).
\end{equation}
\textbf{Current-Path Mean} (CPM) uses the same path-local predictions, but reports their mean correctness:
\begin{equation}
q_j =
\frac{1}{m_j+1}
\left(
\mathbf{1}[\hat{y}_0 = y]
+
\sum_{t=1}^{m_j}
\mathbf{1}[\hat{y}^{(j)}_t = y]
\right).
\end{equation}
The plotted CV and CPV values are obtained by averaging $\mathbf{1}[c_j=y]$ and $\mathbf{1}[v_j=y]$ across examples, respectively, while CPM averages $q_j$ across examples. Finally, \textbf{All-Path Vote} (APV) and \textbf{All-Path Mean} (APM) summarize the average path-local performance across all paths:
\begin{equation}
\mathrm{APV}
=
\frac{1}{n}\sum_{j=1}^{n} \mathrm{CPV}_j,
\qquad
\mathrm{APM}
=
\frac{1}{n}\sum_{j=1}^{n} \mathrm{CPM}_j.
\end{equation}
These two quantities are shown as flat reference lines because they average over all paths rather than varying cumulatively with the path index.

\begin{figure}[t]
    \centering
    \includegraphics[width=.78\linewidth]{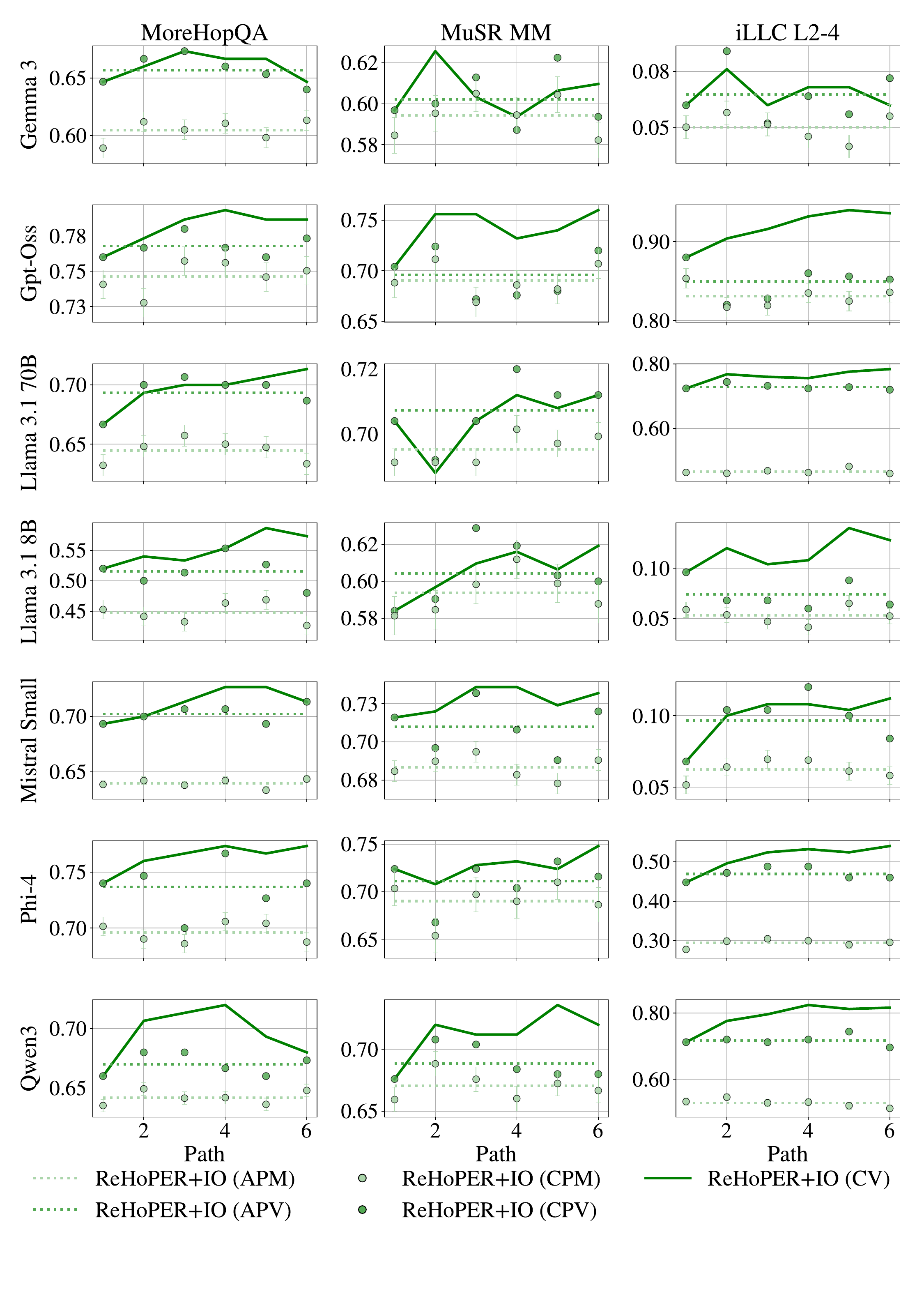}
    \caption{
    Path-wise aggregation analysis for ReHoPER under the \textbf{IO} answer format.
    CV cumulatively votes over predictions from paths $1$ through $j$, including the shared initial prediction only once.
    CPV and CPM summarize the current path alone, while APV and APM show the average path-local vote and mean correctness across all paths.
    }
    \label{fig:agg-path-io}
\end{figure}

\begin{figure}[t]
    \centering
    \includegraphics[width=.78\linewidth]{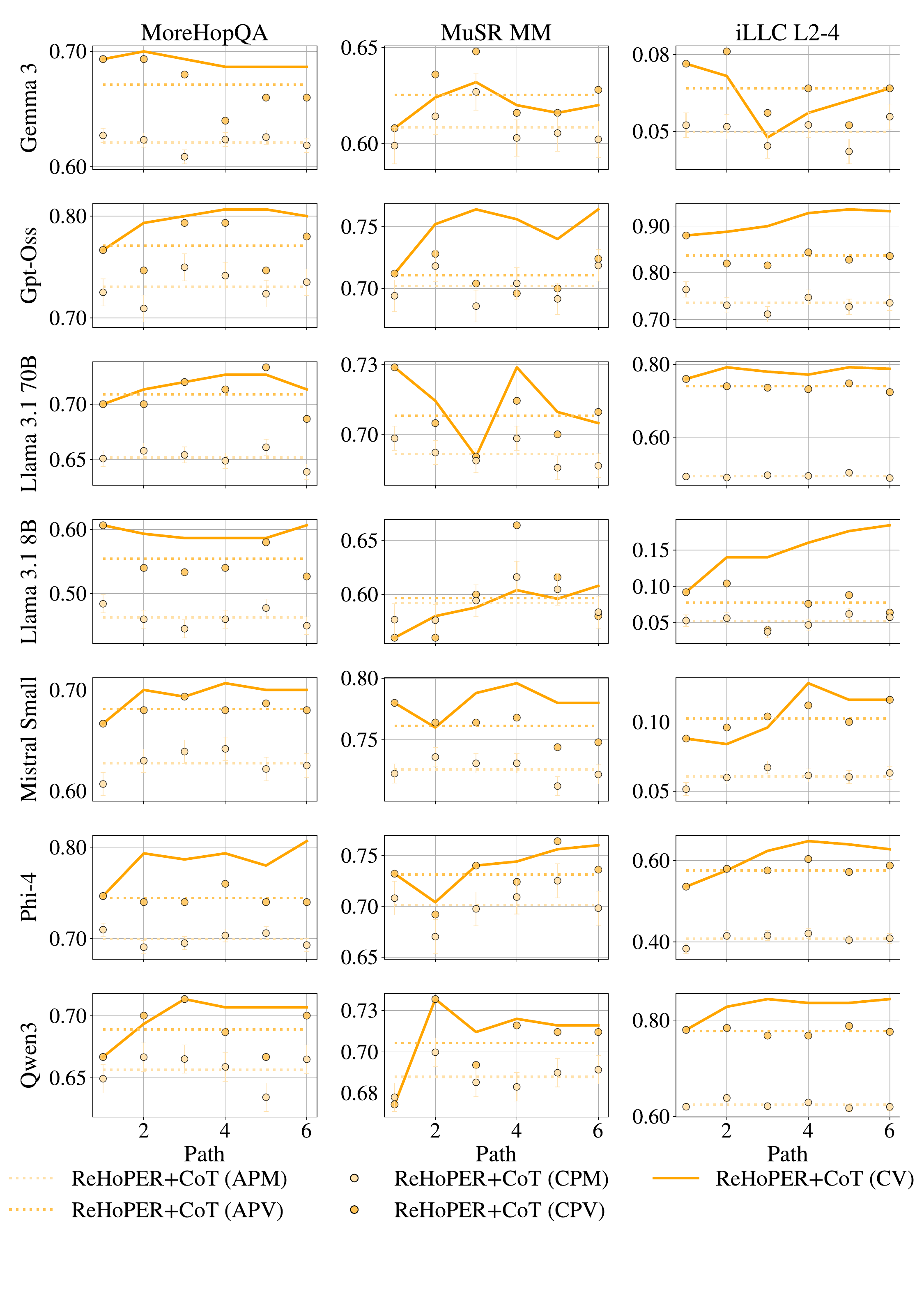}
    \caption{
    Path-wise aggregation analysis for ReHoPER under the \textbf{CoT} answer format.
    CV cumulatively votes over predictions from paths $1$ through $j$, including the shared initial prediction only once.
    CPV and CPM summarize the current path alone, while APV and APM show the average path-local vote and mean correctness across all paths.
    }
    \label{fig:agg-path-cot}
\end{figure}

\begin{figure}[t]
    \centering
    \includegraphics[width=.78\linewidth]{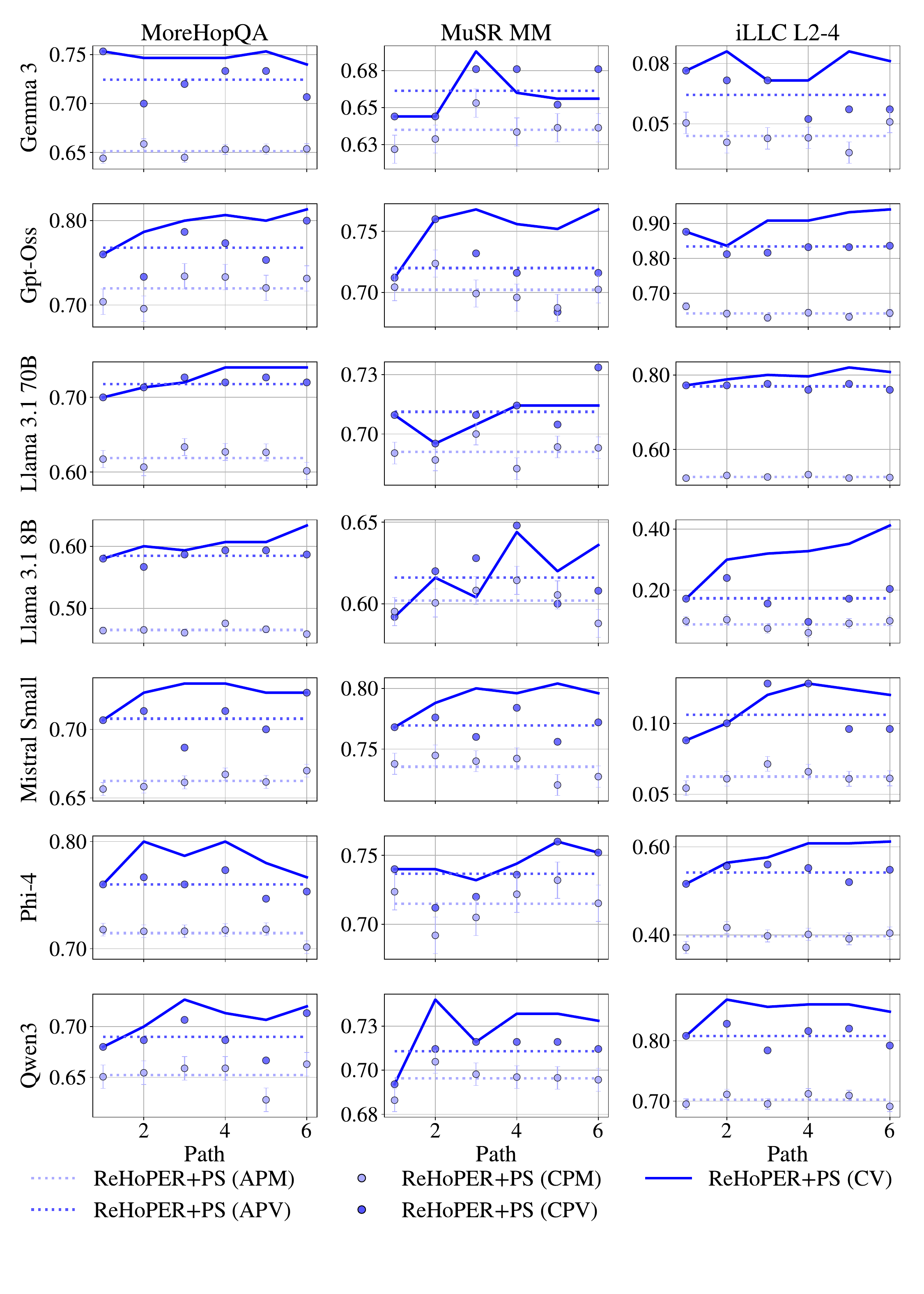}
    \caption{
    Path-wise aggregation analysis for ReHoPER under the \textbf{PS} answer format.
    CV cumulatively votes over predictions from paths $1$ through $j$, including the shared initial prediction only once.
    CPV and CPM summarize the current path alone, while APV and APM show the average path-local vote and mean correctness across all paths.
    }
    \label{fig:agg-path-ps}
\end{figure}

Figures~\ref{fig:agg-path-io}--\ref{fig:agg-path-ps} show the resulting path-wise aggregation trends for IO, CoT, and PS. Overall, cumulative voting across paths is generally more reliable than relying on a single path in isolation. CPV and CPM show that individual paths can already provide useful predictions, while the gap between CV and the path-local summaries highlights the value of combining complementary evidence from multiple stochastic trajectories. The flat APV and APM lines provide a useful reference for the average quality of individual paths; when CV rises above these lines, it indicates that cumulative aggregation benefits from pooling complementary predictions across paths rather than merely matching the average path-local behavior.

These results further support ReHoPER's aggregation design. The method benefits from both within-path reasoning depth and across-path diversity: each path explores a different reasoning trajectory, and cumulative voting pools the useful predictions contributed by different paths while preserving the evidence accumulated in earlier ones. This makes the final prediction less dependent on any single trajectory and better reflects the collective evidence gathered during the full ReHoPER rollout.

\clearpage
\section{Additional paths vs.\ steps results}
\label{app:pvs}

\begin{figure}[htbp]
  \centering
  \includegraphics[width=.79\columnwidth]{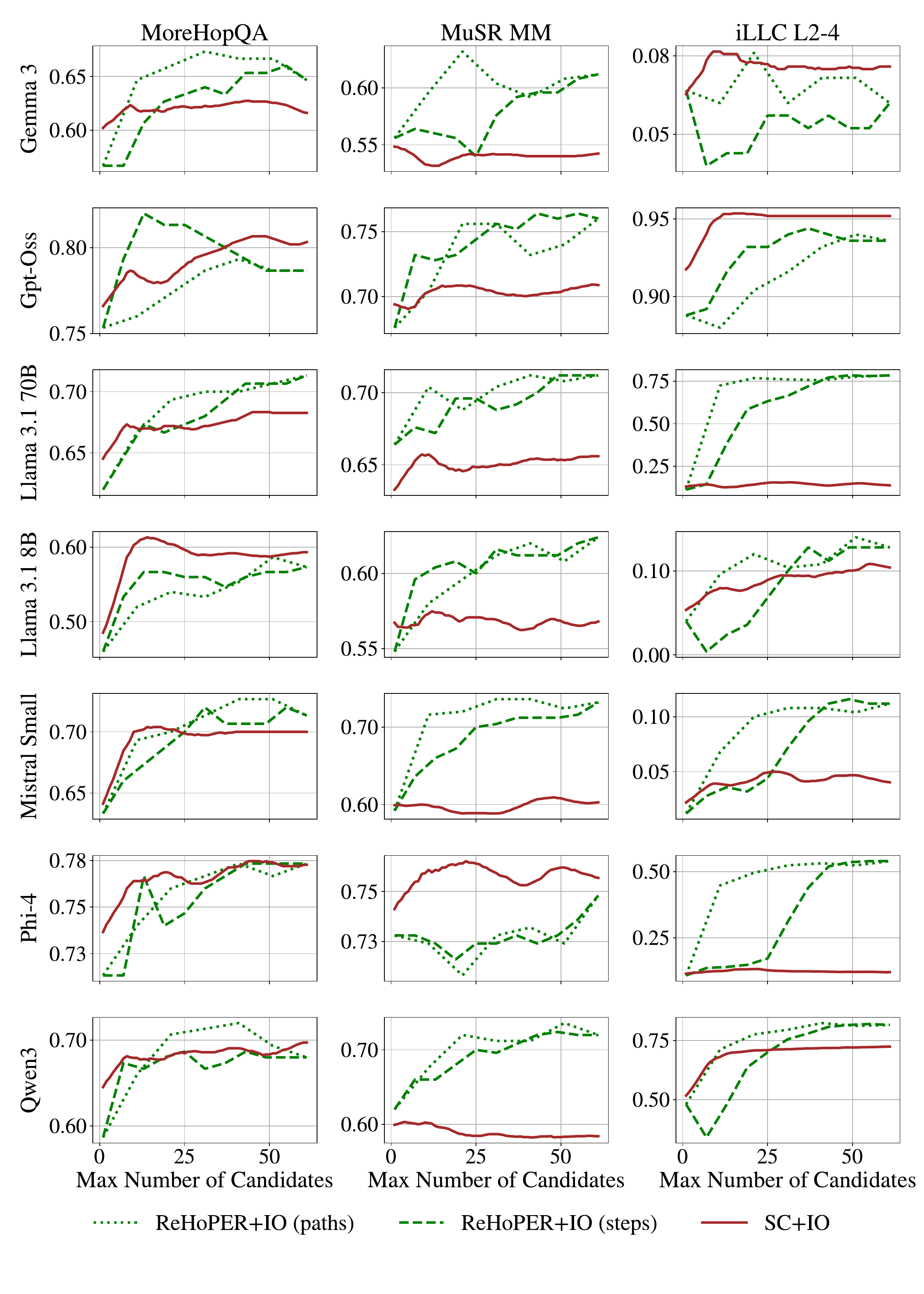}
  \caption{Accuracy gains from allocating iterations to steps vs.\ paths under the \textbf{IO} answer format.}
  \label{fig:pvs-io}
\end{figure}

\begin{figure}[htbp]
  \centering
  \includegraphics[width=.79\columnwidth]{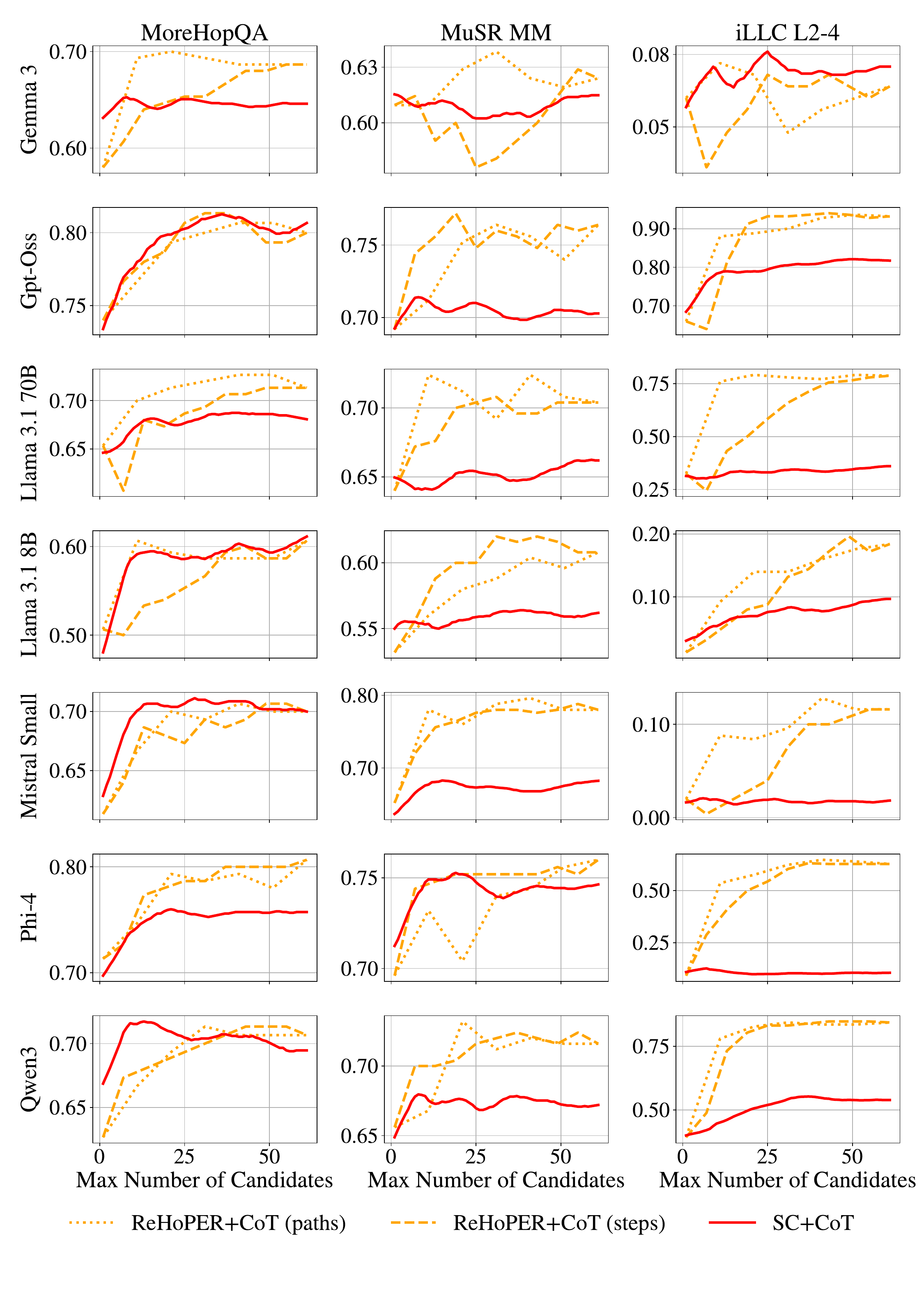}
  \caption{Accuracy gains from allocating iterations to steps vs.\ paths under the \textbf{CoT} answer format.}
  \label{fig:pvs-cot}
\end{figure}

\begin{figure}[htbp]
  \centering
  \includegraphics[width=.79\columnwidth]{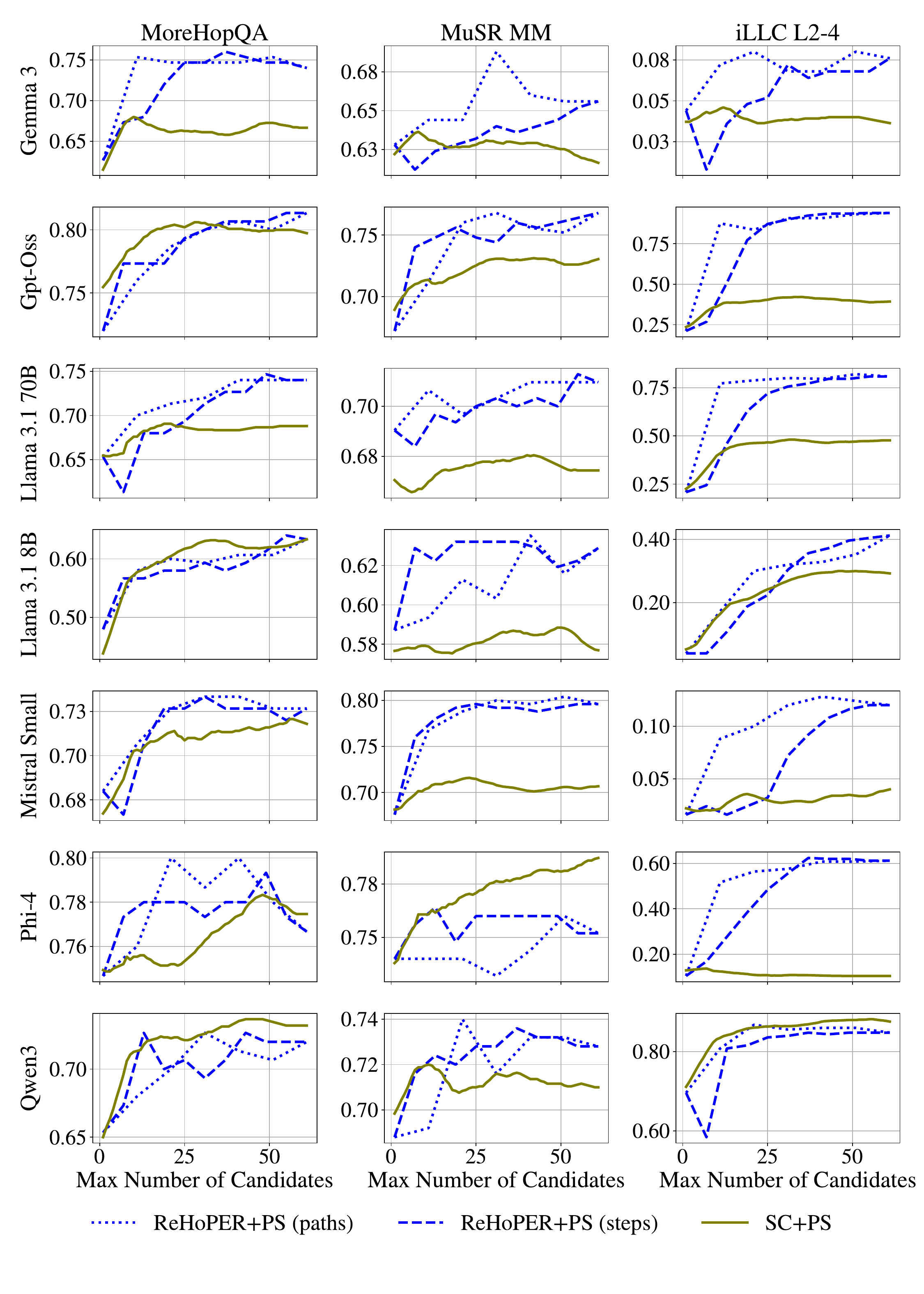}
  \caption{Accuracy gains from allocating iterations to steps vs.\ paths under the \textbf{PS} answer format.}
  \label{fig:pvs-ps}
\end{figure}

Figures~\ref{fig:pvs-io}--\ref{fig:pvs-ps} extend the paths-vs-steps analysis in Figure~\ref{fig:pvs} to additional model backbones and output prefixes (IO, CoT, and PS). At each budget, accuracy is computed by majority vote over a list of candidate predictions. Let \(\hat{y}_0\) be the initial prediction, \(\hat{y}^{(i)}_k\) the prediction at step \(k\) of path \(i\), and \(m_i\) the number of steps completed on that path. For the dashed (steps) curves, we fix \(p=6\) paths and increase the maximum step cap \(s\) from 0 to 10. The candidate list is
\(\langle \hat{y}_0\rangle \;\mathbin{\|}\; \Big( \mathbin{\|}_{i=1}^{6}\; \mathbin{\|}_{k=1}^{\min(m_i,\ s)} \langle \hat{y}^{(i)}_{k}\rangle \Big)\),
so the maximum number of candidates at budget \(s\) is \(N_{\max}(s)=1+6s\). For the dotted (paths) curves, we fix the per-path step cap to \(s=10\) and instead increase the number of paths \(p\) from 0 to 6. The candidate list is
\(\langle \hat{y}_0\rangle \;\mathbin{\|}\; \Big( \mathbin{\|}_{i=1}^{p}\; \mathbin{\|}_{k=1}^{\min(m_i,\ 10)} \langle \hat{y}^{(i)}_{k}\rangle \Big)\),
so the maximum number of candidates at budget \(p\) is \(N_{\max}(p)=1+10p\). These maxima are upper bounds: some paths terminate early, so fewer candidates may be included in the vote. The solid SC curve is computed analogously by aggregating over the first \(t=1,\ldots,\,N_{\max}\) resampled candidates. For readability, we smooth its accuracy trajectory with a 10-point moving average.

\clearpage
\section{Intermediate question statistics}
\label{app:iq-stats}

Tables~\ref{tab:iq-statistics-rh-1}--\ref{tab:iq-statistics-on} summarize the intermediate questions (IQs) produced by ReHoPER and its two ablations. The column \#IQ reports the average total number of valid IQs generated per example across all paths, while \#IQ/Path reports the corresponding average per path; its subscript is the average standard deviation of the number of IQs across paths for each example. Similarly, \#uniqIQ counts the average number of distinct valid IQs across all paths, and \#uniqIQ/Path reports the average number of distinct IQs within each path, again with the corresponding across-path standard deviation. Finally, \%uniqIQ measures global diversity across paths, i.e., \#uniqIQ divided by \#IQ, while \%uniqIQ/Path measures within-path diversity, i.e., \#uniqIQ/Path divided by \#IQ/Path.

\looseness=-1
A useful sanity check is that \%uniqIQ/Path is essentially \(100\%\) for ReHoPER across all datasets in Tables~\ref{tab:iq-statistics-rh-1} and~\ref{tab:iq-statistics-rh-2}. This is expected because ReHoPER explicitly selects the first candidate IQ that has not already been asked in the current path, and terminates the path if all candidates repeat. The ablation without replanning shows the same pattern in Table~\ref{tab:iq-statistics-off}: although it does not replan, it commits to a single ordered list of IQs at the beginning of each path, and repeated questions within that fixed list are rare. In contrast, the ablation without look-ahead generates only one next IQ at a time without a horizon or selection step, so it can repeatedly ask the same or near-identical questions. Importantly, this repetition happens even though the instruction for the ablation without look-ahead explicitly asks the model not to repeat previously asked questions (Figure~\ref{fig:i-single} in Appendix~\ref{app:instruct}). This suggests that prompt-level discouragement alone is not sufficient to reliably enforce non-repetition; the explicit selection mechanism in ReHoPER is what keeps within-path uniqueness at \(100\%\).

On the three datasets used for the planning ablations, trajectory lengths and question diversity differ across configurations. The ablation without look-ahead almost always reaches the maximum path length, generating about \(60\) IQs in total, or \(10\) IQs per path, because its termination criterion is tied to the step cap rather than the availability of useful new questions. However, this larger number of generated IQs does not necessarily correspond to more distinct questions. On MoreHopQA, the ablation without look-ahead generates \(60.00\) IQs on average but only \(20.51\) unique IQs, while ReHoPER generates fewer total IQs (\(43.07\)) but more unique IQs (\(24.25\)). The same pattern appears in the within-path counts: the ablation without look-ahead asks \(10.00\) IQs per path but only \(5.77\) unique IQs per path, whereas ReHoPER asks \(7.18\) IQs per path and all \(7.18\) are unique. The ablation without replanning, by contrast, produces much shorter trajectories: averaged across models, it generates only \(3.03\) IQs per path on MoreHopQA and \(3.89\) on iLLC L2-4, compared with \(7.18\) and \(7.33\) for ReHoPER. ReHoPER therefore produces substantially more valid and unique IQs than the ablation without replanning, while avoiding the repeated-question behavior observed in the ablation without look-ahead.

Across datasets, ReHoPER produces the most saturated trajectories on MuSR. In Table~\ref{tab:iq-statistics-rh-1}, the average \#IQ/Path is close to the maximum on MuSR MM, OP, and TA, and global uniqueness remains high, ranging from \(90.04\%\) to \(96.70\%\) on average across the three MuSR subsets. MoreHopQA shows a different pattern, as the global uniqueness in ReHoPER drops to \(55.47\%\) on average. This means that different stochastic paths often converge to similar IQs, suggesting that the task admits a more constrained set of useful decompositions.

The iLLC results in Table~\ref{tab:iq-statistics-rh-2} show a consistent increase in the amount and diversity of generated IQs as the task becomes more compositional. For example, moving from iLLC L1-4 to iLLC L2-4 increases the average \#IQ from \(38.62\) to \(43.99\) and the average \#uniqIQ from \(17.95\) to \(22.87\). A similar pattern holds at longer lengths: iLLC L2-8 yields more total and unique IQs than iLLC L1-8, with \#uniqIQ increasing from \(23.27\) to \(27.37\). The global uniqueness rate also tends to be higher in the harder L2 variants than in the corresponding L1 variants. This suggests that harder iLLC settings induce a broader set of useful intermediate questions, while ReHoPER still preserves non-redundancy within each individual path.

Finally, the gap between \%uniqIQ and \%uniqIQ/Path is informative. Since \%uniqIQ/Path is always \(100\%\) for ReHoPER, any reduction in \%uniqIQ reflects overlap across different paths rather than repetition within a path. This overlap is not necessarily undesirable, as on some tasks, multiple paths may independently identify the same useful subquestions. At the same time, model-level variation shows that some models explore more diverse decompositions than others. For instance, on iLLC, Llama 3.1 8B and gpt-oss often maintain relatively high global uniqueness, whereas Phi-4 and Llama 3.1 70B tend to generate more overlapping IQs across paths. Thus, these statistics complement accuracy results by showing not only how many intermediate questions are generated, but also whether the reasoning paths provide genuinely different decompositions or repeatedly revisit the same subproblems.

\input{tables/iq-statistics}

\clearpage
\section{Trajectory changes under replanning}
\label{app:trajectory-changes}

To further examine whether ReHoPER simply follows its initial plan or actively adapts its reasoning trajectory, we analyze how often replanning changes the next intermediate question. At each step $i$, the current plan contains a predicted next question and a ``second-next'' question. After answering the current question, ReHoPER replans from the updated history. We compare the second-next question from the previous plan with the actual next question selected after replanning. If they differ, we count the transition as replanned.

Tables~\ref{tab:trajectory-1} and~\ref{tab:trajectory-2} report the resulting trajectory statistics. For a step pair $i \rightarrow i+1$, $T$ denotes the average number of instances that are still active and have not terminated by that comparison point, $R$ denotes the average number of those active instances whose next question changes after replanning, and $N$ denotes the total number of instances in the dataset. Thus, $\%T/N$ measures how much of the dataset remains active at that step, $\%R/T$ measures how often replanning changes the next question among active instances, and $\%R/N$ measures the same change rate relative to the full dataset. The \textsc{ANY} row reports whether an instance has at least one such replanning-induced change across all step pairs. All entries are reported as mean$_{\mathrm{std}}$ across paths.

Overall, replanning frequently changes the reasoning trajectory rather than merely continuing a fixed plan. This pattern is especially clear on MuSR MM, where nearly all instances remain active across steps and $\%R/T$ is consistently high across models. MoreHopQA shows a similar trend among active instances, although $\%T/N$ decreases at later steps as more paths terminate. The iLLC results require more nuance. Because iLLC admits a natural word-by-word procedure, once the model starts processing words sequentially, the second-next question from the previous plan often matches the next question regenerated after replanning. This is most visible around steps 2--5, where many trajectories enter this stable extraction template, yielding lower $\%R/T$ despite continued replanning. However, this local reduction does not mean that replanning is absent: other steps often still show high replanning rates, and the \textsc{ANY} rows show that most iLLC instances experience at least one replanning-induced change. Thus, lower step-level replanning on some iLLC transitions often reflects preservation of a useful sequential decomposition rather than inactive replanning. Together, these results support the interpretation that ReHoPER's receding-horizon mechanism repeatedly updates the intermediate-question trajectory as new information is added, while still preserving stable subplans when they remain appropriate.

\input{tables/trajectory}

\clearpage
\section{Hyperparameter optimization}
\label{app:hyperparams}

We perform a stage-wise hyperparameter optimization for ReHoPER.
We first tune the inference temperature \(T \in \{0.2,\,0.3,\,0.5,\,0.7\}\) while fixing the path/step configuration to \(10-6\).
After selecting the best temperature on a development split of iLLC~L2-4 with 250 newly sampled instances, we fix this temperature and tune the path/step budget \(\{n\}-\{m_{\max}\} \in \{5-12,\ 6-10,\ 10-6,\ 12-5\}\).
All configurations are budget-matched to produce at most 61 candidate answers per instance.
Finally, we rerun the selected configuration on the MoreHopQA dataset to assess whether the same trends generalize across datasets.
Tables~\ref{tab:temp} and~\ref{tab:pathstep} summarize the results.

Based on Table~\ref{tab:temp}, we fix the sampling temperature to \(T=0.5\). On iLLC~L2-4, \(T=0.5\) yields the highest average accuracy across all three answer-format prefixes, improving over the best lower temperatures (\(T\in\{0.2,0.3\}\)) by roughly 3pp for IO and CoT and about 1pp for PS, and still outperforming the higher temperature \(T=0.7\) by around 1-2pp on IO/CoT and about 0.5pp on PS. The same trend largely holds on MoreHopQA: \(T=0.5\) remains the best choice for IO and PS, and is the second best setting for CoT, indicating that this temperature transfers well across datasets. For the path/step budget in Table~\ref{tab:pathstep}, we select the \(6-10\) configuration. On iLLC~L2-4, \(6-10\) achieves the best average performance for PS, is tied for best for CoT, and is second-best for IO. On MoreHopQA, \(6-10\) is again consistently competitive and typically within 1pp of the best configuration across IO, CoT, and PS, even though \(10-6\) occasionally attains the highest score. We therefore adopt \(T=0.5\) and a \(6-10\) path/step budget as a single, dataset-agnostic configuration for all subsequent experiments.

\input{tables/temp}

\input{tables/pathstep}

\clearpage
\section{Robustness across repeated runs}
\label{app:robustness}

To assess whether ReHoPER's gains are robust under stochastic decoding, we repeat each experiment three times and report the mean accuracy with standard deviation in Table~\ref{tab:robustness}. Overall, ReHoPER remains stable across runs, with standard deviations generally comparable to those of Self-Consistency (SC). This suggests that the observed improvements are not artifacts of a single favorable sample, but persist across independent decoding runs.

The gains are especially pronounced on the harder iLLC L2-4 benchmark. Averaged across models, ReHoPER improves over SC under all answer-format prefixes, increasing accuracy from 8.6 to 22.3 with IO, 8.1 to 24.2 with CoT, and 11.7 to 30.8 with PS. These improvements are consistent across most models and are particularly large for Phi-4, where ReHoPER improves from roughly 10--13\% accuracy to more than 54--62\% across prefixes, while maintaining low run-to-run variation.

On MoreHopQA, where the base SC performance is already substantially higher, the improvements are smaller but still generally positive on average. ReHoPER matches or improves the average SC result under all three prefixes, with the clearest gains under CoT and PS. While a few individual model-prefix combinations show small decreases, the overall pattern indicates that ReHoPER does not rely on a fragile prompting choice and its benefits are strongest in the more compositional iLLC setting and remain stable across repeated runs and answer-format prefixes.

\input{tables/robustness}

\section{Planning ablation details}
\label{app:planning-ablations}

The planning ablations retain ReHoPER's initial prediction, Answer, Update, Predict, and Aggregation steps (Section~\ref{sec:rehoper}), with the same decoding temperature, output prefixes, number of paths, and maximum step count. They change question generation, selection, and path termination as follows.

\paragraph{Without replanning.}
At the start of each path \(i\), a single call \(f(C,I_{\mathrm{fixed}})\) generates an ordered list \(\mathbf{IQ}^{(i)}=\langle IQ^{(i)}_1,\ldots,IQ^{(i)}_{m_i}\rangle\), whose length is determined by the model and can vary across paths. At step \(k\), the Plan and Select steps are replaced by indexing this list. Questions are answered in order without revising the list or applying ReHoPER's candidate-selection rule. The path ends after \(\min(m_i,m_{\max})\) steps.

\paragraph{Without look-ahead.}
At each step, a single call \(f(C,\mathbf{H}^{(i)}_{k-1},I_{\mathrm{single}})\) generates the next intermediate question. This replaces both Plan and Select: there is no candidate horizon or separate check for selecting an unasked question. The path runs to \(m_{\max}\), even if a question repeats. The prompt nevertheless instructs the model to avoid repetition.

The exact instructions are shown in Figures~\ref{fig:i-fixed} and~\ref{fig:i-single}. Because these configurations also change selection and termination, the comparisons assess the combined planning design rather than isolate a single mechanism with all other behavior held fixed. Equal path and step limits do not imply equal realized trajectory lengths or compute use. Table~\ref{tab:planning-ablations-full} in the main paper reports the complete accuracy breakdown.

\clearpage
\section{iLLC L1-8 and iLLC L2-8 results}
\label{app:illc}
Table \ref{tab:illc} reports the full per-model accuracy breakdown on iLLC L1-8 and iLLC L2-8 under IO, CoT, and PS prefixes.

\input{tables/illc}

\section{Model details}
\label{app:models}

Table~\ref{tab:models} summarizes the language models used in our experiments, including their model families, parameter scales, context lengths, and the exact checkpoints used for inference. Our model suite is chosen to cover a deliberately diverse range of instruction-tuned \gls*{llm}s: smaller dense models such as \textsc{Llama-3.1-8B-Instruct} \citep{grattafiori2024llama}, mid-sized dense models such as \textsc{Gemma-3-12B-IT} \citep{team2025gemma}, \textsc{Phi-4} \citep{abdin2024phi}, and \textsc{Mistral-Small-24B-Instruct-2501} \citep{mistral_small_24b_instruct_2501_modelcard}, as well as larger-capacity models obtained either through scale, such as \textsc{Llama-3.1-70B}, or sparsity, such as \textsc{gpt-oss-20b} \citep{agarwal2025gpt} and \textsc{Qwen3-30B-A3B} \citep{yang2025qwen3}. This selection covers both general instruction-following models and models explicitly positioned toward stronger reasoning behavior.

From a systems perspective, we use quantized deployments for larger models when needed to reduce inference memory requirements. Specifically, we use 4-bit weight-only AWQ variants \citep{lin2024awq} for Llama 3.1 70B and Mistral Small, and a GPTQ variant \citep{frantar2023optq} for Qwen3. For Qwen3, we report two configurations: thinking disabled and thinking enabled, using the model's built-in switching mechanism; unless otherwise specified, Qwen3 refers to the non-thinking configuration. We implement our experiments in PyTorch \citep{paszke2019pytorch} and perform inference using vLLM \citep{kwon2023efficient}, together with the transformers \citep{wolf2019huggingface}, NumPy \citep{harris2020array}, and Matplotlib \citep{Hunter:2007} libraries. Experiments were conducted on a multi-GPU setup equipped with NVIDIA RTX A6000 48GB and NVIDIA RTX A5000 24GB GPUs.

\begin{table*}[htb]
\centering
\footnotesize
\caption{Models used in our experiments. We list the exact checkpoint, model family, parameter scale, and context length reported by the corresponding model card.}
\label{tab:models}
\setlength{\tabcolsep}{5pt}
\resizebox{1.\columnwidth}{!}{
\begin{tabular}{l l l l}
\toprule
\textbf{Model (as used)} & \textbf{Org / Family} & \textbf{Params} & \textbf{Context} \\
\midrule
google/gemma-3-12b-it & Google / Gemma 3 & 12B & 128K \\
openai/gpt-oss-20b & OpenAI / gpt-oss & 21B (3.6B active) & 128K \\
meta-llama/Llama-3.1-8B-Instruct & Meta / Llama 3.1 & 8B & 128K \\
hugging-quants/Meta-Llama-3.1-70B-Instruct-AWQ-INT4 & Meta / Llama 3.1 & 70B & 128K \\
stelterlab/Mistral-Small-24B-Instruct-2501-AWQ & Mistral / Mistral Small 3 & 24B & 32K \\
microsoft/phi-4 & Microsoft / Phi & 14B & 16K \\
Qwen/Qwen3-30B-A3B-GPTQ-Int4 & Qwen / Qwen3 & 30.5B (3.3B active) & 32K \\
\bottomrule
\end{tabular}
}
\end{table*}

\clearpage
\section{Intermediate question instructions}
\label{app:instruct}

\begin{figure}[htb]
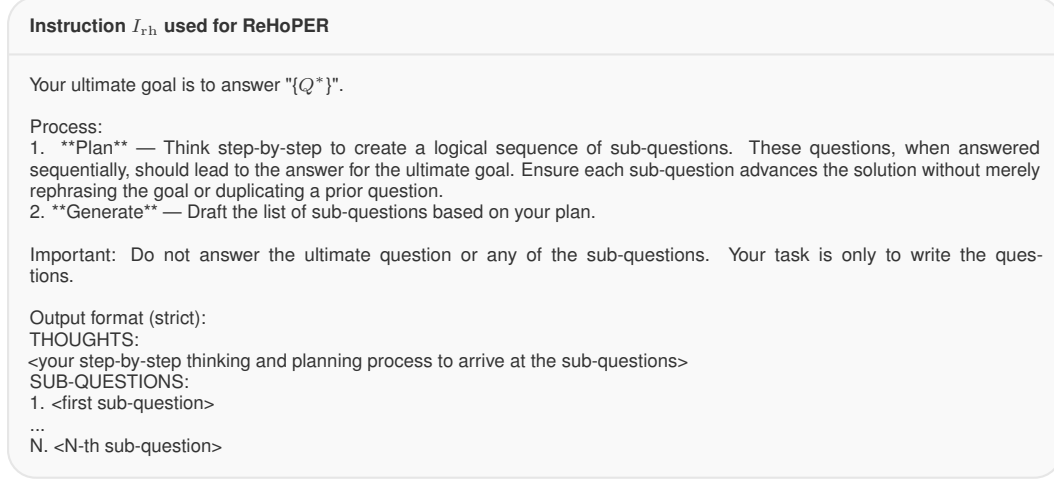

\centering
\begin{softbox}[Instruction \(I_{\mathrm{rh}}\) used for ReHoPER]
Your ultimate goal is to answer "\{\(Q^*\)\}".\\
\\
Process:\\
1. **Plan** — Think step-by-step to create a logical sequence of sub-questions. These questions, when answered sequentially, should lead to the answer for the ultimate goal. Ensure each sub-question advances the solution without merely rephrasing the goal or duplicating a prior question.\\
2. **Generate** — Draft the list of sub-questions based on your plan.\\
\\
Important: Do not answer the ultimate question or any of the sub-questions. Your task is only to write the questions.\\
\\
Output format (strict):\\
THOUGHTS:\\
<your step-by-step thinking and planning process to arrive at the sub-questions>\\
SUB-QUESTIONS:\\
1. <first sub-question>\\
...\\
N. <N-th sub-question>
\end{softbox}
\caption{Instruction \(I_{\mathrm{rh}}\) used for ReHoPER.}
\label{fig:i-rh}
\end{figure}

\begin{figure}[htb]
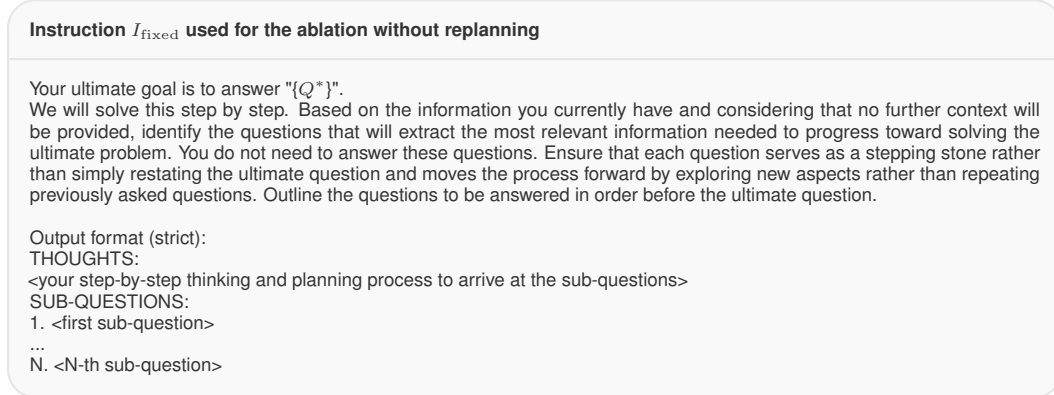

\centering
\begin{softbox}[Instruction \(I_{\mathrm{fixed}}\) used for the ablation without replanning]
Your ultimate goal is to answer "\{\(Q^*\)\}".\\
We will solve this step by step. Based on the information you currently have and considering that no further context will be provided, identify the questions that will extract the most relevant information needed to progress toward solving the ultimate problem. You do not need to answer these questions. Ensure that each question serves as a stepping stone rather than simply restating the ultimate question and moves the process forward by exploring new aspects rather than repeating previously asked questions. Outline the questions to be answered in order before the ultimate question.\\
\\
Output format (strict):\\
THOUGHTS:\\
<your step-by-step thinking and planning process to arrive at the sub-questions>\\
SUB-QUESTIONS:\\
1. <first sub-question>\\
...\\
N. <N-th sub-question>
\end{softbox}
\caption{Instruction \(I_{\mathrm{fixed}}\) used for the ablation without replanning.}
\label{fig:i-fixed}
\end{figure}

\begin{figure}[htb]
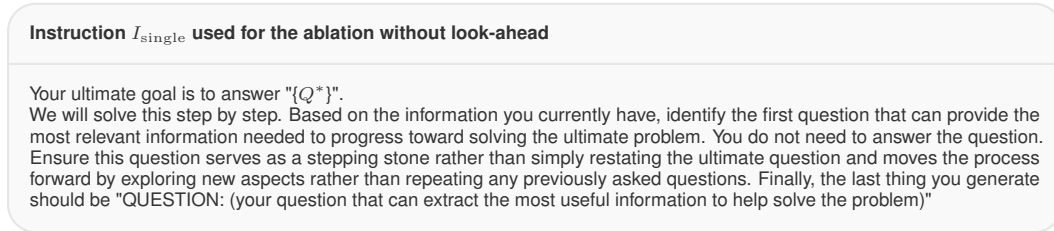

\centering
\begin{softbox}[Instruction \(I_{\mathrm{single}}\) used for the ablation without look-ahead]
Your ultimate goal is to answer "\{\(Q^*\)\}".\\
We will solve this step by step. Based on the information you currently have, identify the first question that can provide the most relevant information needed to progress toward solving the ultimate problem. You do not need to answer the question. Ensure this question serves as a stepping stone rather than simply restating the ultimate question and moves the process forward by exploring new aspects rather than repeating any previously asked questions. Finally, the last thing you generate should be "QUESTION: (your question that can extract the most useful information to help solve the problem)"
\end{softbox}
\caption{Instruction \(I_{\mathrm{single}}\) used for the ablation without look-ahead.}
\label{fig:i-single}
\end{figure}

This appendix summarizes the exact intermediate question generation instructions used in our experiments. Figures~\ref{fig:i-rh}, \ref{fig:i-fixed}, and~\ref{fig:i-single} show the prompts for ReHoPER (\(I_{\mathrm{rh}}\)), the ablation without replanning (\(I_{\mathrm{fixed}}\)), and the ablation without look-ahead (\(I_{\mathrm{single}}\)), respectively.

\clearpage
\section{Dataset details and examples}
\label{app:data-examples}

\paragraph{MuSR.}
MuSR evaluates multi-step reasoning over long, story-like inputs where the answer requires integrating scattered clues and applying commonsense inference rather than relying on surface matching~\citep{sprague2024musr}. It contains three tasks: Murder Mysteries (MM), where the model identifies the culprit from a narrative; Object Placement (OP), where the model tracks objects as they move or are relocated through a sequence of events; and Team Allocation (TA), where the model infers valid group assignments under narrative constraints.

\paragraph{MoreHopQA.}
MoreHopQA is a multi-hop question answering benchmark designed to test whether models can connect multiple evidence pieces across a context and perform sequential reasoning hops to reach the correct answer~\citep{Schnitzler2024MoreHopQAMT}. We use the 150-example subset selected by the dataset authors, following the original evaluation setup.

\paragraph{iLLC.}
iLLC is our controlled diagnostic benchmark for compositional reasoning. It generalizes Last Letter Concatenation (LLC)~\citep{wei2022chain} by requiring the model to extract the \(i\)-th-to-last character from each word in a given list and concatenate the resulting letters. We denote subsets as L\{\(i\)\}-\{\(w\)\}, where \(i\) specifies the target character position from the end and \(w\) specifies the number of words in the input. For example, L1-4 corresponds to standard last-letter concatenation over four words, while L2-4 requires extracting the second-to-last character from each of four words.

\paragraph{Post-processing.}
Let \(R\) denote the model response to the goal question \(Q^*\), and let \(\hat{y}\) denote the extracted prediction. For MoreHopQA and iLLC, we obtain \(\hat{y}\) by extracting the first word that appears immediately after \texttt{"ANSWER:"}. For MuSR, where answers are multiple-choice option numbers, we instead extract the most frequent option number that appears after \texttt{"ANSWER:"} in \(R\).

Table~\ref{tab:datasets} summarizes the datasets used in our evaluation, including their task variants, test-set sizes, and answer formats. The examples below illustrate MuSR (MM/OP/TA), MoreHopQA, and iLLC (L1-4/L2-4).

\begin{table*}[htb]
\centering
\small
\caption{Summary of reasoning datasets used for evaluation.}
\label{tab:datasets}
\resizebox{.8\linewidth}{!}{
\begin{tabular}{l l c l}
\toprule
\multirow{2}{*}{\textbf{Dataset}} & \multirow{2}{*}{\textbf{Task}} & \textbf{Test Set Size} & \multirow{2}{*}{\textbf{Answer Format}} \\
& & \textbf{Per Task} & \\
\midrule
\multirow{2}{*}{MuSR} & Murder Mysteries, Team Allocation & 250 & \multirow{2}{*}{Choice number} \\
 & Object Placement & 256 & \\

\cdashlinelr{1-4}

MoreHopQA &
- &
150 &
Short text answer \\

\cdashlinelr{1-4}

iLLC & L1-4, L1-6, L1-8, L2-4, L2-6, L2-8 & 250 & Concatenated letters \\

\bottomrule
\end{tabular}
}
\end{table*}

\input{figures/data-examples}

\clearpage

\section{Sample iLLC L2-4 dialogues}
\label{app:output-samples}
The positive and negative examples below illustrate two iLLC L2-4 instances, showing intermediate questions, answers, and final predictions from our ReHoPER experiments.

\vspace{.5em}
\begin{softbox}[A positive iLLC L2-4 example using ReHoPER]
\textbf{User:} (\(IQ_1\))\\
What are the individual words in the name "Zoe Brock Clark Sanders"? \\
\\
\textbf{Assistant:} (\(A_1\))\\
The individual words in the name "Zoe Brock Clark Sanders" are:\\\\1. Zoe\\2. Brock\\3. Clark\\4. Sanders \\
\\
\textbf{User:} (\(IQ_2\))\\
What is the second-to-last letter of the word "Zoe"? \\
\\
\textbf{Assistant:} (\(A_2\))\\
The word "Zoe" has three letters: Z, o, and e. The second-to-last letter is "o". \\
\\
\textbf{User:} (\(IQ_3\))\\
What is the second-to-last letter of the word "Brock"? \\
\\
\textbf{Assistant:} (\(A_3\))\\
The word "Brock" has five letters: B, r, o, c, and k. The second-to-last letter is "c". \\
\\
\textbf{User:} (\(IQ_4\))\\
What is the second-to-last letter of the word "Clark"? \\
\\
\textbf{Assistant:} (\(A_4\))\\
The word "Clark" has five letters: C, l, a, r, and k. The second-to-last letter is "r". \\
\\
\textbf{User:} (\(IQ_5\))\\
What is the second-to-last letter of the word "Sanders"? \\
\\
\textbf{Assistant:} (\(A_5\))\\
The word "Sanders" has seven letters: S, a, n, d, e, r, and s. The second-to-last letter is "r". \\
\\
\textbf{User:} (\(IQ_6\))\\
What is the result of concatenating the second-to-last letters of the words "Zoe", "Brock", "Clark", and "Sanders"? \\
\\
\textbf{Assistant:} (\(A_6\))\\
The second-to-last letters of the words are as follows:\\- "Zoe": o\\- "Brock": c\\- "Clark": r\\- "Sanders": r\\\\Concatenating these letters results in the string "ocrr". \\
\\
\textbf{User:} (\(IQ^*\))\\
Question: Take the second-to-last letters of the words in “Zoe Brock Clark Sanders” and concatenate them.\\The last thing you generate should be "ANSWER: (the concatenated letters without any spaces or special characters)" \\
\\
\textbf{Assistant:} (\(R_6\))\\
Let's think step by step. We need to identify the second-to-last letter of each word in "Zoe Brock Clark Sanders" and then concatenate them.\\\\1. **Zoe**: The second-to-last letter is "o".\\2. **Brock**: The second-to-last letter is "c".\\3. **Clark**: The second-to-last letter is "r".\\4. **Sanders**: The second-to-last letter is "r".\\\\Now, concatenate these letters: "ocrr".\\\\ANSWER: ocrr \\
\\
\noindent\rule[0.5pt]{\linewidth}{0.5pt}\\
\\
\textbf{Ground Truth:} (\(y\)) \\
ocrr

\end{softbox}
\vspace{.5em}

\begin{softbox}[A negative iLLC L2-4 example using ReHoPER]
\textbf{User:} (\(IQ_1\))\\
What are the individual words in the phrase "Dan Rudy Kash Yolanda"? \\
\\
\textbf{Assistant:} (\(A_1\))\\
The phrase "Dan Rudy Kash Yolanda" is composed of the following individual words:\\\\1. Dan\\2. Rudy\\3. Kash\\4. Yolanda\\\\Each of these appears to be a proper noun, likely representing names. \\
\\
\textbf{User:} (\(IQ_2\))\\
What is the length of each word in the phrase "Dan Rudy Kash Yolanda"? \\
\\
\textbf{Assistant:} (\(A_2\))\\
Here are the lengths of each word in the phrase "Dan Rudy Kash Yolanda":\\\\1. Dan - 3 letters\\2. Rudy - 4 letters\\3. Kash - 4 letters\\4. Yolanda - 7 letters\\\\The total number of letters in the phrase is 18. \\
\\
\textbf{User:} (\(IQ_3\))\\
How many letters are in each word: Dan, Rudy, Kash, Yolanda? \\
\\
\textbf{Assistant:} (\(A_3\))\\
Here are the number of letters in each word:\\\\1. Dan - 3 letters\\2. Rudy - 4 letters\\3. Kash - 4 letters\\4. Yolanda - 7 letters\\\\In total, there are 18 letters in the phrase "Dan Rudy Kash Yolanda." \\
\\
\textbf{User:} (\(IQ_4\))\\
How many letters are in the word "Dan"? \\
\\
\textbf{Assistant:} (\(A_4\))\\
The word "Dan" has 3 letters. \\
\\
\textbf{User:} (\(IQ_5\))\\
What is the second-to-last letter of the word "Dan"? \\
\\
\textbf{Assistant:} (\(A_5\))\\
The second-to-last letter of the word "Dan" is "a". \\
\\
\textbf{User:} (\(IQ_6\))\\
What is the length of the word "Dan"? \\
\\
\textbf{Assistant:} (\(A_6\))\\
The word "Dan" has a length of 3 letters. \\
\\
\textbf{User:} (\(IQ_7\))\\
What is the second-to-last letter of the word "Rudy"? \\
\\
\textbf{Assistant:} (\(A_7\))\\
The second-to-last letter of the word "Rudy" is "d". \\
\\
\textbf{User:} (\(IQ_8\))\\
What is the second-to-last letter of the word "Kash"? \\
\\
\textbf{Assistant:} (\(A_8\))\\
The second-to-last letter of the word "Kash" is "a". \\
\\
\textbf{User:} (\(IQ_9\))\\
What is the second-to-last letter of the word "Yolanda"? \\
\\
\textbf{Assistant:} (\(A_9\))\\
The second-to-last letter of the word "Yolanda" is "n". \\
\\
\textbf{User:} (\(IQ_{10}\))\\
Concatenate the second-to-last letters of the words "Dan", "Rudy", "Kash", and "Yolanda" in that order. \\
\\
\textbf{Assistant:} (\(A_{10}\))\\
The second-to-last letters of the words "Dan", "Rudy", "Kash", and "Yolanda" are "a", "d", "a", and "n", respectively. Concatenating these letters in that order gives you "adad". \\
\\
\textbf{User:} (\(IQ^*\))\\
Question: Take the second-to-last letters of the words in “Dan Rudy Kash Yolanda” and concatenate them.\\The last thing you generate should be "ANSWER: (the concatenated letters without any spaces or special characters)" \\
\\
\textbf{Assistant:} (\(R_{10}\))\\
Let's think step by step. The phrase is "Dan Rudy Kash Yolanda". We will determine the second-to-last letter of each word:\\\\1. "Dan" - The second-to-last letter is "a".\\2. "Rudy" - The second-to-last letter is "d".\\3. "Kash" - The second-to-last letter is "a".\\4. "Yolanda" - The second-to-last letter is "n".\\\\Now, we concatenate these letters in the order they appear: "a", "d", "a", "n".\\\\ANSWER: adan \\
\\
\noindent\rule[0.5pt]{\linewidth}{0.5pt}\\
\\
\textbf{Ground Truth:} (\(y\)) \\
adsd

\end{softbox}
\vspace{.5em}

%% file: tables/result-macro.tex
\begin{table*}[htb]
\centering
\small
\caption{Macro-aggregated accuracy (\%) on three benchmark families. MuSR averages MM, OP, and TA; iLLC averages L1-4, L1-6, L1-8, L2-4, L2-6, and L2-8; and MoreHopQA is reported separately. The final block is an equal-weight average of these three family-level results. SD denotes Self-Discover. \colorbox{green!40}{\small Green}, \colorbox{red!25}{\small red}, and \colorbox{gray!25}{\small gray} indicate improvements, decreases, and ties of ReHoPER (RH) relative to Self-Consistency (SC), respectively.}
\label{tab:result-macro}
\resizebox{0.67\textwidth}{!}{\begin{tabular}{l p{0pt} c cc cc cc p{5pt} c cc cc cc}
\toprule
&& \multicolumn{7}{c}{\textbf{MuSR}} && \multicolumn{7}{c}{\textbf{iLLC}} \\
\cmidrule(lr){3-9}\cmidrule(lr){11-17}
&& & \multicolumn{2}{c}{\textbf{IO}} & \multicolumn{2}{c}{\textbf{CoT}} & \multicolumn{2}{c}{\textbf{PS}} && & \multicolumn{2}{c}{\textbf{IO}} & \multicolumn{2}{c}{\textbf{CoT}} & \multicolumn{2}{c}{\textbf{PS}} \\
\cmidrule(lr){4-5}\cmidrule(lr){6-7}\cmidrule(lr){8-9}\cmidrule(lr){12-13}\cmidrule(lr){14-15}\cmidrule(lr){16-17}
\textbf{Model} && \textbf{SD}&\textbf{SC}&\textbf{RH}&\textbf{SC}&\textbf{RH}&\textbf{SC}&\textbf{RH} && \textbf{SD}&\textbf{SC}&\textbf{RH}&\textbf{SC}&\textbf{RH}&\textbf{SC}&\textbf{RH} \\
\midrule
Gemma 3 && 52.5&54.5&\gain{54.6}&56.3&\loss{56.2}&58.1&\gain{59.2} && 24.7&7.8&\gain{16.6}&30.5&\loss{30.5}&37.1&\loss{32.5} \\
gpt-oss && 66.5&65.2&\gain{66.0}&60.0&\gain{64.4}&65.1&\gain{66.1} && 62.2&85.9&\loss{83.9}&78.3&\gain{83.4}&52.7&\gain{83.9} \\
Llama 3.1 8B && 44.4&47.1&\gain{53.0}&50.5&\gain{52.3}&51.5&\gain{53.2} && 12.2&28.3&\loss{26.9}&26.5&\gain{31.7}&35.0&\gain{37.1} \\
Llama 3.1 70B && 55.0&57.9&\gain{61.6}&58.0&\gain{61.4}&58.5&\gain{62.1} && 39.9&40.3&\gain{69.9}&44.5&\gain{69.9}&56.6&\gain{77.3} \\
Mistral Small && 53.1&53.1&\gain{59.8}&55.5&\gain{63.3}&58.0&\gain{62.8} && 22.9&38.3&\gain{38.3}&38.9&\gain{39.3}&39.1&\gain{40.6} \\
Phi-4 && 57.4&61.3&\loss{60.4}&59.5&\gain{62.5}&61.1&\gain{63.3} && 38.7&38.2&\gain{52.9}&40.8&\gain{61.7}&40.8&\gain{61.3} \\
Qwen3 w/o Thinking && 53.3&56.2&\gain{62.5}&63.1&\gain{65.0}&63.2&\gain{64.4} && 29.7&53.2&\gain{68.1}&48.3&\gain{68.3}&74.7&\loss{72.4} \\
Qwen3 w/ Thinking && 63.2&65.2&\gain{65.4}&63.1&\gain{65.2}&63.2&\gain{64.8} && 1.6&87.3&\loss{86.4}&48.3&\gain{84.9}&74.7&\gain{85.6} \\
\cdashlinelr{1-17}
Average && 55.7&57.6&\gain{60.4}&58.2&\gain{61.3}&59.8&\gain{62.0} && 29.0&47.4&\gain{55.4}&44.5&\gain{58.7}&51.3&\gain{61.3} \\

\midrule

&& \multicolumn{7}{c}{\textbf{MoreHopQA}} && \multicolumn{7}{c}{\textbf{Average}} \\
\cmidrule(lr){3-9}\cmidrule(lr){11-17}
&& & \multicolumn{2}{c}{\textbf{IO}} & \multicolumn{2}{c}{\textbf{CoT}} & \multicolumn{2}{c}{\textbf{PS}} && & \multicolumn{2}{c}{\textbf{IO}} & \multicolumn{2}{c}{\textbf{CoT}} & \multicolumn{2}{c}{\textbf{PS}} \\
\cmidrule(lr){4-5}\cmidrule(lr){6-7}\cmidrule(lr){8-9}\cmidrule(lr){12-13}\cmidrule(lr){14-15}\cmidrule(lr){16-17}
\textbf{Model} && \textbf{SD}&\textbf{SC}&\textbf{RH}&\textbf{SC}&\textbf{RH}&\textbf{SC}&\textbf{RH} && \textbf{SD}&\textbf{SC}&\textbf{RH}&\textbf{SC}&\textbf{RH}&\textbf{SC}&\textbf{RH} \\
\midrule
Gemma 3 && 44.0&61.3&\gain{64.7}&64.7&\gain{68.7}&66.7&\gain{74.0} && 40.4&41.2&\gain{45.3}&50.5&\gain{51.8}&54.0&\gain{55.2} \\
gpt-oss && 39.3&80.7&\loss{78.7}&81.3&\loss{80.0}&79.3&\gain{81.3} && 56.0&77.2&\loss{76.2}&73.2&\gain{75.9}&65.7&\gain{77.1} \\
Llama 3.1 8B && 34.0&59.3&\loss{57.3}&61.3&\loss{60.7}&64.0&\loss{63.3} && 30.2&44.9&\gain{45.7}&46.1&\gain{48.2}&50.2&\gain{51.2} \\
Llama 3.1 70B && 62.0&68.0&\gain{71.3}&68.0&\gain{71.3}&68.7&\gain{74.0} && 52.3&55.4&\gain{67.6}&56.8&\gain{67.6}&61.3&\gain{71.1} \\
Mistral Small && 54.7&70.0&\gain{71.3}&70.0&\same{70.0}&71.3&\gain{72.7} && 43.5&53.8&\gain{56.5}&54.8&\gain{57.6}&56.2&\gain{58.7} \\
Phi-4 && 64.0&77.3&\same{77.3}&76.0&\gain{80.7}&77.3&\loss{76.7} && 53.4&58.9&\gain{63.5}&58.8&\gain{68.3}&59.7&\gain{67.1} \\
Qwen3 w/o Thinking && 34.0&70.0&\loss{68.0}&69.3&\gain{70.7}&73.3&\loss{72.0} && 39.0&59.8&\gain{66.2}&60.2&\gain{68.0}&70.4&\loss{69.6} \\
Qwen3 w/ Thinking && 42.0&75.3&\gain{80.0}&69.3&\gain{78.7}&73.3&\gain{78.0} && 35.6&76.0&\gain{77.3}&60.2&\gain{76.3}&70.4&\gain{76.1} \\
\cdashlinelr{1-17}
Average && 46.8&70.3&\gain{71.1}&70.0&\gain{72.6}&71.8&\gain{74.0} && 43.8&58.4&\gain{62.3}&57.6&\gain{64.2}&61.0&\gain{65.8} \\
\bottomrule
\end{tabular}}
\end{table*}

%% file: tables/iq-statistics.tex
\begin{table*}[htb]
\centering
\caption{
Statistics of generated intermediate questions (IQs) in \textbf{ReHoPER} on \textbf{MuSR} and \textbf{MoreHopQA}. We report the total number of generated IQs, the average number of IQs per path with standard deviation, the number of unique IQs, the average number of unique IQs per path with standard deviation, and the corresponding uniqueness percentages.
}
\label{tab:iq-statistics-rh-1}
\resizebox{.9\textwidth}{!}{
\begin{tabular}{
l p{0pt}
cc p{0pt} cc p{0pt} cc
}

\toprule

& &
\multicolumn{8}{c}{\textbf{MuSR MM}} \\

\cmidrule(lr){3-10}

\textbf{Model} & &
\textbf{\#IQ} & \textbf{\#IQ/Path$_{\mathrm{std}}$} & & \textbf{\#uniqIQ} & \textbf{\#uniqIQ/Path$_{\mathrm{std}}$} & & \textbf{\%uniqIQ} & \textbf{\%uniqIQ/Path} \\

\midrule
Gemma 3 & & 60.00 & $10.00_{\mathrm{0.00}}$ & & 59.29 & $10.00_{\mathrm{0.00}}$ & & 98.82\% & 100.00\% \\
gpt-oss & & 39.38 & $6.56_{\mathrm{3.72}}$ & & 39.29 & $6.56_{\mathrm{3.72}}$ & & 99.77\% & 100.00\% \\
Llama 3.1 8B & & 59.95 & $9.99_{\mathrm{0.02}}$ & & 59.54 & $9.99_{\mathrm{0.02}}$ & & 99.32\% & 100.00\% \\
Llama 3.1 70B & & 60.00 & $10.00_{\mathrm{0.00}}$ & & 56.72 & $10.00_{\mathrm{0.00}}$ & & 94.53\% & 100.00\% \\
Mistral Small & & 59.96 & $9.99_{\mathrm{0.01}}$ & & 55.86 & $9.99_{\mathrm{0.01}}$ & & 93.16\% & 100.00\% \\
Phi-4 & & 59.97 & $10.00_{\mathrm{0.01}}$ & & 57.08 & $10.00_{\mathrm{0.01}}$ & & 95.17\% & 100.00\% \\
Qwen3 & & 59.56 & $9.93_{\mathrm{0.15}}$ & & 57.24 & $9.93_{\mathrm{0.15}}$ & & 96.12\% & 100.00\% \\
\cdashlinelr{1-10}
Average & & 56.97 & $9.50_{\mathrm{0.56}}$ & & 55.00 & $9.50_{\mathrm{0.56}}$ & & 96.70\% & 100.00\% \\

\midrule

& &
\multicolumn{8}{c}{\textbf{MuSR OP}} \\

\cmidrule(lr){3-10}

\textbf{Model} & &
\textbf{\#IQ} & \textbf{\#IQ/Path$_{\mathrm{std}}$} & & \textbf{\#uniqIQ} & \textbf{\#uniqIQ/Path$_{\mathrm{std}}$} & & \textbf{\%uniqIQ} & \textbf{\%uniqIQ/Path} \\

\midrule
Gemma 3 & & 59.98 & $10.00_{\mathrm{0.01}}$ & & 52.47 & $10.00_{\mathrm{0.01}}$ & & 87.48\% & 100.00\% \\
gpt-oss & & 59.42 & $9.90_{\mathrm{0.22}}$ & & 57.66 & $9.90_{\mathrm{0.22}}$ & & 97.04\% & 100.00\% \\
Llama 3.1 8B & & 59.27 & $9.88_{\mathrm{0.25}}$ & & 57.52 & $9.88_{\mathrm{0.25}}$ & & 97.05\% & 100.00\% \\
Llama 3.1 70B & & 59.99 & $10.00_{\mathrm{0.00}}$ & & 51.86 & $10.00_{\mathrm{0.00}}$ & & 86.45\% & 100.00\% \\
Mistral Small & & 55.77 & $9.29_{\mathrm{0.85}}$ & & 46.51 & $9.29_{\mathrm{0.85}}$ & & 83.39\% & 100.00\% \\
Phi-4 & & 59.84 & $9.97_{\mathrm{0.05}}$ & & 55.41 & $9.97_{\mathrm{0.05}}$ & & 92.60\% & 100.00\% \\
Qwen3 & & 55.66 & $9.28_{\mathrm{0.84}}$ & & 48.01 & $9.28_{\mathrm{0.84}}$ & & 86.26\% & 100.00\% \\
\cdashlinelr{1-10}
Average & & 58.56 & $9.76_{\mathrm{0.32}}$ & & 52.78 & $9.76_{\mathrm{0.32}}$ & & 90.04\% & 100.00\% \\

\midrule

& &
\multicolumn{8}{c}{\textbf{MuSR TA}} \\

\cmidrule(lr){3-10}

\textbf{Model} & &
\textbf{\#IQ} & \textbf{\#IQ/Path$_{\mathrm{std}}$} & & \textbf{\#uniqIQ} & \textbf{\#uniqIQ/Path$_{\mathrm{std}}$} & & \textbf{\%uniqIQ} & \textbf{\%uniqIQ/Path} \\

\midrule
Gemma 3 & & 60.00 & $10.00_{\mathrm{0.00}}$ & & 59.88 & $10.00_{\mathrm{0.00}}$ & & 99.81\% & 100.00\% \\
gpt-oss & & 55.38 & $9.23_{\mathrm{1.33}}$ & & 54.56 & $9.23_{\mathrm{1.33}}$ & & 98.52\% & 100.00\% \\
Llama 3.1 8B & & 59.67 & $9.94_{\mathrm{0.12}}$ & & 58.76 & $9.94_{\mathrm{0.12}}$ & & 98.48\% & 100.00\% \\
Llama 3.1 70B & & 59.99 & $10.00_{\mathrm{0.00}}$ & & 55.94 & $10.00_{\mathrm{0.00}}$ & & 93.25\% & 100.00\% \\
Mistral Small & & 60.00 & $10.00_{\mathrm{0.00}}$ & & 55.43 & $10.00_{\mathrm{0.00}}$ & & 92.38\% & 100.00\% \\
Phi-4 & & 60.00 & $10.00_{\mathrm{0.00}}$ & & 46.76 & $10.00_{\mathrm{0.00}}$ & & 77.94\% & 100.00\% \\
Qwen3 & & 59.98 & $10.00_{\mathrm{0.01}}$ & & 54.42 & $10.00_{\mathrm{0.01}}$ & & 90.72\% & 100.00\% \\
\cdashlinelr{1-10}
Average & & 59.29 & $9.88_{\mathrm{0.21}}$ & & 55.11 & $9.88_{\mathrm{0.21}}$ & & 93.01\% & 100.00\% \\

\midrule

& &
\multicolumn{8}{c}{\textbf{MoreHopQA}} \\

\cmidrule(lr){3-10}

\textbf{Model} & &
\textbf{\#IQ} & \textbf{\#IQ/Path$_{\mathrm{std}}$} & & \textbf{\#uniqIQ} & \textbf{\#uniqIQ/Path$_{\mathrm{std}}$} & & \textbf{\%uniqIQ} & \textbf{\%uniqIQ/Path} \\

\midrule
Gemma 3 & & 43.25 & $7.21_{\mathrm{1.39}}$ & & 25.79 & $7.21_{\mathrm{1.39}}$ & & 59.64\% & 100.00\% \\
gpt-oss & & 54.83 & $9.14_{\mathrm{0.91}}$ & & 37.68 & $9.14_{\mathrm{0.91}}$ & & 68.72\% & 100.00\% \\
Llama 3.1 8B & & 39.47 & $6.58_{\mathrm{2.33}}$ & & 29.14 & $6.58_{\mathrm{2.33}}$ & & 73.83\% & 100.00\% \\
Llama 3.1 70B & & 48.99 & $8.17_{\mathrm{0.89}}$ & & 22.33 & $8.17_{\mathrm{0.89}}$ & & 45.57\% & 100.00\% \\
Mistral Small & & 33.12 & $5.52_{\mathrm{0.99}}$ & & 13.12 & $5.52_{\mathrm{0.99}}$ & & 39.61\% & 100.00\% \\
Phi-4 & & 46.07 & $7.68_{\mathrm{1.08}}$ & & 24.90 & $7.68_{\mathrm{1.08}}$ & & 54.04\% & 100.00\% \\
Qwen3 & & 35.76 & $5.96_{\mathrm{1.22}}$ & & 16.77 & $5.96_{\mathrm{1.22}}$ & & 46.89\% & 100.00\% \\
\cdashlinelr{1-10}
Average & & 43.07 & $7.18_{\mathrm{1.26}}$ & & 24.25 & $7.18_{\mathrm{1.26}}$ & & 55.47\% & 100.00\% \\

\bottomrule

\end{tabular}
}
\vspace{-1.1em}
\end{table*}

\begin{table*}[htb]
\centering
\caption{
Statistics of generated intermediate questions (IQs) in \textbf{ReHoPER} on \textbf{iLLC}. We report the total number of generated IQs, the average number of IQs per path with standard deviation, the number of unique IQs, the average number of unique IQs per path with standard deviation, and the corresponding uniqueness percentages.
}
\label{tab:iq-statistics-rh-2}
\resizebox{.9\textwidth}{!}{
\begin{tabular}{
l p{0pt}
cc p{0pt} cc p{0pt} cc
}

\toprule

& &
\multicolumn{8}{c}{\textbf{iLLC L1-4}} \\

\cmidrule(lr){3-10}

\textbf{Model} & &
\textbf{\#IQ} & \textbf{\#IQ/Path$_{\mathrm{std}}$} & & \textbf{\#uniqIQ} & \textbf{\#uniqIQ/Path$_{\mathrm{std}}$} & & \textbf{\%uniqIQ} & \textbf{\%uniqIQ/Path} \\

\midrule
Gemma 3       & & 34.98 & $5.83_{\mathrm{0.90}}$ & & 17.23 & $5.83_{\mathrm{0.90}}$ & & 49.25\% & 100.00\% \\
gpt-oss       & & 53.96 & $8.99_{\mathrm{1.31}}$ & & 35.02 & $8.99_{\mathrm{1.31}}$ & & 64.89\% & 100.00\% \\
Llama 3.1 8B  & & 23.58 & $3.93_{\mathrm{2.47}}$ & & 17.78 & $3.93_{\mathrm{2.47}}$ & & 75.40\% & 100.00\% \\
Llama 3.1 70B & & 39.54 & $6.59_{\mathrm{0.58}}$ & & 10.14 & $6.59_{\mathrm{0.58}}$ & & 25.65\% & 100.00\% \\
Mistral Small & & 38.87 & $6.48_{\mathrm{1.66}}$ & & 17.58 & $6.48_{\mathrm{1.66}}$ & & 45.23\% & 100.00\% \\
Phi-4         & & 42.53 & $7.09_{\mathrm{0.80}}$ & & 12.82 & $7.09_{\mathrm{0.80}}$ & & 30.14\% & 100.00\% \\
Qwen3         & & 36.85 & $6.14_{\mathrm{1.27}}$ & & 15.08 & $6.14_{\mathrm{1.27}}$ & & 40.94\% & 100.00\% \\
\cdashlinelr{1-10}
Average & & 38.62 & $6.44_{\mathrm{1.28}}$ & & 17.95 & $6.44_{\mathrm{1.28}}$ & & 47.36\% & 100.00\% \\

\midrule

& &
\multicolumn{8}{c}{\textbf{iLLC L1-6}} \\

\cmidrule(lr){3-10}

\textbf{Model} & &
\textbf{\#IQ} & \textbf{\#IQ/Path$_{\mathrm{std}}$} & & \textbf{\#uniqIQ} & \textbf{\#uniqIQ/Path$_{\mathrm{std}}$} & & \textbf{\%uniqIQ} & \textbf{\%uniqIQ/Path} \\

\midrule
Gemma 3       & & 41.16 & $6.86_{\mathrm{1.24}}$ & & 26.40 & $6.86_{\mathrm{1.24}}$ & & 64.12\% & 100.00\% \\
gpt-oss       & & 52.29 & $8.71_{\mathrm{1.49}}$ & & 34.24 & $8.71_{\mathrm{1.49}}$ & & 65.49\% & 100.00\% \\
Llama 3.1 8B  & & 29.36 & $4.89_{\mathrm{3.05}}$ & & 25.14 & $4.89_{\mathrm{3.05}}$ & & 85.61\% & 100.00\% \\
Llama 3.1 70B & & 46.94 & $7.82_{\mathrm{0.60}}$ & & 11.29 & $7.82_{\mathrm{0.60}}$ & & 24.06\% & 100.00\% \\
Mistral Small & & 35.98 & $6.00_{\mathrm{1.74}}$ & & 17.06 & $6.00_{\mathrm{1.74}}$ & & 47.43\% & 100.00\% \\
Phi-4         & & 54.28 & $9.05_{\mathrm{0.88}}$ & & 18.84 & $9.05_{\mathrm{0.88}}$ & & 34.70\% & 100.00\% \\
Qwen3         & & 38.41 & $6.40_{\mathrm{2.04}}$ & & 23.74 & $6.40_{\mathrm{2.04}}$ & & 61.81\% & 100.00\% \\
\cdashlinelr{1-10}
Average & & 42.63 & $7.10_{\mathrm{1.58}}$ & & 22.39 & $7.10_{\mathrm{1.58}}$ & & 54.75\% & 100.00\% \\

\midrule

& &
\multicolumn{8}{c}{\textbf{iLLC L1-8}} \\

\cmidrule(lr){3-10}

\textbf{Model} & &
\textbf{\#IQ} & \textbf{\#IQ/Path$_{\mathrm{std}}$} & & \textbf{\#uniqIQ} & \textbf{\#uniqIQ/Path$_{\mathrm{std}}$} & & \textbf{\%uniqIQ} & \textbf{\%uniqIQ/Path} \\

\midrule
Gemma 3       & & 41.31 & $6.88_{\mathrm{1.60}}$ & & 27.96 & $6.88_{\mathrm{1.60}}$ & & 67.70\% & 100.00\% \\
gpt-oss       & & 51.82 & $8.64_{\mathrm{1.57}}$ & & 33.97 & $8.64_{\mathrm{1.57}}$ & & 65.55\% & 100.00\% \\
Llama 3.1 8B  & & 33.90 & $5.65_{\mathrm{3.24}}$ & & 29.18 & $5.65_{\mathrm{3.24}}$ & & 86.06\% & 100.00\% \\
Llama 3.1 70B & & 36.72 & $6.12_{\mathrm{1.90}}$ & & 13.51 & $6.12_{\mathrm{1.90}}$ & & 36.79\% & 100.00\% \\
Mistral Small & & 30.15 & $5.02_{\mathrm{2.14}}$ & & 17.86 & $5.02_{\mathrm{2.14}}$ & & 59.23\% & 100.00\% \\
Phi-4         & & 58.45 & $9.74_{\mathrm{0.35}}$ & & 16.44 & $9.74_{\mathrm{0.35}}$ & & 28.13\% & 100.00\% \\
Qwen3         & & 37.19 & $6.20_{\mathrm{2.16}}$ & & 23.97 & $6.20_{\mathrm{2.16}}$ & & 64.46\% & 100.00\% \\
\cdashlinelr{1-10}
Average & & 41.36 & $6.89_{\mathrm{1.85}}$ & & 23.27 & $6.89_{\mathrm{1.85}}$ & & 58.27\% & 100.00\% \\

\midrule

& &
\multicolumn{8}{c}{\textbf{iLLC L2-4}} \\

\cmidrule(lr){3-10}

\textbf{Model} & &
\textbf{\#IQ} & \textbf{\#IQ/Path$_{\mathrm{std}}$} & & \textbf{\#uniqIQ} & \textbf{\#uniqIQ/Path$_{\mathrm{std}}$} & & \textbf{\%uniqIQ} & \textbf{\%uniqIQ/Path} \\

\midrule
Gemma 3       & & 43.20 & $7.20_{\mathrm{1.66}}$ & & 21.57 & $7.20_{\mathrm{1.66}}$ & & 49.93\% & 100.00\% \\
gpt-oss       & & 56.82 & $9.47_{\mathrm{0.85}}$ & & 40.82 & $9.47_{\mathrm{0.85}}$ & & 71.84\% & 100.00\% \\
Llama 3.1 8B  & & 41.58 & $6.93_{\mathrm{2.68}}$ & & 35.16 & $6.93_{\mathrm{2.68}}$ & & 84.57\% & 100.00\% \\
Llama 3.1 70B & & 49.00 & $8.17_{\mathrm{0.80}}$ & & 16.16 & $8.17_{\mathrm{0.80}}$ & & 32.99\% & 100.00\% \\
Mistral Small & & 36.17 & $6.03_{\mathrm{1.58}}$ & & 17.66 & $6.03_{\mathrm{1.58}}$ & & 48.81\% & 100.00\% \\
Phi-4         & & 42.29 & $7.05_{\mathrm{0.72}}$ & & 11.84 & $7.05_{\mathrm{0.72}}$ & & 28.00\% & 100.00\% \\
Qwen3         & & 38.90 & $6.48_{\mathrm{1.04}}$ & & 16.85 & $6.48_{\mathrm{1.04}}$ & & 43.32\% & 100.00\% \\
\cdashlinelr{1-10}
Average & & 43.99 & $7.33_{\mathrm{1.33}}$ & & 22.87 & $7.33_{\mathrm{1.33}}$ & & 51.35\% & 100.00\% \\

\midrule

& &
\multicolumn{8}{c}{\textbf{iLLC L2-6}} \\

\cmidrule(lr){3-10}

\textbf{Model} & &
\textbf{\#IQ} & \textbf{\#IQ/Path$_{\mathrm{std}}$} & & \textbf{\#uniqIQ} & \textbf{\#uniqIQ/Path$_{\mathrm{std}}$} & & \textbf{\%uniqIQ} & \textbf{\%uniqIQ/Path} \\

\midrule
Gemma 3       & & 45.94 & $7.66_{\mathrm{1.70}}$ & & 30.04 & $7.66_{\mathrm{1.70}}$ & & 65.40\% & 100.00\% \\
gpt-oss       & & 56.45 & $9.41_{\mathrm{0.97}}$ & & 40.37 & $9.41_{\mathrm{0.97}}$ & & 71.52\% & 100.00\% \\
Llama 3.1 8B  & & 36.38 & $6.06_{\mathrm{2.75}}$ & & 30.31 & $6.06_{\mathrm{2.75}}$ & & 83.32\% & 100.00\% \\
Llama 3.1 70B & & 53.71 & $8.95_{\mathrm{0.76}}$ & & 18.47 & $8.95_{\mathrm{0.76}}$ & & 34.39\% & 100.00\% \\
Mistral Small & & 34.03 & $5.67_{\mathrm{1.74}}$ & & 16.84 & $5.67_{\mathrm{1.74}}$ & & 49.49\% & 100.00\% \\
Phi-4         & & 53.99 & $9.00_{\mathrm{0.67}}$ & & 17.04 & $9.00_{\mathrm{0.67}}$ & & 31.56\% & 100.00\% \\
Qwen3         & & 39.52 & $6.59_{\mathrm{2.19}}$ & & 21.28 & $6.59_{\mathrm{2.19}}$ & & 53.86\% & 100.00\% \\
\cdashlinelr{1-10}
Average & & 45.72 & $7.62_{\mathrm{1.54}}$ & & 24.91 & $7.62_{\mathrm{1.54}}$ & & 55.65\% & 100.00\% \\

\midrule

& &
\multicolumn{8}{c}{\textbf{iLLC L2-8}} \\

\cmidrule(lr){3-10}

\textbf{Model} & &
\textbf{\#IQ} & \textbf{\#IQ/Path$_{\mathrm{std}}$} & & \textbf{\#uniqIQ} & \textbf{\#uniqIQ/Path$_{\mathrm{std}}$} & & \textbf{\%uniqIQ} & \textbf{\%uniqIQ/Path} \\

\midrule
Gemma 3       & & 50.66 & $8.44_{\mathrm{1.60}}$ & & 39.44 & $8.44_{\mathrm{1.60}}$ & & 77.85\% & 100.00\% \\
gpt-oss       & & 55.85 & $9.31_{\mathrm{1.04}}$ & & 40.71 & $9.31_{\mathrm{1.04}}$ & & 72.89\% & 100.00\% \\
Llama 3.1 8B  & & 38.24 & $6.37_{\mathrm{2.98}}$ & & 32.79 & $6.37_{\mathrm{2.98}}$ & & 85.74\% & 100.00\% \\
Llama 3.1 70B & & 52.11 & $8.68_{\mathrm{1.12}}$ & & 23.75 & $8.68_{\mathrm{1.12}}$ & & 45.58\% & 100.00\% \\
Mistral Small & & 28.25 & $4.71_{\mathrm{2.08}}$ & & 18.62 & $4.71_{\mathrm{2.08}}$ & & 65.91\% & 100.00\% \\
Phi-4         & & 59.25 & $9.87_{\mathrm{0.22}}$ & & 16.70 & $9.87_{\mathrm{0.22}}$ & & 28.18\% & 100.00\% \\
Qwen3         & & 31.70 & $5.28_{\mathrm{2.22}}$ & & 19.56 & $5.28_{\mathrm{2.22}}$ & & 61.70\% & 100.00\% \\
\cdashlinelr{1-10}
Average & & 45.15 & $7.52_{\mathrm{1.61}}$ & & 27.37 & $7.52_{\mathrm{1.61}}$ & & 62.55\% & 100.00\% \\

\bottomrule

\end{tabular}
}
\vspace{-1.1em}
\end{table*}

\begin{table*}[htb]
\centering
\caption{
Statistics of generated intermediate questions (IQs) in ReHoPER \textbf{without replanning} on MuSR MM, MoreHopQA, and iLLC L2-4. We report the total number of generated IQs, the average number of IQs per path with standard deviation, the number of unique IQs, the average number of unique IQs per path with standard deviation, and the corresponding uniqueness percentages.
}
\label{tab:iq-statistics-off}
\resizebox{.8\textwidth}{!}{
\begin{tabular}{
l p{0pt}
cc p{0pt} cc p{0pt} cc
}

\toprule

& &
\multicolumn{8}{c}{\textbf{MuSR MM}} \\

\cmidrule(lr){3-10}

\textbf{Model} & &
\textbf{\#IQ} & \textbf{\#IQ/Path$_{\mathrm{std}}$} & &\textbf{\#uniqIQ} & \textbf{\#uniqIQ/Path$_{\mathrm{std}}$} & & \textbf{\%uniqIQ} & \textbf{\%uniqIQ/Path} \\

\midrule
Gemma 3 & & 48.73 & 8.12$_{\mathrm{1.07}}$ & & 48.53 & 8.12$_{\mathrm{1.07}}$ & & 99.58\% & 100.00\% \\
gpt-oss & & 59.70 & 9.95$_{\mathrm{0.10}}$ & & 59.62 & 9.95$_{\mathrm{0.10}}$ & & 99.87\% & 99.99\% \\
Llama 3.1 8B & & 46.69 & 7.78$_{\mathrm{2.42}}$ & & 46.64 & 7.78$_{\mathrm{2.42}}$ & & 99.90\% & 99.97\% \\
Llama 3.1 70B & & 50.04 & 8.34$_{\mathrm{1.20}}$ & & 49.74 & 8.34$_{\mathrm{1.20}}$ & & 99.41\% & 100.00\% \\
Mistral Small & & 58.78 & 9.80$_{\mathrm{0.38}}$ & & 58.12 & 9.79$_{\mathrm{0.39}}$ & & 98.87\% & 99.94\% \\
Phi-4 & & 58.67 & 9.78$_{\mathrm{0.40}}$ & & 58.04 & 9.78$_{\mathrm{0.40}}$ & & 98.92\% & 100.00\% \\
Qwen3 & & 59.44 & 9.91$_{\mathrm{0.19}}$ & & 58.16 & 9.91$_{\mathrm{0.19}}$ & & 97.86\% & 99.99\% \\
\cdashlinelr{1-10}
Average & & 54.58 & $9.10_{\mathrm{0.82}}$ & & 54.12 & $9.10_{\mathrm{0.82}}$ & & 99.20\% & 99.98\% \\

\midrule

& &
\multicolumn{8}{c}{\textbf{MoreHopQA}} \\

\cmidrule(lr){3-10}

\textbf{Model} & &
\textbf{\#IQ} & \textbf{\#IQ/Path$_{\mathrm{std}}$} & &\textbf{\#uniqIQ} & \textbf{\#uniqIQ/Path$_{\mathrm{std}}$} & & \textbf{\%uniqIQ} & \textbf{\%uniqIQ/Path} \\

\midrule
Gemma 3 & & 19.27 & 3.21$_{\mathrm{0.56}}$ & & 10.95 & 3.21$_{\mathrm{0.56}}$ & & 56.83\% & 100.00\% \\
gpt-oss & & 24.09 & 4.01$_{\mathrm{1.16}}$ & & 19.32 & 4.01$_{\mathrm{1.16}}$ & & 80.21\% & 100.00\% \\
Llama 3.1 8B & & 16.32 & 2.72$_{\mathrm{1.49}}$ & & 14.09 & 2.71$_{\mathrm{1.49}}$ & & 86.32\% & 99.80\% \\
Llama 3.1 70B & & 18.97 & 3.16$_{\mathrm{0.78}}$ & & 13.02 & 3.15$_{\mathrm{0.77}}$ & & 68.65\% & 99.72\% \\
Mistral Small & & 15.95 & 2.66$_{\mathrm{0.69}}$ & & 11.03 & 2.66$_{\mathrm{0.69}}$ & & 69.19\% & 99.96\% \\
Phi-4 & & 11.68 & 1.95$_{\mathrm{1.72}}$ & & 10.11 & 1.95$_{\mathrm{1.72}}$ & & 86.59\% & 100.00\% \\
Qwen3 & & 20.83 & 3.47$_{\mathrm{0.32}}$ & & 8.36 & 3.47$_{\mathrm{0.31}}$ & & 40.13\% & 99.90\% \\
\cdashlinelr{1-10}
Average & & 18.16 & $3.03_{\mathrm{0.96}}$ & & 12.41 & $3.02_{\mathrm{0.96}}$ & & 69.70\% & 99.91\% \\

\midrule

& &
\multicolumn{8}{c}{\textbf{iLLC L2-4}} \\

\cmidrule(lr){3-10}

\textbf{Model} & &
\textbf{\#IQ} & \textbf{\#IQ/Path$_{\mathrm{std}}$} & &\textbf{\#uniqIQ} & \textbf{\#uniqIQ/Path$_{\mathrm{std}}$} & & \textbf{\%uniqIQ} & \textbf{\%uniqIQ/Path} \\

\midrule
Gemma 3 & & 22.95 & 3.82$_{\mathrm{0.96}}$ & & 17.18 & 3.82$_{\mathrm{0.96}}$ & & 74.88\% & 99.97\% \\
gpt-oss & & 28.90 & 4.82$_{\mathrm{2.14}}$ & & 25.44 & 4.82$_{\mathrm{2.14}}$ & & 88.03\% & 100.00\% \\
Llama 3.1 8B & & 7.78 & 1.30$_{\mathrm{0.64}}$ & & 6.48 & 1.30$_{\mathrm{0.64}}$ & & 83.20\% & 100.00\% \\
Llama 3.1 70B & & 26.34 & 4.39$_{\mathrm{1.12}}$ & & 23.05 & 4.39$_{\mathrm{1.12}}$ & & 87.53\% & 100.00\% \\
Mistral Small & & 19.37 & 3.23$_{\mathrm{0.63}}$ & & 14.13 & 3.23$_{\mathrm{0.63}}$ & & 72.97\% & 100.00\% \\
Phi-4 & & 27.20 & 4.53$_{\mathrm{1.52}}$ & & 19.00 & 4.53$_{\mathrm{1.52}}$ & & 69.86\% & 100.00\% \\
Qwen3 & & 30.76 & 5.13$_{\mathrm{1.09}}$ & & 21.58 & 5.13$_{\mathrm{1.09}}$ & & 70.14\% & 100.00\% \\
\cdashlinelr{1-10}
Average & & 23.33 & $3.89_{\mathrm{1.16}}$ & & 18.12 & $3.89_{\mathrm{1.16}}$ & & 78.09\% & 100.00\% \\

\bottomrule

\end{tabular}
}
\vspace{-1.1em}
\end{table*}

\begin{table*}[htb]
\centering
\caption{
Statistics of generated intermediate questions (IQs) in ReHoPER \textbf{without look-ahead} on MuSR MM, MoreHopQA, and iLLC L2-4. We report the total number of generated IQs, the average number of IQs per path with standard deviation, the number of unique IQs, the average number of unique IQs per path with standard deviation, and the corresponding uniqueness percentages.
}
\label{tab:iq-statistics-on}
\resizebox{.8\textwidth}{!}{
\begin{tabular}{
l p{0pt}
cc p{0pt} cc p{0pt} cc
}

\toprule

& &
\multicolumn{8}{c}{\textbf{MuSR MM}} \\

\cmidrule(lr){3-10}

\textbf{Model} & &
\textbf{\#IQ} & \textbf{\#IQ/Path$_{\mathrm{std}}$} & &\textbf{\#uniqIQ} & \textbf{\#uniqIQ/Path$_{\mathrm{std}}$} & & \textbf{\%uniqIQ} & \textbf{\%uniqIQ/Path} \\

\midrule

Gemma 3 & & 60.00 & 10.00$_{\mathrm{0.00}}$ & & 58.76 & 9.95$_{\mathrm{0.09}}$ & & 97.93\% & 99.53\% \\
gpt-oss & & 59.97 & 10.00$_{\mathrm{0.01}}$ & & 59.29 & 9.88$_{\mathrm{0.21}}$ & & 98.86\% & 98.88\% \\
Llama 3.1 8B & & 60.00 & 10.00$_{\mathrm{0.00}}$ & & 57.86 & 9.69$_{\mathrm{0.47}}$ & & 96.43\% & 96.87\% \\
Llama 3.1 70B & & 60.00 & 10.00$_{\mathrm{0.00}}$ & & 59.11 & 9.95$_{\mathrm{0.10}}$ & & 98.51\% & 99.49\% \\
Mistral Small & & 60.00 & 10.00$_{\mathrm{0.00}}$ & & 57.53 & 9.62$_{\mathrm{0.52}}$ & & 95.88\% & 96.15\% \\
Phi-4 & & 60.00 & 10.00$_{\mathrm{0.00}}$ & & 48.68 & 8.12$_{\mathrm{1.28}}$ & & 81.13\% & 81.20\% \\
Qwen3 & & 60.00 & 10.00$_{\mathrm{0.00}}$ & & 56.44 & 9.59$_{\mathrm{0.63}}$ & & 94.07\% & 95.86\% \\
\cdashlinelr{1-10}
Average & & 60.00 & $10.00_{\mathrm{0.00}}$ & & 56.81 & $9.54_{\mathrm{0.47}}$ & & 94.69\% & 95.43\% \\

\midrule

& &
\multicolumn{8}{c}{\textbf{MoreHopQA}} \\

\cmidrule(lr){3-10}

\textbf{Model} & &
\textbf{\#IQ} & \textbf{\#IQ/Path$_{\mathrm{std}}$} & &\textbf{\#uniqIQ} & \textbf{\#uniqIQ/Path$_{\mathrm{std}}$} & & \textbf{\%uniqIQ} & \textbf{\%uniqIQ/Path} \\

\midrule

Gemma 3 & & 60.00 & 10.00$_{\mathrm{0.00}}$ & & 9.95 & 5.01$_{\mathrm{0.68}}$ & & 16.58\% & 50.08\% \\
gpt-oss & & 60.00 & 10.00$_{\mathrm{0.00}}$ & & 24.18 & 5.65$_{\mathrm{1.34}}$ & & 40.30\% & 56.50\% \\
Llama 3.1 70B & & 59.97 & 9.99$_{\mathrm{0.00}}$ & & 32.85 & 8.34$_{\mathrm{1.00}}$ & & 54.77\% & 83.44\% \\
Llama 3.1 8B & & 60.00 & 10.00$_{\mathrm{0.00}}$ & & 31.94 & 6.76$_{\mathrm{1.32}}$ & & 53.23\% & 67.61\% \\
Mistral Small & & 60.00 & 10.00$_{\mathrm{0.00}}$ & & 21.03 & 7.01$_{\mathrm{0.98}}$ & & 35.06\% & 70.07\% \\
Phi-4 & & 60.00 & 10.00$_{\mathrm{0.00}}$ & & 17.67 & 5.23$_{\mathrm{1.09}}$ & & 29.46\% & 52.33\% \\
Qwen3 & & 60.00 & 10.00$_{\mathrm{0.00}}$ & & 5.95 & 2.39$_{\mathrm{0.51}}$ & & 9.92\% & 23.92\% \\
\cdashlinelr{1-10}
Average & & 60.00 & $10.00_{\mathrm{0.00}}$ & & 20.51 & $5.77_{\mathrm{0.99}}$ & & 34.19\% & 57.71\% \\

\midrule

& &
\multicolumn{8}{c}{\textbf{iLLC L2-4}} \\

\cmidrule(lr){3-10}

\textbf{Model} & &
\textbf{\#IQ} & \textbf{\#IQ/Path$_{\mathrm{std}}$} & &\textbf{\#uniqIQ} & \textbf{\#uniqIQ/Path$_{\mathrm{std}}$} & & \textbf{\%uniqIQ} & \textbf{\%uniqIQ/Path} \\

\midrule

Gemma 3 & & 60.00 & 10.00$_{\mathrm{0.00}}$ & & 12.93 & 4.73$_{\mathrm{0.73}}$ & & 21.55\% & 47.27\% \\
gpt-oss & & 59.94 & 9.99$_{\mathrm{0.02}}$ & & 47.28 & 8.81$_{\mathrm{1.10}}$ & & 78.89\% & 88.15\% \\
Llama 3.1 70B & & 60.00 & 10.00$_{\mathrm{0.00}}$ & & 31.00 & 9.41$_{\mathrm{0.64}}$ & & 51.67\% & 94.11\% \\
Llama 3.1 8B & & 60.00 & 10.00$_{\mathrm{0.00}}$ & & 27.67 & 6.41$_{\mathrm{1.30}}$ & & 46.12\% & 64.12\% \\
Mistral Small & & 60.00 & 10.00$_{\mathrm{0.00}}$ & & 37.13 & 7.88$_{\mathrm{1.11}}$ & & 61.88\% & 78.84\% \\
Phi-4 & & 60.00 & 10.00$_{\mathrm{0.00}}$ & & 31.83 & 7.16$_{\mathrm{1.30}}$ & & 53.05\% & 71.63\% \\
Qwen3 & & 60.00 & 10.00$_{\mathrm{0.00}}$ & & 3.67 & 2.03$_{\mathrm{0.24}}$ & & 6.12\% & 20.34\% \\
\cdashlinelr{1-10}
Average & & 59.99 & $10.00_{\mathrm{0.00}}$ & & 27.36 & $6.63_{\mathrm{0.92}}$ & & 45.61\% & 66.35\% \\

\bottomrule

\end{tabular}
}
\vspace{-1.1em}
\end{table*}

%% file: tables/trajectory.tex
\begin{table*}[htb]
\vspace{1em}
\centering
\Large
\caption{
Trajectory statistics under the second-next analysis. $T$ is the number of valid step transitions, $R$ is the number of replanned transitions, and $N$ is the total number of instances.
}
\label{tab:trajectory-1}
\resizebox{\textwidth}{!}{
\begin{tabular}{
l p{0pt}
ccccc p{0pt}
ccccc p{0pt}
ccccc
}

\toprule

& &
\multicolumn{17}{c}{\textbf{Gemma 3}} \\

\cmidrule(lr){3-19}

& &
\multicolumn{5}{c}{\textbf{MuSR MM (N: 250)}} & & \multicolumn{5}{c}{\textbf{MoreHopQA (N: 150)}} & & \multicolumn{5}{c}{\textbf{iLLC L2-4 (N: 250)}} \\

\cmidrule(lr){3-7} \cmidrule(lr){9-13} \cmidrule(lr){15-19}

\textbf{Step} & &
\textbf{T$_{\mathrm{std}}$} & \textbf{\%T/N$_{\mathrm{std}}$} & \textbf{R$_{\mathrm{std}}$} & \textbf{\%R/T$_{\mathrm{std}}$} & \textbf{\%R/N$_{\mathrm{std}}$} & &
\textbf{T$_{\mathrm{std}}$} & \textbf{\%T/N$_{\mathrm{std}}$} & \textbf{R$_{\mathrm{std}}$} & \textbf{\%R/T$_{\mathrm{std}}$} & \textbf{\%R/N$_{\mathrm{std}}$} & &\textbf{T$_{\mathrm{std}}$} & \textbf{\%T/N$_{\mathrm{std}}$} & \textbf{R$_{\mathrm{std}}$} & \textbf{\%R/T$_{\mathrm{std}}$} & \textbf{\%R/N$_{\mathrm{std}}$} \\

\midrule

1$\rightarrow$2 & & 250.00$_{0.00}$ & 100.0$_{0.0}$ & 249.83$_{0.37}$ & 99.9$_{0.1}$ & 99.9$_{0.1}$ & & 138.17$_{2.27}$ & 92.1$_{1.5}$ & 122.50$_{3.69}$ & 88.7$_{3.0}$ & 81.7$_{2.5}$ & & 250.00$_{0.00}$ & 100.0$_{0.0}$ & 209.33$_{21.02}$ & 83.7$_{8.4}$ & 83.7$_{8.4}$ \\
2$\rightarrow$3 & & 250.00$_{0.00}$ & 100.0$_{0.0}$ & 250.00$_{0.00}$ & 100.0$_{0.0}$ & 100.0$_{0.0}$ & & 124.33$_{2.49}$ & 82.9$_{1.7}$ & 105.50$_{2.14}$ & 84.9$_{2.6}$ & 70.3$_{1.4}$ & & 250.00$_{0.00}$ & 100.0$_{0.0}$ & 131.00$_{34.24}$ & 52.4$_{13.7}$ & 52.4$_{13.7}$ \\
3$\rightarrow$4 & & 250.00$_{0.00}$ & 100.0$_{0.0}$ & 249.83$_{0.37}$ & 99.9$_{0.1}$ & 99.9$_{0.1}$ & & 118.17$_{3.13}$ & 78.8$_{2.1}$ & 101.33$_{3.68}$ & 85.8$_{2.0}$ & 67.6$_{2.5}$ & & 248.17$_{1.21}$ & 99.3$_{0.5}$ & 73.83$_{16.87}$ & 29.7$_{6.8}$ & 29.5$_{6.7}$ \\
4$\rightarrow$5 & & 250.00$_{0.00}$ & 100.0$_{0.0}$ & 250.00$_{0.00}$ & 100.0$_{0.0}$ & 100.0$_{0.0}$ & & 106.00$_{5.32}$ & 70.7$_{3.5}$ & 90.50$_{3.91}$ & 85.4$_{2.3}$ & 60.3$_{2.6}$ & & 241.17$_{7.60}$ & 96.5$_{3.0}$ & 74.83$_{18.96}$ & 31.1$_{8.0}$ & 29.9$_{7.6}$ \\
5$\rightarrow$6 & & 250.00$_{0.00}$ & 100.0$_{0.0}$ & 249.83$_{0.37}$ & 99.9$_{0.1}$ & 99.9$_{0.1}$ & & 98.83$_{3.18}$ & 65.9$_{2.1}$ & 85.00$_{3.65}$ & 86.0$_{3.5}$ & 56.7$_{2.4}$ & & 155.83$_{39.33}$ & 62.3$_{15.7}$ & 65.00$_{15.93}$ & 42.7$_{11.6}$ & 26.0$_{6.4}$ \\
6$\rightarrow$7 & & 250.00$_{0.00}$ & 100.0$_{0.0}$ & 250.00$_{0.00}$ & 100.0$_{0.0}$ & 100.0$_{0.0}$ & & 81.17$_{3.80}$ & 54.1$_{2.5}$ & 70.33$_{4.82}$ & 86.6$_{2.7}$ & 46.9$_{3.2}$ & & 117.00$_{23.33}$ & 46.8$_{9.3}$ & 58.83$_{13.64}$ & 50.3$_{8.3}$ & 23.5$_{5.5}$ \\
7$\rightarrow$8 & & 250.00$_{0.00}$ & 100.0$_{0.0}$ & 250.00$_{0.00}$ & 100.0$_{0.0}$ & 100.0$_{0.0}$ & & 69.33$_{3.14}$ & 46.2$_{2.1}$ & 59.33$_{5.15}$ & 85.5$_{4.8}$ & 39.6$_{3.4}$ & & 110.50$_{24.61}$ & 44.2$_{9.8}$ & 56.00$_{16.42}$ & 49.9$_{7.5}$ & 22.4$_{6.6}$ \\
8$\rightarrow$9 & & 250.00$_{0.00}$ & 100.0$_{0.0}$ & 250.00$_{0.00}$ & 100.0$_{0.0}$ & 100.0$_{0.0}$ & & 56.17$_{3.48}$ & 37.4$_{2.3}$ & 47.83$_{4.06}$ & 85.2$_{5.3}$ & 31.9$_{2.7}$ & & 102.33$_{23.49}$ & 40.9$_{9.4}$ & 51.33$_{11.93}$ & 50.5$_{7.1}$ & 20.5$_{4.8}$ \\
9$\rightarrow$10 & & 250.00$_{0.00}$ & 100.0$_{0.0}$ & 250.00$_{0.00}$ & 100.0$_{0.0}$ & 100.0$_{0.0}$ & & 48.33$_{3.40}$ & 32.2$_{2.3}$ & 42.33$_{2.75}$ & 87.7$_{2.3}$ & 28.2$_{1.8}$ & & 48.00$_{8.79}$ & 19.2$_{3.5}$ & 26.67$_{7.65}$ & 54.5$_{8.6}$ & 10.7$_{3.1}$ \\

\cdashlinelr{1-19}

ANY & & 250.00$_{0.00}$ & 100.0$_{0.0}$ & 250.00$_{0.00}$ & 100.0$_{0.0}$ & 100.0$_{0.0}$ & & 142.50$_{1.50}$ & 95.0$_{1.0}$ & 137.83$_{2.27}$ & 96.7$_{1.3}$ & 91.9$_{1.5}$ & & 250.00$_{0.00}$ & 100.0$_{0.0}$ & 242.00$_{7.48}$ & 96.8$_{3.0}$ & 96.8$_{3.0}$ \\

\midrule

& &
\multicolumn{17}{c}{\textbf{gpt-oss}} \\

\cmidrule(lr){3-19}

& &
\multicolumn{5}{c}{\textbf{MuSR MM (N: 250)}} & & \multicolumn{5}{c}{\textbf{MoreHopQA (N: 150)}} & & \multicolumn{5}{c}{\textbf{iLLC L2-4 (N: 250)}} \\

\cmidrule(lr){3-7} \cmidrule(lr){9-13} \cmidrule(lr){15-19}

\textbf{Step} & &
\textbf{T$_{\mathrm{std}}$} & \textbf{\%T/N$_{\mathrm{std}}$} & \textbf{R$_{\mathrm{std}}$} & \textbf{\%R/T$_{\mathrm{std}}$} & \textbf{\%R/N$_{\mathrm{std}}$} & &
\textbf{T$_{\mathrm{std}}$} & \textbf{\%T/N$_{\mathrm{std}}$} & \textbf{R$_{\mathrm{std}}$} & \textbf{\%R/T$_{\mathrm{std}}$} & \textbf{\%R/N$_{\mathrm{std}}$} & &\textbf{T$_{\mathrm{std}}$} & \textbf{\%T/N$_{\mathrm{std}}$} & \textbf{R$_{\mathrm{std}}$} & \textbf{\%R/T$_{\mathrm{std}}$} & \textbf{\%R/N$_{\mathrm{std}}$} \\

\midrule

1$\rightarrow$2 & & 173.67$_{10.40}$ & 69.5$_{4.2}$ & 173.33$_{10.45}$ & 99.8$_{0.3}$ & 69.3$_{4.2}$ & & 147.67$_{1.11}$ & 98.4$_{0.7}$ & 133.67$_{3.77}$ & 90.5$_{2.3}$ & 89.1$_{2.5}$ & & 248.50$_{0.96}$ & 99.4$_{0.4}$ & 242.83$_{3.58}$ & 97.7$_{1.4}$ & 97.1$_{1.4}$ \\
2$\rightarrow$3 & & 166.00$_{10.86}$ & 66.4$_{4.3}$ & 165.00$_{10.69}$ & 99.4$_{0.3}$ & 66.0$_{4.3}$ & & 148.67$_{1.11}$ & 99.1$_{0.7}$ & 132.83$_{3.24}$ & 89.3$_{1.8}$ & 88.6$_{2.2}$ & & 249.00$_{1.00}$ & 99.6$_{0.4}$ & 243.67$_{2.43}$ & 97.9$_{1.0}$ & 97.5$_{1.0}$ \\
3$\rightarrow$4 & & 165.50$_{9.22}$ & 66.2$_{3.7}$ & 164.33$_{8.67}$ & 99.3$_{0.6}$ & 65.7$_{3.5}$ & & 144.00$_{2.52}$ & 96.0$_{1.7}$ & 130.50$_{2.63}$ & 90.6$_{1.3}$ & 87.0$_{1.8}$ & & 245.67$_{1.80}$ & 98.3$_{0.7}$ & 236.00$_{2.52}$ & 96.1$_{0.9}$ & 94.4$_{1.0}$ \\
4$\rightarrow$5 & & 160.50$_{11.12}$ & 64.2$_{4.4}$ & 158.67$_{11.64}$ & 98.8$_{0.6}$ & 63.5$_{4.7}$ & & 137.33$_{2.56}$ & 91.6$_{1.7}$ & 126.17$_{2.03}$ & 91.9$_{2.3}$ & 84.1$_{1.4}$ & & 235.67$_{4.92}$ & 94.3$_{2.0}$ & 225.50$_{6.75}$ & 95.7$_{1.0}$ & 90.2$_{2.7}$ \\
5$\rightarrow$6 & & 157.67$_{9.96}$ & 63.1$_{4.0}$ & 156.67$_{10.14}$ & 99.4$_{0.5}$ & 62.7$_{4.1}$ & & 133.33$_{2.98}$ & 88.9$_{2.0}$ & 121.50$_{2.50}$ & 91.2$_{3.3}$ & 81.0$_{1.7}$ & & 230.00$_{4.97}$ & 92.0$_{2.0}$ & 219.50$_{7.83}$ & 95.4$_{1.7}$ & 87.8$_{3.1}$ \\
6$\rightarrow$7 & & 155.17$_{9.25}$ & 62.1$_{3.7}$ & 153.00$_{9.83}$ & 98.6$_{0.9}$ & 61.2$_{3.9}$ & & 123.50$_{4.43}$ & 82.3$_{3.0}$ & 113.17$_{6.41}$ & 91.6$_{2.5}$ & 75.4$_{4.3}$ & & 209.50$_{5.35}$ & 83.8$_{2.1}$ & 202.00$_{4.76}$ & 96.4$_{0.8}$ & 80.8$_{1.9}$ \\
7$\rightarrow$8 & & 149.83$_{10.35}$ & 59.9$_{4.1}$ & 148.33$_{9.41}$ & 99.0$_{0.7}$ & 59.3$_{3.8}$ & & 119.33$_{3.68}$ & 79.6$_{2.5}$ & 110.33$_{4.42}$ & 92.4$_{2.0}$ & 73.6$_{2.9}$ & & 207.50$_{5.12}$ & 83.0$_{2.0}$ & 202.33$_{6.32}$ & 97.5$_{0.9}$ & 80.9$_{2.5}$ \\
8$\rightarrow$9 & & 147.17$_{8.93}$ & 58.9$_{3.6}$ & 144.33$_{8.16}$ & 98.1$_{1.2}$ & 57.7$_{3.3}$ & & 109.83$_{2.19}$ & 73.2$_{1.5}$ & 101.50$_{4.27}$ & 92.4$_{2.5}$ & 67.7$_{2.8}$ & & 177.50$_{5.32}$ & 71.0$_{2.1}$ & 173.67$_{4.75}$ & 97.8$_{0.3}$ & 69.5$_{1.9}$ \\
9$\rightarrow$10 & & 143.17$_{7.95}$ & 57.3$_{3.2}$ & 140.33$_{7.27}$ & 98.0$_{1.2}$ & 56.1$_{2.9}$ & & 100.67$_{4.07}$ & 67.1$_{2.7}$ & 93.67$_{3.35}$ & 93.1$_{2.2}$ & 62.4$_{2.2}$ & & 171.33$_{6.42}$ & 68.5$_{2.6}$ & 167.67$_{7.36}$ & 97.8$_{1.2}$ & 67.1$_{2.9}$ \\

\cdashlinelr{1-19}

ANY & & 181.50$_{10.59}$ & 72.6$_{4.2}$ & 181.50$_{10.59}$ & 100.0$_{0.0}$ & 72.6$_{4.2}$ & & 149.17$_{0.69}$ & 99.4$_{0.5}$ & 148.50$_{0.96}$ & 99.6$_{0.5}$ & 99.0$_{0.6}$ & & 249.17$_{1.07}$ & 99.7$_{0.4}$ & 249.17$_{1.07}$ & 100.0$_{0.0}$ & 99.7$_{0.4}$ \\

\bottomrule

\end{tabular}
}
\end{table*}

\begin{table*}[htb]
\centering
\Large
\caption{
Trajectory statistics under the second-next analysis. $T$ is the number of valid step transitions, $R$ is the number of replanned transitions, and $N$ is the total number of instances.
}
\label{tab:trajectory-2}
\resizebox{\textwidth}{!}{
\begin{tabular}{
l p{0pt}
ccccc p{0pt}
ccccc p{0pt}
ccccc
}

\toprule

& &
\multicolumn{17}{c}{\textbf{Llama 3.1 8B}} \\

\cmidrule(lr){3-19}

& &
\multicolumn{5}{c}{\textbf{MuSR MM (N: 250)}} & & \multicolumn{5}{c}{\textbf{MoreHopQA (N: 150)}} & & \multicolumn{5}{c}{\textbf{iLLC L2-4 (N: 250)}} \\

\cmidrule(lr){3-7} \cmidrule(lr){9-13} \cmidrule(lr){15-19}

\textbf{Step} & &
\textbf{T$_{\mathrm{std}}$} & \textbf{\%T/N$_{\mathrm{std}}$} & \textbf{R$_{\mathrm{std}}$} & \textbf{\%R/T$_{\mathrm{std}}$} & \textbf{\%R/N$_{\mathrm{std}}$} & &
\textbf{T$_{\mathrm{std}}$} & \textbf{\%T/N$_{\mathrm{std}}$} & \textbf{R$_{\mathrm{std}}$} & \textbf{\%R/T$_{\mathrm{std}}$} & \textbf{\%R/N$_{\mathrm{std}}$} & &\textbf{T$_{\mathrm{std}}$} & \textbf{\%T/N$_{\mathrm{std}}$} & \textbf{R$_{\mathrm{std}}$} & \textbf{\%R/T$_{\mathrm{std}}$} & \textbf{\%R/N$_{\mathrm{std}}$} \\

\midrule

1$\rightarrow$2 & & 250.00$_{0.00}$ & 100.0$_{0.0}$ & 250.00$_{0.00}$ & 100.0$_{0.0}$ & 100.0$_{0.0}$ & & 109.67$_{5.12}$ & 73.1$_{3.4}$ & 101.83$_{4.52}$ & 92.9$_{1.4}$ & 67.9$_{3.0}$ & & 221.83$_{8.63}$ & 88.7$_{3.5}$ & 219.17$_{9.10}$ & 98.8$_{1.0}$ & 87.7$_{3.6}$ \\
2$\rightarrow$3 & & 250.00$_{0.00}$ & 100.0$_{0.0}$ & 249.00$_{1.00}$ & 99.6$_{0.4}$ & 99.6$_{0.4}$ & & 105.67$_{4.15}$ & 70.4$_{2.8}$ & 99.00$_{6.14}$ & 93.6$_{2.8}$ & 66.0$_{4.1}$ & & 210.33$_{9.62}$ & 84.1$_{3.8}$ & 199.00$_{9.27}$ & 94.6$_{2.3}$ & 79.6$_{3.7}$ \\
3$\rightarrow$4 & & 249.67$_{0.47}$ & 99.9$_{0.2}$ & 248.83$_{0.90}$ & 99.7$_{0.3}$ & 99.5$_{0.4}$ & & 94.67$_{2.49}$ & 63.1$_{1.7}$ & 88.67$_{2.21}$ & 93.7$_{2.3}$ & 59.1$_{1.5}$ & & 201.83$_{11.91}$ & 80.7$_{4.8}$ & 186.50$_{9.39}$ & 92.5$_{1.0}$ & 74.6$_{3.8}$ \\
4$\rightarrow$5 & & 250.00$_{0.00}$ & 100.0$_{0.0}$ & 249.00$_{0.82}$ & 99.6$_{0.3}$ & 99.6$_{0.3}$ & & 79.67$_{2.29}$ & 53.1$_{1.5}$ & 74.17$_{2.91}$ & 93.1$_{1.7}$ & 49.4$_{1.9}$ & & 176.50$_{13.46}$ & 70.6$_{5.4}$ & 154.50$_{13.51}$ & 87.5$_{1.9}$ & 61.8$_{5.4}$ \\
5$\rightarrow$6 & & 250.00$_{0.00}$ & 100.0$_{0.0}$ & 247.33$_{1.11}$ & 98.9$_{0.4}$ & 98.9$_{0.4}$ & & 73.83$_{3.72}$ & 49.2$_{2.5}$ & 68.00$_{4.04}$ & 92.1$_{2.6}$ & 45.3$_{2.7}$ & & 153.17$_{15.02}$ & 61.3$_{6.0}$ & 131.83$_{11.20}$ & 86.2$_{1.5}$ & 52.7$_{4.5}$ \\
6$\rightarrow$7 & & 249.67$_{0.47}$ & 99.9$_{0.2}$ & 247.50$_{1.61}$ & 99.1$_{0.5}$ & 99.0$_{0.6}$ & & 65.17$_{3.39}$ & 43.4$_{2.3}$ & 59.67$_{3.14}$ & 91.6$_{3.6}$ & 39.8$_{2.1}$ & & 129.00$_{11.21}$ & 51.6$_{4.5}$ & 112.33$_{11.47}$ & 87.1$_{4.4}$ & 44.9$_{4.6}$ \\
7$\rightarrow$8 & & 249.50$_{0.76}$ & 99.8$_{0.3}$ & 245.67$_{2.69}$ & 98.5$_{0.9}$ & 98.3$_{1.1}$ & & 55.67$_{4.07}$ & 37.1$_{2.7}$ & 52.17$_{2.41}$ & 93.9$_{3.5}$ & 34.8$_{1.6}$ & & 107.83$_{9.12}$ & 43.1$_{3.6}$ & 90.83$_{10.95}$ & 84.1$_{5.1}$ & 36.3$_{4.4}$ \\
8$\rightarrow$9 & & 249.17$_{1.07}$ & 99.7$_{0.4}$ & 245.33$_{2.62}$ & 98.5$_{1.2}$ & 98.1$_{1.0}$ & & 46.33$_{4.11}$ & 30.9$_{2.7}$ & 43.67$_{4.15}$ & 94.2$_{2.8}$ & 29.1$_{2.8}$ & & 91.33$_{7.61}$ & 36.5$_{3.0}$ & 79.50$_{10.10}$ & 86.8$_{4.5}$ & 31.8$_{4.0}$ \\
9$\rightarrow$10 & & 248.50$_{0.76}$ & 99.4$_{0.3}$ & 245.33$_{1.25}$ & 98.7$_{0.5}$ & 98.1$_{0.5}$ & & 41.17$_{6.31}$ & 27.4$_{4.2}$ & 39.33$_{5.47}$ & 95.8$_{3.0}$ & 26.2$_{3.6}$ & & 80.83$_{13.01}$ & 32.3$_{5.2}$ & 70.83$_{14.78}$ & 87.0$_{4.6}$ & 28.3$_{5.9}$ \\

\cdashlinelr{1-19}

ANY & & 250.00$_{0.00}$ & 100.0$_{0.0}$ & 250.00$_{0.00}$ & 100.0$_{0.0}$ & 100.0$_{0.0}$ & & 135.83$_{3.13}$ & 90.6$_{2.1}$ & 134.17$_{2.67}$ & 98.8$_{0.7}$ & 89.4$_{1.8}$ & & 228.00$_{8.02}$ & 91.2$_{3.2}$ & 228.00$_{8.02}$ & 100.0$_{0.0}$ & 91.2$_{3.2}$ \\

\midrule

& &
\multicolumn{17}{c}{\textbf{Llama 3.1 70B}} \\

\cmidrule(lr){3-19}

& &
\multicolumn{5}{c}{\textbf{MuSR MM (N: 250)}} & & \multicolumn{5}{c}{\textbf{MoreHopQA (N: 150)}} & & \multicolumn{5}{c}{\textbf{iLLC L2-4 (N: 250)}} \\

\cmidrule(lr){3-7} \cmidrule(lr){9-13} \cmidrule(lr){15-19}

\textbf{Step} & &
\textbf{T$_{\mathrm{std}}$} & \textbf{\%T/N$_{\mathrm{std}}$} & \textbf{R$_{\mathrm{std}}$} & \textbf{\%R/T$_{\mathrm{std}}$} & \textbf{\%R/N$_{\mathrm{std}}$} & &
\textbf{T$_{\mathrm{std}}$} & \textbf{\%T/N$_{\mathrm{std}}$} & \textbf{R$_{\mathrm{std}}$} & \textbf{\%R/T$_{\mathrm{std}}$} & \textbf{\%R/N$_{\mathrm{std}}$} & &\textbf{T$_{\mathrm{std}}$} & \textbf{\%T/N$_{\mathrm{std}}$} & \textbf{R$_{\mathrm{std}}$} & \textbf{\%R/T$_{\mathrm{std}}$} & \textbf{\%R/N$_{\mathrm{std}}$} \\

\midrule

1$\rightarrow$2 & & 250.00$_{0.00}$ & 100.0$_{0.0}$ & 249.33$_{0.94}$ & 99.7$_{0.4}$ & 99.7$_{0.4}$ & & 150.00$_{0.00}$ & 100.0$_{0.0}$ & 124.00$_{3.32}$ & 82.7$_{2.2}$ & 82.7$_{2.2}$ & & 250.00$_{0.00}$ & 100.0$_{0.0}$ & 243.00$_{8.81}$ & 97.2$_{3.5}$ & 97.2$_{3.5}$ \\
2$\rightarrow$3 & & 250.00$_{0.00}$ & 100.0$_{0.0}$ & 249.83$_{0.37}$ & 99.9$_{0.1}$ & 99.9$_{0.1}$ & & 146.33$_{1.11}$ & 97.6$_{0.7}$ & 122.33$_{3.04}$ & 83.6$_{1.6}$ & 81.6$_{2.0}$ & & 250.00$_{0.00}$ & 100.0$_{0.0}$ & 238.00$_{6.71}$ & 95.2$_{2.7}$ & 95.2$_{2.7}$ \\
3$\rightarrow$4 & & 250.00$_{0.00}$ & 100.0$_{0.0}$ & 249.17$_{0.69}$ & 99.7$_{0.3}$ & 99.7$_{0.3}$ & & 133.00$_{2.83}$ & 88.7$_{1.9}$ & 110.50$_{3.95}$ & 83.1$_{1.6}$ & 73.7$_{2.6}$ & & 250.00$_{0.00}$ & 100.0$_{0.0}$ & 169.33$_{6.16}$ & 67.7$_{2.5}$ & 67.7$_{2.5}$ \\
4$\rightarrow$5 & & 250.00$_{0.00}$ & 100.0$_{0.0}$ & 249.00$_{1.41}$ & 99.6$_{0.6}$ & 99.6$_{0.6}$ & & 120.00$_{1.73}$ & 80.0$_{1.2}$ & 101.00$_{2.38}$ & 84.2$_{2.0}$ & 67.3$_{1.6}$ & & 238.67$_{3.09}$ & 95.5$_{1.2}$ & 171.17$_{5.79}$ & 71.7$_{2.4}$ & 68.5$_{2.3}$ \\
5$\rightarrow$6 & & 250.00$_{0.00}$ & 100.0$_{0.0}$ & 249.17$_{0.69}$ & 99.7$_{0.3}$ & 99.7$_{0.3}$ & & 108.67$_{2.98}$ & 72.4$_{2.0}$ & 96.17$_{2.91}$ & 88.5$_{1.9}$ & 64.1$_{1.9}$ & & 217.50$_{5.19}$ & 87.0$_{2.1}$ & 156.67$_{10.35}$ & 72.0$_{3.7}$ & 62.7$_{4.1}$ \\
6$\rightarrow$7 & & 250.00$_{0.00}$ & 100.0$_{0.0}$ & 248.00$_{1.53}$ & 99.2$_{0.6}$ & 99.2$_{0.6}$ & & 94.17$_{1.67}$ & 62.8$_{1.1}$ & 80.50$_{3.99}$ & 85.4$_{2.9}$ & 53.7$_{2.7}$ & & 185.00$_{4.65}$ & 74.0$_{1.9}$ & 140.17$_{4.37}$ & 75.8$_{1.3}$ & 56.1$_{1.7}$ \\
7$\rightarrow$8 & & 250.00$_{0.00}$ & 100.0$_{0.0}$ & 247.67$_{1.49}$ & 99.1$_{0.6}$ & 99.1$_{0.6}$ & & 82.33$_{1.97}$ & 54.9$_{1.3}$ & 66.67$_{2.62}$ & 81.0$_{3.0}$ & 44.4$_{1.7}$ & & 143.00$_{6.40}$ & 57.2$_{2.6}$ & 114.00$_{5.89}$ & 79.7$_{2.5}$ & 45.6$_{2.4}$ \\
8$\rightarrow$9 & & 250.00$_{0.00}$ & 100.0$_{0.0}$ & 246.50$_{1.71}$ & 98.6$_{0.7}$ & 98.6$_{0.7}$ & & 75.33$_{3.59}$ & 50.2$_{2.4}$ & 67.67$_{3.59}$ & 89.9$_{4.1}$ & 45.1$_{2.4}$ & & 114.17$_{5.01}$ & 45.7$_{2.0}$ & 99.00$_{5.60}$ & 86.7$_{2.2}$ & 39.6$_{2.2}$ \\
9$\rightarrow$10 & & 250.00$_{0.00}$ & 100.0$_{0.0}$ & 246.50$_{0.96}$ & 98.6$_{0.4}$ & 98.6$_{0.4}$ & & 67.33$_{3.04}$ & 44.9$_{2.0}$ & 60.17$_{3.58}$ & 89.3$_{2.4}$ & 40.1$_{2.4}$ & & 88.33$_{5.79}$ & 35.3$_{2.3}$ & 73.67$_{5.37}$ & 83.4$_{1.4}$ & 29.5$_{2.1}$ \\

\cdashlinelr{1-19}

ANY & & 250.00$_{0.00}$ & 100.0$_{0.0}$ & 250.00$_{0.00}$ & 100.0$_{0.0}$ & 100.0$_{0.0}$ & & 150.00$_{0.00}$ & 100.0$_{0.0}$ & 148.83$_{0.69}$ & 99.2$_{0.5}$ & 99.2$_{0.5}$ & & 250.00$_{0.00}$ & 100.0$_{0.0}$ & 250.00$_{0.00}$ & 100.0$_{0.0}$ & 100.0$_{0.0}$ \\

\midrule

& &
\multicolumn{17}{c}{\textbf{Mistral Small}} \\

\cmidrule(lr){3-19}

& &
\multicolumn{5}{c}{\textbf{MuSR MM (N: 250)}} & & \multicolumn{5}{c}{\textbf{MoreHopQA (N: 150)}} & & \multicolumn{5}{c}{\textbf{iLLC L2-4 (N: 250)}} \\

\cmidrule(lr){3-7} \cmidrule(lr){9-13} \cmidrule(lr){15-19}

\textbf{Step} & &
\textbf{T$_{\mathrm{std}}$} & \textbf{\%T/N$_{\mathrm{std}}$} & \textbf{R$_{\mathrm{std}}$} & \textbf{\%R/T$_{\mathrm{std}}$} & \textbf{\%R/N$_{\mathrm{std}}$} & &
\textbf{T$_{\mathrm{std}}$} & \textbf{\%T/N$_{\mathrm{std}}$} & \textbf{R$_{\mathrm{std}}$} & \textbf{\%R/T$_{\mathrm{std}}$} & \textbf{\%R/N$_{\mathrm{std}}$} & &\textbf{T$_{\mathrm{std}}$} & \textbf{\%T/N$_{\mathrm{std}}$} & \textbf{R$_{\mathrm{std}}$} & \textbf{\%R/T$_{\mathrm{std}}$} & \textbf{\%R/N$_{\mathrm{std}}$} \\

\midrule

1$\rightarrow$2 & & 250.00$_{0.00}$ & 100.0$_{0.0}$ & 246.33$_{1.37}$ & 98.5$_{0.5}$ & 98.5$_{0.5}$ & & 134.67$_{3.14}$ & 89.8$_{2.1}$ & 106.50$_{6.40}$ & 79.1$_{4.4}$ & 71.0$_{4.3}$ & & 250.00$_{0.00}$ & 100.0$_{0.0}$ & 169.33$_{21.30}$ & 67.7$_{8.5}$ & 67.7$_{8.5}$ \\
2$\rightarrow$3 & & 249.83$_{0.37}$ & 99.9$_{0.1}$ & 244.67$_{2.05}$ & 97.9$_{0.9}$ & 97.9$_{0.8}$ & & 114.83$_{1.77}$ & 76.6$_{1.2}$ & 78.50$_{1.89}$ & 68.4$_{1.3}$ & 52.3$_{1.3}$ & & 213.67$_{11.59}$ & 85.5$_{4.6}$ & 103.83$_{11.82}$ & 48.6$_{5.2}$ & 41.5$_{4.7}$ \\
3$\rightarrow$4 & & 249.00$_{0.82}$ & 99.6$_{0.3}$ & 239.17$_{2.79}$ & 96.0$_{0.9}$ & 95.7$_{1.1}$ & & 91.83$_{2.34}$ & 61.2$_{1.6}$ & 66.83$_{4.22}$ & 72.8$_{3.7}$ & 44.6$_{2.8}$ & & 196.67$_{7.41}$ & 78.7$_{3.0}$ & 58.67$_{17.30}$ & 30.0$_{9.2}$ & 23.5$_{6.9}$ \\
4$\rightarrow$5 & & 249.83$_{0.37}$ & 99.9$_{0.1}$ & 241.33$_{2.36}$ & 96.6$_{0.9}$ & 96.5$_{0.9}$ & & 69.50$_{3.82}$ & 46.3$_{2.5}$ & 56.00$_{3.37}$ & 80.6$_{2.4}$ & 37.3$_{2.2}$ & & 189.67$_{10.66}$ & 75.9$_{4.3}$ & 38.83$_{4.74}$ & 20.6$_{3.2}$ & 15.5$_{1.9}$ \\
5$\rightarrow$6 & & 249.83$_{0.37}$ & 99.9$_{0.1}$ & 235.17$_{2.11}$ & 94.1$_{0.9}$ & 94.1$_{0.8}$ & & 56.83$_{1.46}$ & 37.9$_{1.0}$ & 45.00$_{2.16}$ & 79.2$_{2.9}$ & 30.0$_{1.4}$ & & 105.33$_{13.05}$ & 42.1$_{5.2}$ & 36.50$_{3.25}$ & 35.2$_{5.2}$ & 14.6$_{1.3}$ \\
6$\rightarrow$7 & & 249.67$_{0.47}$ & 99.9$_{0.2}$ & 229.33$_{7.27}$ & 91.9$_{2.9}$ & 91.7$_{2.9}$ & & 41.67$_{1.11}$ & 27.8$_{0.7}$ & 33.50$_{2.06}$ & 80.4$_{3.7}$ & 22.3$_{1.4}$ & & 59.67$_{10.06}$ & 23.9$_{4.0}$ & 30.83$_{5.79}$ & 51.8$_{6.2}$ & 12.3$_{2.3}$ \\
7$\rightarrow$8 & & 249.67$_{0.47}$ & 99.9$_{0.2}$ & 230.83$_{4.45}$ & 92.5$_{1.7}$ & 92.3$_{1.8}$ & & 28.50$_{1.89}$ & 19.0$_{1.3}$ & 21.83$_{2.11}$ & 76.5$_{3.5}$ & 14.6$_{1.4}$ & & 38.83$_{7.73}$ & 15.5$_{3.1}$ & 22.00$_{3.70}$ & 57.1$_{5.7}$ & 8.8$_{1.5}$ \\
8$\rightarrow$9 & & 249.50$_{0.50}$ & 99.8$_{0.2}$ & 231.17$_{5.79}$ & 92.7$_{2.4}$ & 92.5$_{2.3}$ & & 23.83$_{1.07}$ & 15.9$_{0.7}$ & 18.17$_{1.95}$ & 76.1$_{6.6}$ & 12.1$_{1.3}$ & & 30.33$_{8.48}$ & 12.1$_{3.4}$ & 13.67$_{4.68}$ & 45.4$_{10.2}$ & 5.5$_{1.9}$ \\
9$\rightarrow$10 & & 249.33$_{0.47}$ & 99.7$_{0.2}$ & 229.50$_{2.87}$ & 92.0$_{1.3}$ & 91.8$_{1.1}$ & & 14.67$_{1.49}$ & 9.8$_{1.0}$ & 10.33$_{1.37}$ & 70.7$_{8.1}$ & 6.9$_{0.9}$ & & 22.17$_{8.13}$ & 8.9$_{3.3}$ & 12.50$_{5.32}$ & 55.8$_{5.2}$ & 5.0$_{2.1}$ \\

\cdashlinelr{1-19}

ANY & & 250.00$_{0.00}$ & 100.0$_{0.0}$ & 250.00$_{0.00}$ & 100.0$_{0.0}$ & 100.0$_{0.0}$ & & 141.00$_{1.15}$ & 94.0$_{0.8}$ & 130.17$_{3.98}$ & 92.3$_{2.6}$ & 86.8$_{2.7}$ & & 250.00$_{0.00}$ & 100.0$_{0.0}$ & 195.83$_{10.84}$ & 78.3$_{4.3}$ & 78.3$_{4.3}$ \\

\midrule

& &
\multicolumn{17}{c}{\textbf{Phi-4}} \\

\cmidrule(lr){3-19}

& &
\multicolumn{5}{c}{\textbf{MuSR MM (N: 250)}} & & \multicolumn{5}{c}{\textbf{MoreHopQA (N: 150)}} & & \multicolumn{5}{c}{\textbf{iLLC L2-4 (N: 250)}} \\

\cmidrule(lr){3-7} \cmidrule(lr){9-13} \cmidrule(lr){15-19}

\textbf{Step} & &
\textbf{T$_{\mathrm{std}}$} & \textbf{\%T/N$_{\mathrm{std}}$} & \textbf{R$_{\mathrm{std}}$} & \textbf{\%R/T$_{\mathrm{std}}$} & \textbf{\%R/N$_{\mathrm{std}}$} & &
\textbf{T$_{\mathrm{std}}$} & \textbf{\%T/N$_{\mathrm{std}}$} & \textbf{R$_{\mathrm{std}}$} & \textbf{\%R/T$_{\mathrm{std}}$} & \textbf{\%R/N$_{\mathrm{std}}$} & &\textbf{T$_{\mathrm{std}}$} & \textbf{\%T/N$_{\mathrm{std}}$} & \textbf{R$_{\mathrm{std}}$} & \textbf{\%R/T$_{\mathrm{std}}$} & \textbf{\%R/N$_{\mathrm{std}}$} \\

\midrule

1$\rightarrow$2 & & 250.00$_{0.00}$ & 100.0$_{0.0}$ & 237.00$_{3.32}$ & 94.8$_{1.3}$ & 94.8$_{1.3}$ & & 149.67$_{0.47}$ & 99.8$_{0.3}$ & 117.33$_{2.56}$ & 78.4$_{1.7}$ & 78.2$_{1.7}$ & & 250.00$_{0.00}$ & 100.0$_{0.0}$ & 82.50$_{32.41}$ & 33.0$_{13.0}$ & 33.0$_{13.0}$ \\
2$\rightarrow$3 & & 250.00$_{0.00}$ & 100.0$_{0.0}$ & 234.50$_{3.86}$ & 93.8$_{1.5}$ & 93.8$_{1.5}$ & & 138.00$_{2.08}$ & 92.0$_{1.4}$ & 113.83$_{2.73}$ & 82.5$_{1.5}$ & 75.9$_{1.8}$ & & 250.00$_{0.00}$ & 100.0$_{0.0}$ & 19.83$_{3.53}$ & 7.9$_{1.4}$ & 7.9$_{1.4}$ \\
3$\rightarrow$4 & & 249.83$_{0.37}$ & 99.9$_{0.1}$ & 234.83$_{1.95}$ & 94.0$_{0.8}$ & 93.9$_{0.8}$ & & 124.50$_{3.50}$ & 83.0$_{2.3}$ & 100.00$_{3.21}$ & 80.3$_{1.1}$ & 66.7$_{2.1}$ & & 250.00$_{0.00}$ & 100.0$_{0.0}$ & 15.17$_{3.44}$ & 6.1$_{1.4}$ & 6.1$_{1.4}$ \\
4$\rightarrow$5 & & 249.83$_{0.37}$ & 99.9$_{0.1}$ & 236.33$_{3.04}$ & 94.6$_{1.2}$ & 94.5$_{1.2}$ & & 111.17$_{4.56}$ & 74.1$_{3.0}$ & 91.50$_{4.07}$ & 82.3$_{2.8}$ & 61.0$_{2.7}$ & & 250.00$_{0.00}$ & 100.0$_{0.0}$ & 11.50$_{3.10}$ & 4.6$_{1.2}$ & 4.6$_{1.2}$ \\
5$\rightarrow$6 & & 249.83$_{0.37}$ & 99.9$_{0.1}$ & 236.00$_{5.03}$ & 94.5$_{2.0}$ & 94.4$_{2.0}$ & & 100.17$_{3.02}$ & 66.8$_{2.0}$ & 84.00$_{3.87}$ & 83.8$_{1.5}$ & 56.0$_{2.6}$ & & 249.00$_{0.58}$ & 99.6$_{0.2}$ & 200.50$_{8.26}$ & 80.5$_{3.4}$ & 80.2$_{3.3}$ \\
6$\rightarrow$7 & & 249.83$_{0.37}$ & 99.9$_{0.1}$ & 234.00$_{4.00}$ & 93.7$_{1.5}$ & 93.6$_{1.6}$ & & 87.00$_{0.82}$ & 58.0$_{0.5}$ & 74.67$_{2.81}$ & 85.8$_{3.4}$ & 49.8$_{1.9}$ & & 53.67$_{6.97}$ & 21.5$_{2.8}$ & 42.00$_{7.28}$ & 77.9$_{5.6}$ & 16.8$_{2.9}$ \\
7$\rightarrow$8 & & 249.83$_{0.37}$ & 99.9$_{0.1}$ & 232.67$_{3.90}$ & 93.1$_{1.6}$ & 93.1$_{1.6}$ & & 76.67$_{3.14}$ & 51.1$_{2.1}$ & 62.50$_{2.57}$ & 81.7$_{4.7}$ & 41.7$_{1.7}$ & & 19.67$_{4.23}$ & 7.9$_{1.7}$ & 11.33$_{3.25}$ & 57.2$_{10.5}$ & 4.5$_{1.3}$ \\
8$\rightarrow$9 & & 249.83$_{0.37}$ & 99.9$_{0.1}$ & 230.33$_{1.97}$ & 92.2$_{0.7}$ & 92.1$_{0.8}$ & & 68.00$_{2.77}$ & 45.3$_{1.8}$ & 57.50$_{3.82}$ & 84.5$_{4.0}$ & 38.3$_{2.5}$ & & 11.67$_{2.92}$ & 4.7$_{1.2}$ & 7.67$_{3.82}$ & 62.9$_{13.5}$ & 3.1$_{1.5}$ \\
9$\rightarrow$10 & & 249.83$_{0.37}$ & 99.9$_{0.1}$ & 228.83$_{4.88}$ & 91.6$_{2.0}$ & 91.5$_{2.0}$ & & 54.17$_{5.81}$ & 36.1$_{3.9}$ & 45.67$_{6.10}$ & 84.2$_{5.5}$ & 30.4$_{4.1}$ & & 6.17$_{1.46}$ & 2.5$_{0.6}$ & 5.50$_{1.61}$ & 89.0$_{12.6}$ & 2.2$_{0.6}$ \\

\cdashlinelr{1-19}

ANY & & 250.00$_{0.00}$ & 100.0$_{0.0}$ & 250.00$_{0.00}$ & 100.0$_{0.0}$ & 100.0$_{0.0}$ & & 149.83$_{0.37}$ & 99.9$_{0.2}$ & 144.67$_{1.70}$ & 96.6$_{1.1}$ & 96.4$_{1.1}$ & & 250.00$_{0.00}$ & 100.0$_{0.0}$ & 244.17$_{2.19}$ & 97.7$_{0.9}$ & 97.7$_{0.9}$ \\

\midrule

& &
\multicolumn{17}{c}{\textbf{Qwen3}} \\

\cmidrule(lr){3-19}

& &
\multicolumn{5}{c}{\textbf{MuSR MM (N: 250)}} & & \multicolumn{5}{c}{\textbf{MoreHopQA (N: 150)}} & & \multicolumn{5}{c}{\textbf{iLLC L2-4 (N: 250)}} \\

\cmidrule(lr){3-7} \cmidrule(lr){9-13} \cmidrule(lr){15-19}

\textbf{Step} & &
\textbf{T$_{\mathrm{std}}$} & \textbf{\%T/N$_{\mathrm{std}}$} & \textbf{R$_{\mathrm{std}}$} & \textbf{\%R/T$_{\mathrm{std}}$} & \textbf{\%R/N$_{\mathrm{std}}$} & &
\textbf{T$_{\mathrm{std}}$} & \textbf{\%T/N$_{\mathrm{std}}$} & \textbf{R$_{\mathrm{std}}$} & \textbf{\%R/T$_{\mathrm{std}}$} & \textbf{\%R/N$_{\mathrm{std}}$} & &\textbf{T$_{\mathrm{std}}$} & \textbf{\%T/N$_{\mathrm{std}}$} & \textbf{R$_{\mathrm{std}}$} & \textbf{\%R/T$_{\mathrm{std}}$} & \textbf{\%R/N$_{\mathrm{std}}$} \\

\midrule

1$\rightarrow$2 & & 250.00$_{0.00}$ & 100.0$_{0.0}$ & 249.33$_{0.75}$ & 99.7$_{0.3}$ & 99.7$_{0.3}$ & & 150.00$_{0.00}$ & 100.0$_{0.0}$ & 110.50$_{2.69}$ & 73.7$_{1.8}$ & 73.7$_{1.8}$ & & 250.00$_{0.00}$ & 100.0$_{0.0}$ & 226.00$_{8.21}$ & 90.4$_{3.3}$ & 90.4$_{3.3}$ \\
2$\rightarrow$3 & & 250.00$_{0.00}$ & 100.0$_{0.0}$ & 240.67$_{1.97}$ & 96.3$_{0.8}$ & 96.3$_{0.8}$ & & 130.83$_{2.73}$ & 87.2$_{1.8}$ & 91.17$_{6.26}$ & 69.7$_{4.3}$ & 60.8$_{4.2}$ & & 250.00$_{0.00}$ & 100.0$_{0.0}$ & 86.83$_{6.52}$ & 34.7$_{2.6}$ & 34.7$_{2.6}$ \\
3$\rightarrow$4 & & 250.00$_{0.00}$ & 100.0$_{0.0}$ & 239.33$_{3.30}$ & 95.7$_{1.3}$ & 95.7$_{1.3}$ & & 100.50$_{1.89}$ & 67.0$_{1.3}$ & 68.00$_{4.16}$ & 67.7$_{3.8}$ & 45.3$_{2.8}$ & & 230.17$_{11.98}$ & 92.1$_{4.8}$ & 25.67$_{5.73}$ & 11.2$_{2.5}$ & 10.3$_{2.3}$ \\
4$\rightarrow$5 & & 249.83$_{0.37}$ & 99.9$_{0.1}$ & 238.00$_{3.32}$ & 95.3$_{1.4}$ & 95.2$_{1.3}$ & & 84.33$_{3.50}$ & 56.2$_{2.3}$ & 55.83$_{3.34}$ & 66.2$_{2.7}$ & 37.2$_{2.2}$ & & 233.67$_{8.79}$ & 93.5$_{3.5}$ & 13.67$_{4.23}$ & 5.9$_{1.8}$ & 5.5$_{1.7}$ \\
5$\rightarrow$6 & & 248.83$_{1.34}$ & 99.5$_{0.5}$ & 235.33$_{4.61}$ & 94.6$_{1.5}$ & 94.1$_{1.8}$ & & 66.83$_{6.04}$ & 44.6$_{4.0}$ & 43.50$_{3.82}$ & 65.1$_{0.7}$ & 29.0$_{2.5}$ & & 181.50$_{4.72}$ & 72.6$_{1.9}$ & 74.67$_{4.82}$ & 41.1$_{2.1}$ & 29.9$_{1.9}$ \\
6$\rightarrow$7 & & 247.83$_{0.69}$ & 99.1$_{0.3}$ & 235.00$_{3.11}$ & 94.8$_{1.3}$ & 94.0$_{1.2}$ & & 47.83$_{3.02}$ & 31.9$_{2.0}$ & 33.50$_{4.03}$ & 70.0$_{6.7}$ & 22.3$_{2.7}$ & & 83.00$_{6.53}$ & 33.2$_{2.6}$ & 48.33$_{5.99}$ & 58.4$_{7.4}$ & 19.3$_{2.4}$ \\
7$\rightarrow$8 & & 245.50$_{1.38}$ & 98.2$_{0.6}$ & 228.17$_{4.88}$ & 92.9$_{1.6}$ & 91.3$_{2.0}$ & & 33.83$_{3.08}$ & 22.6$_{2.1}$ & 24.00$_{4.76}$ & 70.7$_{11.7}$ & 16.0$_{3.2}$ & & 22.33$_{5.50}$ & 8.9$_{2.2}$ & 10.33$_{2.75}$ & 46.6$_{7.2}$ & 4.1$_{1.1}$ \\
8$\rightarrow$9 & & 242.17$_{1.46}$ & 96.9$_{0.6}$ & 226.50$_{2.50}$ & 93.5$_{0.8}$ & 90.6$_{1.0}$ & & 21.83$_{2.03}$ & 14.6$_{1.4}$ & 16.33$_{3.73}$ & 74.1$_{12.0}$ & 10.9$_{2.5}$ & & 9.83$_{3.08}$ & 3.9$_{1.2}$ & 4.67$_{2.05}$ & 49.7$_{16.8}$ & 1.9$_{0.8}$ \\
9$\rightarrow$10 & & 239.83$_{2.67}$ & 95.9$_{1.1}$ & 225.00$_{3.00}$ & 93.8$_{1.0}$ & 90.0$_{1.2}$ & & 16.67$_{3.25}$ & 11.1$_{2.2}$ & 12.67$_{2.69}$ & 75.9$_{7.4}$ & 8.4$_{1.8}$ & & 4.33$_{2.13}$ & 1.7$_{0.9}$ & 1.50$_{0.96}$ & 32.2$_{22.8}$ & 0.6$_{0.4}$ \\

\cdashlinelr{1-19}

ANY & & 250.00$_{0.00}$ & 100.0$_{0.0}$ & 250.00$_{0.00}$ & 100.0$_{0.0}$ & 100.0$_{0.0}$ & & 150.00$_{0.00}$ & 100.0$_{0.0}$ & 141.00$_{1.53}$ & 94.0$_{1.0}$ & 94.0$_{1.0}$ & & 250.00$_{0.00}$ & 100.0$_{0.0}$ & 240.33$_{0.94}$ & 96.1$_{0.4}$ & 96.1$_{0.4}$ \\

\bottomrule

\end{tabular}
}
\end{table*}

%% file: tables/temp.tex
\begin{table*}[htb]
\centering
\Large
\caption{
Accuracy (\%) of ReHoPER under different sampling temperatures (T) on iLLC~L2-4 (development set) and MoreHopQA, using IO, CoT, and PS answer-format prefixes. \colorbox{green!40}{Green} and \colorbox{green!15}{light green} mark the best and second best result, respectively, for each (model, dataset, prompting-style) triplet.
}
\label{tab:temp}
\resizebox{\textwidth}{!}{
\begin{tabular}{
l p{0pt}
cccc p{0pt} cccc p{0pt} cccc p{0pt}
cccc p{0pt} cccc p{0pt} cccc
}

\toprule

& &
\multicolumn{14}{c}{\textbf{iLLC~L2-4 (Development Set)}} & &
\multicolumn{14}{c}{\textbf{MoreHopQA}} \\

\cmidrule(lr){3-16}
\cmidrule(lr){18-31}

& &
\multicolumn{4}{c}{\textbf{IO}} & & \multicolumn{4}{c}{\textbf{CoT}} & & \multicolumn{4}{c}{\textbf{PS}} & &
\multicolumn{4}{c}{\textbf{IO}} & & \multicolumn{4}{c}{\textbf{CoT}} & & \multicolumn{4}{c}{\textbf{PS}} \\

\cmidrule(lr){3-6} \cmidrule(lr){8-11} \cmidrule(lr){13-16}
\cmidrule(lr){18-21} \cmidrule(lr){23-26} \cmidrule(lr){28-31}

\textbf{Model} & &
\textbf{T=.2} & \textbf{T=.3} & \textbf{T=.5} & \textbf{T=.7} & &
\textbf{T=.2} & \textbf{T=.3} & \textbf{T=.5} & \textbf{T=.7} & &
\textbf{T=.2} & \textbf{T=.3} & \textbf{T=.5} & \textbf{T=.7} & &
\textbf{T=.2} & \textbf{T=.3} & \textbf{T=.5} & \textbf{T=.7} & &
\textbf{T=.2} & \textbf{T=.3} & \textbf{T=.5} & \textbf{T=.7} & &
\textbf{T=.2} & \textbf{T=.3} & \textbf{T=.5} & \textbf{T=.7} \\

\midrule

Gemma 3 & &
\smallgain{6.8} & \smallgain{6.8} & 5.6 & \gain{7.6} & &
5.6 & \smallgain{7.6} & 5.6 & \gain{8.4} & &
4.0 & 4.8 & \smallgain{6.0} & \gain{6.8} & &
64.7 & \gain{67.3} & 64.0 & \smallgain{66.0} & &
66.0 & 66.0 & \smallgain{66.7} & \gain{67.3} & &
70.7 & 71.3 & \gain{74.7} & \smallgain{73.3} \\

gpt-oss & &
90.8 & 92.4 & \smallgain{93.2} & \gain{94.8} & &
89.2 & 92.0 & \smallgain{93.6} & \gain{94.8} & &
89.6 & \smallgain{92.8} & 91.2 & \gain{93.6} & &
78.7 & 78.7 & \gain{81.3} & \smallgain{80.0} & &
78.0 & 79.3 & \gain{80.0} & \gain{80.0} & &
80.0 & \smallgain{80.7} & \gain{82.0} & 79.3 \\

Llama 3.1 8B & &
10.8 & 8.8 & \smallgain{16.8} & \gain{20.0} & &
\smallgain{16.4} & 12.4 & \gain{18.4} & 11.2 & &
\smallgain{40.4} & 30.0 & 38.4 & \gain{42.4} & &
\smallgain{57.3} & \gain{58.7} & 56.7 & 55.3 & &
56.0 & 57.3 & \smallgain{58.0} & \gain{62.0} & &
58.7 & 60.7 & \smallgain{63.3} & \gain{64.7} \\

Mistral Small & &
\smallgain{8.0} & 7.2 & \gain{9.2} & 7.2 & &
5.6 & \smallgain{8.8} & \gain{9.2} & 6.0 & &
\gain{8.8} & \smallgain{8.4} & \smallgain{8.4} & 6.4 & &
66.0 & \smallgain{70.7} & \gain{71.3} & 70.0 & &
67.3 & \smallgain{68.0} & 67.3 & \gain{72.0} & &
\gain{70.7} & \gain{70.7} & \gain{70.7} & 68.0 \\

Phi-4 & &
\smallgain{42.4} & 41.6 & \gain{48.4} & 36.4 & &
\smallgain{58.4} & 54.4 & \gain{64.0} & 54.8 & &
\smallgain{65.6} & 61.2 & \gain{67.6} & 58.4 & &
71.3 & 72.7 & \gain{78.7} & \smallgain{78.0} & &
73.3 & 77.3 & \gain{80.7} & \smallgain{80.0} & &
75.3 & 76.0 & \gain{80.0} & \smallgain{78.7} \\

Qwen3 w/o Thinking & &
80.0 & 82.8 & \smallgain{84.4} & \gain{84.8} & &
84.0 & 83.6 & \smallgain{87.2} & \gain{88.0} & &
87.6 & 87.2 & \gain{90.4} & \gain{90.4} & &
\gain{71.3} & \smallgain{70.7} & 68.7 & 68.7 & &
\gain{73.3} & \smallgain{72.0} & 70.0 & 69.3 & &
69.3 & \smallgain{71.3} & \gain{72.7} & 69.3 \\

\cdashlinelr{1-31}

Average & &
39.8 & 39.9 & \gain{42.9} & \smallgain{41.8} & &
43.2 & 43.1 & \gain{46.3} & \smallgain{43.9} & &
49.3 & 47.4 & \gain{50.3} & \smallgain{49.7} & &
68.2 & \smallgain{69.8} & \gain{70.1} & 69.7 & &
69.0 & 70.0 & \smallgain{70.4} & \gain{71.8} & &
70.8 & 71.8 & \gain{73.9} & \smallgain{72.2} \\

\bottomrule

\end{tabular}
}
\end{table*}

%% file: tables/pathstep.tex
\begin{table*}[htb]
\centering
\Large
\caption{
Accuracy (\%) of ReHoPER under different \{\(n\)\}-\{\(m_{\max}\)\} configurations on iLLC~L2-4 (development set) and MoreHopQA, using IO, CoT, and PS answer-format prefixes. \colorbox{green!40}{Green} and \colorbox{green!15}{light green} mark the best and second best result, respectively, for each (model, dataset, prompting-style) triplet.
}
\label{tab:pathstep}
\resizebox{\textwidth}{!}{
\begin{tabular}{
l p{0pt}
cccc p{5pt} cccc p{5pt} cccc p{5pt}
cccc p{5pt} cccc p{5pt} cccc
}

\toprule

& &
\multicolumn{14}{c}{\textbf{iLLC~L2-4 (Development Set)}} & &
\multicolumn{14}{c}{\textbf{MoreHopQA}} \\

\cmidrule(lr){3-16}
\cmidrule(lr){18-31}

& &
\multicolumn{4}{c}{\textbf{IO}} & & \multicolumn{4}{c}{\textbf{CoT}} & & \multicolumn{4}{c}{\textbf{PS}} & &
\multicolumn{4}{c}{\textbf{IO}} & & \multicolumn{4}{c}{\textbf{CoT}} & & \multicolumn{4}{c}{\textbf{PS}} \\

\cmidrule(lr){3-6} \cmidrule(lr){8-11} \cmidrule(lr){13-16}
\cmidrule(lr){18-21} \cmidrule(lr){23-26} \cmidrule(lr){28-31}

\textbf{Model} & &
\textbf{5-12} & \textbf{6-10} & \textbf{10-6} & \textbf{12-5} & &
\textbf{5-12} & \textbf{6-10} & \textbf{10-6} & \textbf{12-5} & &
\textbf{5-12} & \textbf{6-10} & \textbf{10-6} & \textbf{12-5} & &
\textbf{5-12} & \textbf{6-10} & \textbf{10-6} & \textbf{12-5} & &
\textbf{5-12} & \textbf{6-10} & \textbf{10-6} & \textbf{12-5} & &
\textbf{5-12} & \textbf{6-10} & \textbf{10-6} & \textbf{12-5} \\

\midrule

Gemma 3 & &
\smallgain{6.0} & \gain{6.8} & 5.6 & 4.8 & &
\gain{8.0} & \gain{8.0} & 5.6 & 5.6 & &
\smallgain{6.4} & \gain{7.6} & 6.0 & 5.6 & &
\gain{66.0} & \smallgain{64.7} & 64.0 & 62.7 & &
\gain{66.7} & 64.7 & \gain{66.7} & 64.0 & &
74.0 & \gain{74.7} & \gain{74.7} & 74.0 \\

gpt-oss & &
94.0 & \smallgain{94.4} & 93.2 & \gain{94.8} & &
\gain{93.6} & \gain{93.6} & \gain{93.6} & \gain{93.6} & &
\smallgain{93.2} & \gain{94.4} & 91.2 & 92.8 & &
78.0 & 78.0 & \gain{81.3} & \smallgain{80.0} & &
78.0 & \gain{81.3} & \smallgain{80.0} & 79.3 & &
79.3 & \gain{82.0} & \gain{82.0} & 78.7 \\

Llama 3.1 8B & &
\gain{23.6} & \smallgain{21.2} & 16.8 & 10.8 & &
\smallgain{19.6} & \gain{20.8} & 18.4 & 16.0 & &
\smallgain{38.4} & \gain{40.0} & \smallgain{38.4} & 33.6 & &
54.7 & \smallgain{56.0} & \gain{56.7} & 54.7 & &
\gain{60.7} & \smallgain{59.3} & 58.0 & 56.0 & &
\gain{65.3} & \smallgain{63.3} & \smallgain{63.3} & 61.3 \\

Mistral Small & &
\gain{10.0} & \gain{10.0} & 9.2 & 5.6 & &
\gain{12.8} & \gain{12.8} & 9.2 & 8.0 & &
\gain{11.2} & \smallgain{9.6} & 8.4 & 6.4 & &
70.7 & \smallgain{71.3} & \smallgain{71.3} & \gain{72.0} & &
\gain{67.3} & \gain{67.3} & \gain{67.3} & \gain{67.3} & &
\gain{71.3} & 70.7 & 70.7 & \gain{71.3} \\

Phi-4 & &
\gain{52.8} & \smallgain{51.6} & 48.4 & 34.8 & &
\gain{66.4} & \smallgain{66.0} & 64.0 & 64.0 & &
\gain{68.0} & \smallgain{67.6} & \smallgain{67.6} & 64.0 & &
76.7 & \smallgain{78.0} & \gain{78.7} & 77.3 & &
\smallgain{79.3} & 78.7 & \gain{80.7} & 78.7 & &
78.7 & \smallgain{79.3} & \gain{80.0} & 78.0 \\

Qwen3 w/o Thinking & &
\gain{84.4} & \gain{84.4} & \gain{84.4} & 80.8 & &
\gain{87.6} & 86.8 & \smallgain{87.2} & 86.4 & &
88.0 & \gain{90.8} & \smallgain{90.4} & \smallgain{90.4} & &
\gain{70.0} & 67.3 & \smallgain{68.7} & 67.3 & &
68.7 & \smallgain{70.7} & 70.0 & \gain{72.0} & &
\gain{73.3} & 71.3 & \smallgain{72.7} & 71.3 \\

\cdashlinelr{1-31}

Average & &
\gain{45.1} & \smallgain{44.7} & 42.9 & 38.6 & &
\gain{48.0} & \gain{48.0} & 46.3 & 45.6 & &
\smallgain{50.9} & \gain{51.7} & 50.3 & 48.8 & &
\smallgain{69.3} & 69.2 & \gain{70.1} & 69.0 & &
70.1 & \smallgain{70.3} & \gain{70.4} & 69.6 & &
\smallgain{73.7} & 73.6 & \gain{73.9} & 72.4 \\

\bottomrule

\end{tabular}
}
\end{table*}

%% file: tables/robustness.tex
\begin{table*}[htb]
\centering
\large
\caption{
Robustness of ReHoPER (RH) across repeated runs on MoreHopQA and iLLC L2-4 under IO, CoT, and PS answer-format prefixes. We report accuracy (\%) as mean$_{\mathrm{std}}$ over three runs. SC denotes simple question answering with Self-Consistency using the corresponding output prefix, while RH applies our planning approach but evaluates with the same prefix. \colorbox{green!40}{Green}, \colorbox{red!25}{red}, and \colorbox{gray!25}{gray} indicate performance changes relative to SC.
}
\label{tab:robustness}
\resizebox{\textwidth}{!}{
\begin{tabular}{l p{0pt} cccccc p{5pt} cccccc p{5pt} cccccc}

\toprule

& &
\multicolumn{6}{c}{\textbf{MoreHopQA}} & &
\multicolumn{6}{c}{\textbf{iLLC L2-4}} \\

\cmidrule(lr){3-8}
\cmidrule(lr){10-15}

& &
\multicolumn{2}{c}{\textbf{IO}} & \multicolumn{2}{c}{\textbf{CoT}} & \multicolumn{2}{c}{\textbf{PS}} & &
\multicolumn{2}{c}{\textbf{IO}} & \multicolumn{2}{c}{\textbf{CoT}} & \multicolumn{2}{c}{\textbf{PS}} \\

\cmidrule(lr){3-4} \cmidrule(lr){5-6} \cmidrule(lr){7-8}
\cmidrule(lr){10-11} \cmidrule(lr){12-13} \cmidrule(lr){14-15}

\textbf{Model} & &
\textbf{SC} & \textbf{RH} &
\textbf{SC} & \textbf{RH} &
\textbf{SC} & \textbf{RH} & &
\textbf{SC} & \textbf{RH} &
\textbf{SC} & \textbf{RH} &
\textbf{SC} & \textbf{RH} \\

\midrule

Gemma 3 & &
61.8$_{0.4}$ & \gain{65.8$_{1.0}$} &
64.7$_{0.7}$ & \gain{68.7$_{0.0}$} &
66.7$_{0.7}$ & \gain{74.7$_{1.2}$} & &
7.3$_{0.2}$ & \loss{6.9$_{1.3}$} &
7.6$_{0.0}$ & \loss{6.9$_{0.9}$} &
3.6$_{0.8}$ & \gain{8.1$_{1.7}$} \\

Llama 3.1 8B & &
60.7$_{1.2}$ & \loss{57.1$_{1.7}$} &
60.0$_{0.7}$ & \same{60.0$_{1.8}$} &
62.4$_{1.5}$ & \gain{63.1$_{0.4}$} & &
11.9$_{0.8}$ & \gain{17.5$_{4.1}$} &
12.5$_{3.2}$ & \gain{18.4$_{2.0}$} &
30.4$_{1.1}$ & \gain{41.6$_{1.4}$} \\

Mistral Small & &
70.2$_{0.4}$ & \gain{71.3$_{0.0}$} &
69.1$_{1.0}$ & \loss{68.7$_{1.2}$} &
72.2$_{0.4}$ & \loss{71.1$_{1.4}$} & &
1.9$_{0.5}$ & \gain{10.5$_{0.6}$} &
2.4$_{0.4}$ & \gain{9.6$_{1.7}$} &
3.1$_{0.8}$ & \gain{10.8$_{1.2}$} \\

Phi-4 & &
77.3$_{0.7}$ & \loss{76.0$_{1.8}$} &
76.4$_{0.4}$ & \gain{80.4$_{1.0}$} &
77.6$_{0.4}$ & \gain{78.4$_{1.7}$} & &
13.2$_{0.4}$ & \gain{54.3$_{0.8}$} &
10.0$_{0.4}$ & \gain{62.0$_{1.7}$} &
9.7$_{0.8}$ & \gain{62.7$_{1.3}$} \\

\cdashlinelr{1-15}

Average & &
67.5$_{0.7}$ & \gain{67.6$_{1.1}$} &
67.6$_{0.7}$ & \gain{69.5$_{1.0}$} &
69.7$_{0.8}$ & \gain{71.9$_{1.2}$} & &
8.6$_{0.5}$ & \gain{22.3$_{1.7}$} &
8.1$_{1.0}$ & \gain{24.2$_{1.6}$} &
11.7$_{0.9}$ & \gain{30.8$_{1.4}$} \\

\bottomrule

\end{tabular}
}
\end{table*}

%% file: tables/illc.tex
\begin{table*}[htb]
\centering
\caption{
Accuracy (\%) of ReHoPER (RH) across iLLC L1-8 and iLLC L2-8 benchmarks under IO, CoT, and PS answer-format prefixes. SD refers to the Self-Discover baseline. Furthermore, SC denotes simple question answering with self-consistency using the corresponding output prefix, while RH applies our planning approach but evaluates with the same prefix.
\colorbox{green!40}{Green}, \colorbox{red!25}{red}, and \colorbox{gray!25}{gray} indicate performance changes relative to SC.
}
\label{tab:illc}
\resizebox{.9\columnwidth}{!}{
\begin{tabular}{l p{0pt} c c cccccc p{0pt} c c cccccc}

\toprule

& &
\multicolumn{8}{c}{\textbf{iLLC L1-8}} &
& \multicolumn{8}{c}{\textbf{iLLC L2-8}} \\

\cmidrule(lr){3-10} \cmidrule(lr){12-19}

& &
& &
\multicolumn{2}{c}{\textbf{IO}} &
\multicolumn{2}{c}{\textbf{CoT}} &
\multicolumn{2}{c}{\textbf{PS}} &
&
& &
\multicolumn{2}{c}{\textbf{IO}} &
\multicolumn{2}{c}{\textbf{CoT}} &
\multicolumn{2}{c}{\textbf{PS}} \\

\cmidrule(lr){5-6} \cmidrule(lr){7-8} \cmidrule(lr){9-10}
\cmidrule(lr){14-15} \cmidrule(lr){16-17} \cmidrule(lr){18-19}

\textbf{Model} & &
\textbf{SD} & &
\textbf{SC} & \textbf{RH} &
\textbf{SC} & \textbf{RH} &
\textbf{SC} & \textbf{RH} &
&
\textbf{SD} & &
\textbf{SC} & \textbf{RH} &
\textbf{SC} & \textbf{RH} &
\textbf{SC} & \textbf{RH} \\

\midrule

Gemma 3 & &
26.4 & &
7.6  & \gain{12.4} &
36.0 & \loss{28.8} &
52.8 & \loss{32.8} &
&
0.0 & &
0.4 & \loss{0.0} &
0.0 & \same{0.0} &
0.0 & \same{0.0} \\

gpt-oss & &
58.8 & &
84.0 & \loss{76.4} &
78.4 & \loss{74.8} &
77.6 & \loss{76.0} &
&
45.6 & &
76.4 & \gain{80.0} &
61.6 & \gain{80.8} &
6.0  & \gain{81.2} \\

Llama 3.1 8B & &
7.6 & &
31.2 & \loss{27.6} &
24.0 & \gain{39.2} &
39.2 & \loss{34.4} &
&
0.0 & &
0.8 & \gain{2.0} &
0.0 & \gain{4.0} &
0.4 & \gain{2.4} \\

Llama 3.1 70B & &
56.0 & &
59.2 & \gain{62.8} &
48.8 & \gain{63.2} &
68.8 & \gain{70.4} &
&
0.8 & &
1.2  & \gain{44.0} &
3.6  & \gain{46.4} &
12.8 & \gain{54.0} \\

Mistral Small & &
23.6 & &
64.0 & \loss{57.2} &
62.4 & \loss{57.2} &
60.4 & \loss{59.6} &
&
0.0 & &
0.0 & \same{0.0} &
0.0 & \gain{0.4} &
0.0 & \same{0.0} \\

Phi-4 & &
53.6 & &
53.6 & \gain{55.6} &
60.0 & \gain{62.8} &
60.8 & \gain{61.2} &
&
2.0 & &
0.8 & \gain{16.8} &
1.2 & \gain{33.6} &
1.6 & \gain{35.6} \\

Qwen3 w/o Thinking & &
44.4 & &
50.8 & \gain{68.8} &
54.8 & \gain{67.2} &
69.2 & \gain{72.0} &
&
0.0 & &
15.2 & \gain{33.6} &
8.8  & \gain{32.8} &
60.0 & \loss{43.6} \\

Qwen3 w/ Thinking & &
1.2 & &
85.2 & \loss{80.4} &
54.8 & \gain{75.2} &
69.2 & \gain{78.8} &
&
3.6 & &
76.8 & \loss{72.8} &
8.8  & \gain{76.0} &
60.0 & \gain{76.4} \\

\cdashlinelr{1-19}

Average & &
34.0 & &
54.5 & \gain{55.2} &
52.4 & \gain{58.6} &
62.3 & \loss{60.7} &
&
6.5 & &
21.5 & \gain{31.2} &
10.5 & \gain{34.3} &
17.6 & \gain{36.7} \\

\bottomrule

\end{tabular}
}
\end{table*}

%% file: figures/data-examples.tex
\begin{softbox}[An example from MuSR MM]
In an adrenaline inducing bungee jumping site, Mack's thrill-seeking adventure came to a gruesome end by a nunchaku; now, it's up to Detective Winston to unravel the deadly secrets between Mackenzie and Ana.\\\\Winston took a gulp of his black coffee, staring at the notes sprawled across his desk. A murder case at a bungee jumping site was definitely out of the ordinary. Today's victim was a young man named Mack, loud mouthed and cocky by all accounts. \\\\Mack was bungee jumping the day he was killed. Oddly enough, according to the records, no one else was documented at the bungee jumping site that day, making this case even more peculiar. The first stop for the day was to visit one of Mack's housemates, a woman named Ana. They were seen leaving in the same vehicle from their shared housing complex the morning of the murder, and it was time for Winston to dig deeper. \\\\As he pulled into the shared housing driveway, a nondescript car came into sight. He learned from neighbours that it was frequently used by multiple residents, but Ana had a peculiar interest in it. She would insist on driving whenever with a group of friends, later meticulously cleaning the car after each use. An idiosyncrasy of hers maybe, but a part of the puzzle nonetheless.\\\\Winston knocked on the door, Ana opened it warily, twiddling a cleaning cloth and spray in her hands and greeted him with a nervous nod. Ana gets nervous and fidgets with the cleaner and cloth when questioned. Winston could sense palpable unease as he started asking her questions.\\\\"Ana, did you not join Mack and the others for bungee jumping today?" Winston questioned, to which she responded, "I signed up to jump. But I didn't end up going through with it."\\\\"Any particular reason you didn't join the others, Ana?" Winston proceeded. \\\\Ana took a deep breath, "Well sir, my faith doesn't really permit bungee jumping. Truth be told, I was persuaded strongly by Mack. I had even signed up out of peer pressure but couldn't push myself."\\\\It was true – Mack was insisting that everyone in the group should bungee jump. Mack had reportedly also been vocal about ridiculing Ana’s faith, even encouraging others to join him in doing so. It was a significant factor in their relationship.\\\\"Ana, did you and Mack leave in the same car for the bungee jumping event this morning?" Winston gently pushed further.\\\\"Yes. Yes, we did. We always carpool." She responded while anxiously using the cleaner and cloth on her car’s dashboard. Her eyes flickered nervously back to Winston, expecting the next question.\\\\Winston took a deep breath, standing up to leave, "Alright Ana, that should cover everything for now. We'll be in touch."\\\\Ana nervously nodded without looking up from her cleaning, wringing the cloth repeatedly as Winston walked away, left again with another piece to the enigmatic puzzle of Mack's murder.\\\\The day was getting older and Winston was getting more tired, but the case was fresh, and he wasn't one to back down. He tugged on his coat as he approached the bashful teen waiting for him by the police station.\\\\"Mackenzie, it is?" he asked, extending his hand.\\\\"Yeah, that's right." The slight lisp, overlaid with blanket anxiety, confirmed what the school reports suggested.\\\\"You were at the site when Mack... erm... you know," Winston's voice was methodical, calm -- almost robotic. The suspicion on Mackenzie was not unfounded - the security cameras showed him buying nunchaku a week before. \\\\Mackenzie shifted on his feet, looking away before answering, "Yeah, I was there."\\\\Winston pulled out a small notebook, "What were you doing there, Mackenzie?”\\\\“Bungee jumping, like Mack… Then I left. I didn't... I didn't do anything…” Mackenzie replied.\\\\Internally, Winston sighed at the never-ending waterfall of teenage angst this case was turning into. \\\\“Martial arts, huh?” Winston segued, gesturing to a bruise on Mackenzie’s knuckles. “Nunchaku particularly, I see? Training does include the use of those, correct?” \\\\The change in Mackenzie’s demeanor mirrored the bitterness in the last month’s weather – dark eyes replaced with ice-cold ones. “Yeah,” he admitted, shrinking slightly.\\\\Mackenzie always took pride in being the best at everything. So when Mack got everything he wanted - the promotion to team captain, the respect, the attention - it was a hard pill for Mackenzie to swallow. Winston remembered the team talk, Mackenzie was indeed the top candidate but it had gone to Mack instead. \\\\What clinched it was Mackenzie’s remarks about Mack, echoing whispers of dispute and bickering, lost in the crowded lunchroom. There were also multiple witness reports of the two seen arguing at the bungee jumping site previously. Mackenzie had indeed said disparaging, almost emotional things about Mack – all stemming from a potent brew of jealousy, Winston inferred.\\\\Shifting later through the detritus of Mackenzie's life, Winston discovered the nunchaku that matched the forensics report. They were tucked away, but the layer of dust suggested they weren't a favored possession anymore. It wasn’t hidden, it was misplaced – discarded in the throes of developing maturity.\\\\As the sun started to set, Winston could see witnesses, scattered across the park, repeatedly pointing to the bungee jumping scaffolding. It occurred to him, then, the narrative of the past days. Mackenzie, jealous and wronged, over and over, at the same sight. It was quite a sight. \\\\Winston, shuffling back to the station, was left with one thought - Looks like Mackenzie had quite an eventful week.\\\\
Who is the most likely murderer?\\
1 - Mackenzie\\
2 - Ana\\
\\
You must pick one option. Finally, the last thing you generate should be "ANSWER: (your answer here, include the choice number)"
\end{softbox}
\vspace{.5em}

\begin{softbox}[An example from MuSR OP]
In the heart of the bustling studio, Ricky, Emma, and Danny readied themselves for a day of creating magic. Ricky, holding the helm as the gifted singer-songwriter, was poised for perfection, his precious notebook of lyrics awaiting its call to duty on the producer's desk. Emma, their dutiful and talented producer, was just as eager to breathe life into Ricky's lyrics. She was cognizant of the notebook's place at her desk, awaiting the melodies they would cultivate together.\\\\Across the room, Danny, the diligent studio assistant, was doing his due diligence, keeping the earphones nestled in the recording booth. His aim was to ensure an optimized and meticulous environment for recording, a testament to his commitment to their shared mission. They were all aware of the arrangement — the notebook on the producer's desk, the earphones in the recording booth. Their shared consciousness of these items only intensified the anticipation; they were eager to turn the contents of a weathered notebook into a world-class album.\\\\Ricky, with his weathered notebook of potent lyrics in hand, gently places it onto the piano. An air of creativity and anticipation lingers in the room, everyone aware that this was the first instrumental step in the creation of their masterpiece. In sync with the palpable creative energy, Ricky was engrossed in perfecting the rhythm of his song, preparing himself for an intense day ahead. Not too far away, Emma was sincerely engrossed in her role of musically steering the session. She was focussed on Ricky’s progress, her eyes constantly monitoring him and her mind alive with ideas to enhance the music. \\\\Meanwhile, Danny was diligently covering every corner of the studio. He was making his rounds, ensuring that the studio was prim and proper for Ricky’s crucial session. As part of his tasks, he passed by Ricky several times, always careful not to interrupt the artist's flow.\\\\\\Emma, engrossed in her thoughts, deftly moves the earphones to the producer's desk. She is preparing to tweak the sound settings, pre-empting Ricky's need for perfect audio in his performance. Diverting from his rounds, Danny found himself in the midst of a stirring conversation with a visiting sound engineer. Knowledge flowed between them, illuminating the studio's atmosphere, the engineer's insight bringing a new perspective into Danny's role. Ricky, ensconced in his own world, was in deep discussion with the blank page before him. The daunting silence of the empty studio buzzed with his focus, as he honed his lyrics to perfection in a space separate from the producer's. The visitor, oblivious to the careful choreography of the studio session, stood blocking Danny's general overview of the studio space.\\\\Delicately lifting Ricky's notebook, Danny orchestrates its move to the producer's desk. At the desk, he glimpses a pair of earphones indirectly drawing his attention amidst his routine of tidying up. Emma, from the isolated interior of a sound-proofed booth, lent her ears diligently to already recorded tracks, pouring over them for any room for improvement. Being lost in the music was her way of paying homage to her craft - an unspoken ritual she followed each time she embarked on a music production journey. The entirety of her focus was consumed by the musical notes and rhythm filtering through the studio speakers.\\\\Concurrently, Ricky was absorbed in the act of playing his guitar. His fingers navigated deftly over the strings, lost in an intimate dance with the instrument. As he played, the melodic strums reverberated throughout the studio, filling it with an infectious pulse that hinted at the birth of yet another musical masterpiece. Despite the flurry of activity around him, Ricky was lost in a world of his own, operating on a singular vision of delivering his best performance.\\\\In the meantime, Danny was continuing his cautious management of the studio, ensuring that everything fell into place for the optimum recording session. His watchful eyes were scanning every corner, taking stock of the minor details that could impact the session. However, the design of the studio didn't allow for an unrestricted view into all the corners. The sound booth, where Emma was engrossed in her work, was out of his visual range. The seclusion provided by the booth, although crucial for immersive work, also acted as a barrier for Danny's comprehensive vigilance.\\\\As the day progressed, the studio was entwined in a concerted symphony of dedication and workmanship, the trio, each engrossed in their pursuit, working together to create the best version of Ricky's impending album. As the final note of the day rang through the studio, each person revelled in the satisfaction of another day done right, another step closer towards the realization of Ricky's artistic vision.\\\\Within the dynamic dance of the day's events, the relationships of the trio sang a compelling tune. Each individual played their crucial part in the creation of the impending masterpiece - Ricky with his raw talent, Emma with her passion for perfection, and Danny with his meticulous eye for detail. And as the lights faded on another day of creation, they could sense the beginning of an important chapter in their artistry, a silence collecting the scattered notes of the day, signing off on another critical step in the journey of Ricky's upcoming album.\\\\
Which location is the most likely place Danny would look to find the earphones given the story?\\
1 - piano\\
2 - producer's desk\\
3 - recording booth\\
\\
You must pick one option. Finally, the last thing you generate should be "ANSWER: (your answer here, include the choice number)"
\end{softbox}
\vspace{.5em}

\begin{softbox}[An example from MuSR TA]
Amidst the vibrant chaos of the Redwood Zoo, nestled in the heart of the city's sprawling jungle, the task of assigning roles was a crucial cog in the machinery of its operation. As the manager, the responsibility of allocating Olivia, Alex, and Mia to the positions of Animal Caretaker and Exhibit Cleaner presented an intriguing conundrum. Each individual, with their distinct personalities and skill sets, added a layer of complexity to this assignment puzzle.\\\\Let's begin with Alex, the tall lad with bright eyes, whose history with the mighty beast of the animal kingdom, lacked a certain comfort. The lad, known to express an almost innate unease around animals larger than him, fell short of the prerequisites for an Animal Caretaker. His comfort zone extended to the four-legged companions in our homes, a sentiment I withheld from the petting zoo section of our park. Yet, his association and collaboration with Mia had seen quite the successes in their high school club's fundraising initiatives.\\\\However, his relationship with the gentle Olivia was not as seamless. Alex often mentioned feeling ostracized due to Olivia's tendency to maintain her distance. This seemingly innocent avoidance stirred disquiet within our hushed ranks. And all this, stemming from a disagreement rooted in their previous shared workplaces. Unresolved perhaps, but a factor nonetheless.\\\\Then, there was Mia, the determined bright spark, whose affinity for cleanliness would often bemuse us. She would spend her spare time in her immaculate home cleaning and reorganizing, while her enthusiasm for a spotless Exhibit could not be underestimated. However, her overly thorough methods would invariably result in clashes with Olivia, who criticized her for crossing some form of unspoken boundary.\\\\Mia too had her phobias, the gravelly roars of the zoo's majestic lion had once left her shaken and worried. Loud noises had a similar effect leaving her in a state of nervous terror, much like that of the petite animals held within our barriers. Yet, she was all smiles and peasant conversation around Alex during lunch breaks, sharing a sense of humor that lightened the mood of our everyday grind.\\\\Finally, subdued Olivia, a soul strangled with allergies, and a deep-seated fear for wild animals. An incident with a chimp in her past wove tales of nightmarish betrayal, enough to send her away from the animal exhibits during her zoo visits. Potent elements of dust and pollen resulted in uncontrolled sneezing fits, a remainder from her days at the school as a custodian, responsible for the cleanliness and maintenance.\\\\Three souls; Animals to be cared for, Exhibits to be cleaned. Assigning them was always going to be an enigma for anyone navigating the zoological labyrinth. Love for animals, discomfort, alliances, conflicts; each factor extraordinarily crucial in shaping not just the overall productivity but also the personal growth of each of these individuals at the Redwood Zoo.\\\\
Given the story, how would you uniquely allocate each person to make sure both tasks are accomplished efficiently?\\
1 - Animal Caretaker: Alex, Exhibit Cleaner: Mia and Olivia\\
2 - Animal Caretaker: Olivia, Exhibit Cleaner: Alex and Mia\\
3 - Animal Caretaker: Mia, Exhibit Cleaner: Alex and Olivia\\
\\
You must pick one option. Finally, the last thing you generate should be "ANSWER: (your answer here, include the choice number)"
\end{softbox}
\vspace{.5em}

\begin{softbox}[An example from MoreHopQA]
Context: Maroon 5: Maroon 5 is an American pop rock band that originated in Los Angeles, California. It currently consists of lead vocalist Adam Levine, keyboardist and rhythm guitarist Jesse Carmichael, bassist Mickey Madden, lead guitarist James Valentine, drummer Matt Flynn and keyboardist PJ Morton. What Lovers Do: "What Lovers Do" is a song by American pop rock band Maroon 5 featuring American R\&B singer Sza. It was released on August 30, 2017, as the third single from the band's upcoming sixth studio album (2017). The song contains an interpolation of the 2016 song "Sexual" by Neiked featuring Dyo, therefore Victor Rådström, Dyo and Elina Stridh are credited as songwriters.\\
Question: What is the ASCII code of the last letter of the first name of the current drummer of the band who did the song "What Lovers Do"?\\
The last thing you generate should be "ANSWER: (your answer here)"
\end{softbox}
\vspace{.5em}

\begin{softbox}[An example from iLLC L1-4]
Question: Take the last letters of the words in “Taylor Nancy Quintin Earl” and concatenate them.\\
The last thing you generate should be "ANSWER: (the concatenated letters without any spaces or special characters)"
\end{softbox}
\vspace{.5em}

\begin{softbox}[An example from iLLC L2-4]
Question: Take the second-to-last letters of the words in “Taylor Nancy Quintin Earl” and concatenate them.\\
The last thing you generate should be "ANSWER: (the concatenated letters without any spaces or special characters)"
\end{softbox}
\vspace{.5em}